\documentclass[11pt]{article}

\usepackage{iclr2027_conference}
\iclrfinalcopy
\usepackage{times}
\usepackage{latexsym}

\usepackage[T1]{fontenc}

\usepackage[utf8]{inputenc}

\usepackage{microtype}

\usepackage{inconsolata}

\usepackage{graphicx}

\usepackage{url}

\usepackage{hyperref}
\usepackage{pgffor}
\usepackage{xstring}
\usepackage{float}
\usepackage{subcaption}

\usepackage{booktabs}
\usepackage{tabularx}
\usepackage{array}
\usepackage{makecell}
\usepackage{amsmath}
\usepackage{amssymb}
\usepackage[table,svgnames]{xcolor}
\usepackage{longtable}
\usepackage{etoc}
\usepackage{CJKutf8}
\usepackage[most]{tcolorbox}

\newtcolorbox[auto counter]{finding}[1][]{%
  colback=black!3, colframe=black!60, boxrule=0.5pt,
  left=6pt, right=6pt, top=5pt, bottom=5pt, arc=2pt,
  title={\textbf{Finding~\thetcbcounter}}, fonttitle=\small,
  coltitle=black, colbacktitle=black!8, #1}

\hypersetup{
  colorlinks=true,
  linkcolor=DarkSlateGrey,
  citecolor=DarkSlateGrey,
  filecolor=DarkSlateGrey,
  urlcolor=DarkSlateGrey,
  breaklinks=true,
}

\let\cite\citep

\title{Multilingual GSM-Symbolic: What determines capability transfer across languages?}

\newlength{\authw}
\newcommand{\af}[1]{\textsuperscript{#1}}

\newcommand{\authorblock}{%
\parbox[t]{\dimexpr\textwidth-2\tabcolsep}{\vspace{-1.5em}%
\begin{tabular}[t]{@{}p{\authw}p{\authw}p{\authw}@{}}
Kenneth Enevoldsen\af{1,2, *}     & Riley Herchert\af{3, *}          & Sofie Mosegaard\af{4,2} \\
Dan Saattrup Smart\af{5,2}     & Simon Enni\af{1,2}            & Isaac Chung\af{6} \\
Sofie Bruun\af{4,2}            & Ayush Sunil Munot\af{7}       & Max M{\"u}ller-Eberstein\af{8,9} \\
Adnan El-Assadi\af{10}         & Elisa Bassignana\af{11,9}     & Gianluca Barmina\af{12,2} \\
Hafsteinn Einarsson\af{13}     & Iben Nyholm Debess\af{14}     & Linda Freienthal\af{15} \\
Lukas Galke Poech\af{12,2}     & Mike Zhang\af{16,2}           & Nicolas Legrand\af{1,2} \\
Vladimir Salnikov\af{1,2}      & Yevhen Kostiuk\af{1,2}        & Zafar Hussain\af{1,2} \\
Sagandeep Kaur\af{17}          & Agnes Toftg{\aa}rd\af{18}     & Marie Mattson\af{18} \\
Kristoffer Nielbo\af{1,2}      &                               & \\
\end{tabular}
\par\vspace{0.6em}
{\normalfont\footnotesize\raggedright
\af{*}First authors,
\af{1}Aarhus University,
\af{2}Danish Foundation Models,
\af{3}University of Alabama,
\af{4}Alexandra Institute,
\af{5}Syv.ai,
\af{6}Foam.io,
\af{7}Indian Institute of Technology Kharagpur,
\af{8}The University of Tokyo,
\af{9}IT University of Copenhagen,
\af{10}Massachusetts General Hospital,
\af{11}Bocconi University,
\af{12}University of Southern Denmark,
\af{13}University of Iceland,
\af{14}University of the Faroe Islands,
\af{15}Zendesk,
\af{16}University of Copenhagen,
\af{17}Indian Institute of Technology Madras,
\af{18}National Library of Sweden
\par\vspace{0.3em}
Correspondence: \texttt{kenneth.enevoldsen@cas.au.dk}\par}%
}}

\author{\authorblock}

\begin{document}
\maketitle
\lhead{Preprint}
\begin{abstract}
We understand little about how capabilities acquired in one language carry over to another, or what governs this transfer: evaluations rely on incomparable, saturation-prone datasets and rarely examine its determinants jointly. 
Identifying what predicts transfer would let us avoid exhaustive evaluation across all language pairs and let developers target the factors that limit performance in low-resource languages.
To evaluate cross-lingual capability transfer, we introduce \textsc{Multilingual GSM-Symbolic}, an extensible multilingual mathematical dataset covering 30,000 item-matched question--answer pairs and spanning 15 languages. It utilises symbolic templates to prevent overfitting and ensure generalisation by allowing generation of millions of high-quality variations from a single sample.
Using \textsc{Multilingual GSM-Symbolic}, we quantify the largest determinants of capability as \emph{model size} ($\beta=1.77$), 
\emph{language resource level} ($\beta=0.77$),
\emph{reasoning} ($\beta=0.67$) and 
\emph{typological distance} ($\beta=-0.25$). This joint estimation allows these determinants to be expressed in terms of one another: a 32B model evaluated in Marathi performs like a 10B model in English. 
Our findings have important implications for model developers, showing that model size and reasoning narrow the performance gap between low- and high-resource languages ($\beta=-0.27$, $\beta=-0.20$, respectively), while similar levers have little or no effect on typologically distant languages.

Overall, our analysis framework explains 92\% of between-language variation, but only 23\% of the model-by-language variation, and predicts a model's performance on an unseen language within 6.0pp ($r = .96$). Incorporating measurements from just 10 templates in the target language reduces this to 4.19pp, enabling reasonable estimates of performance with little or no downstream dataset.

\noindent
We release the full dataset\footnote{\url{https://github.com/centre-for-humanities-computing/multilingual-gsm-symbolic}} and code\footnote{\url{https://huggingface.co/datasets/danish-foundation-models/multilingual-gsm-symbolic}} to support future research on, and more evaluation of, cross-lingual capability transfer.
\end{abstract}

\section{Introduction}
    \begin{figure}
    \centering 
    \includegraphics[width=\linewidth]{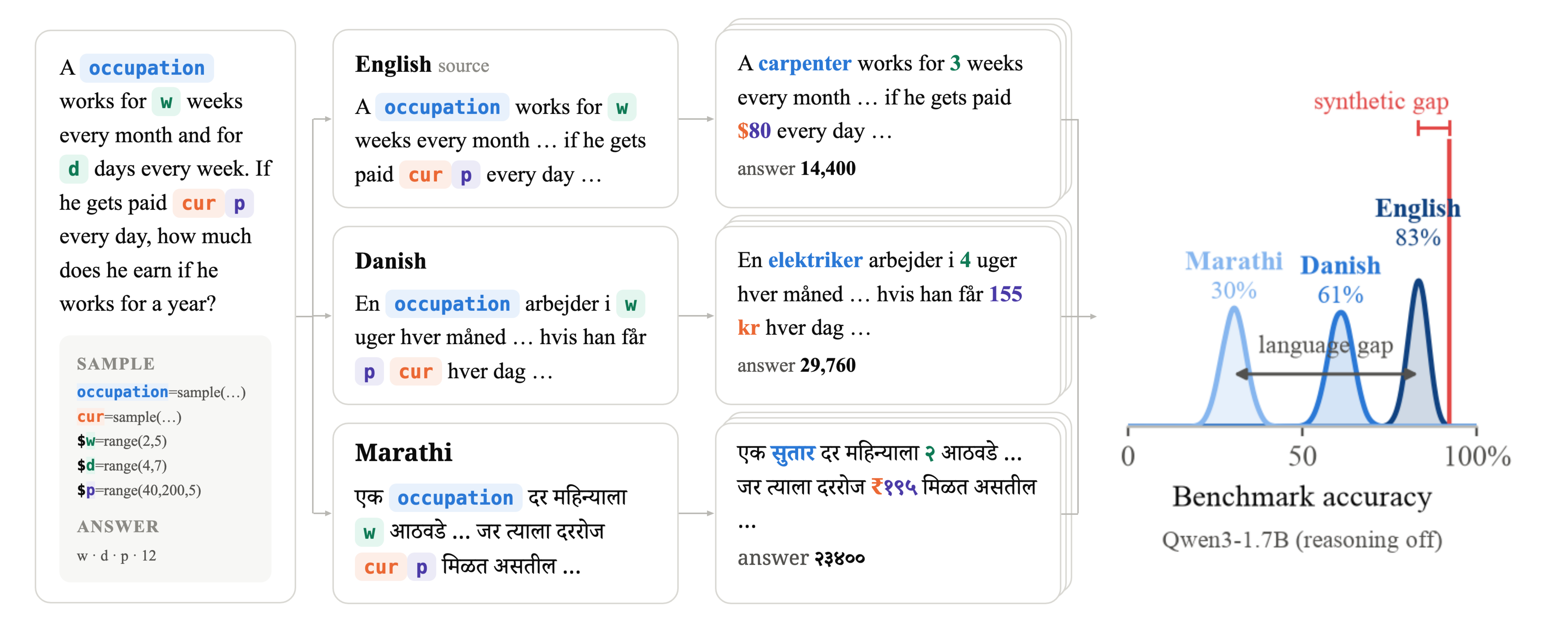}
    \caption{\textbf{Overview of \textsc{Multilingual GSM-Symbolic}}: Consisting of 100 samples in 15 languages, multilingual GSM-Symbolic consists of 1500 verified and localized templates. We sample 20 per template for a total of 30,000 and show an example evaluation of Qwen2.5-7B across English, Danish and Marathi. Showing both the \textit{language gap} between Marathi and English and \textit{synthetic gap} between the original samples and their template samples distributions.}
    \label{fig:gsm-symbolic-templates}
\end{figure}

While modern LLMs show remarkable abilities to solve a wide variety of natural language problems, the transfer of their capabilities across languages is understudied. In order to build better models -- especially for low-resource languages -- a better understanding of the dynamics of capability transfer can reduce evaluation cost by avoiding costly exhaustive evaluation across all language pairs and allow model developers to target factors known to improve capability transfer.

To explore what determines transfer, it is important to isolate central varying components that can contribute to the variance across languages. For instance, in domains like medical question answering, the answers might differ between languages as medical practices vary\footnote{For example, recommended drug choices and pre-prescription tests can differ by region because of population-level differences in adverse-reaction risk.}. 
Contrast this with mathematics, where the answer to any given problem remains fixed across languages, providing an ideal test-bed. However, evaluating models on mathematical benchmarks in various languages does not tell us anything about how a model's capabilities vary across languages, as different benchmarks vary across multiple axes, including formatting, difficulty and subdomains. As such having item-matched samples across languages is required for comparable results and to model how capabilities transfer.

Some datasets already exist which cover mathematics and have item-level matched samples across languages, these are both static and popular, making it hard to isolate capabilities from leakage \cite{Shi2022LanguageMA, singh2025globalmmluunderstandingaddressing}. To overcome this, we utilize a template approach akin to \citet{Mirzadeh2024GSMSymbolicUT}, where a singular template can be used to generate millions of verifiable variants and introduce \textsc{Multilingual GSM-Symbolic}, a benchmark of item-matched word-problems templates across 15 languages, translated with extensive manual review and automatic validation.

We use this benchmark to study how capabilities transfer predicting performance using two classes of features, namely, model-level features such as size, and language-level features such as typological distance. While prior work has examined some of these dimensions
none has coherently modeled transfer across model features and their interactions.

\textbf{Our main contributions are as follows}:
\textit{1) We introduce \textsc{multilingual GSM-Symbolic}}, the first verified, symbolic framework that includes the code making it possible to extend and resample the dataset, \textit{2) we model how capabilities transfer across languages} and determine actionable interventions in order of effectiveness, \textit{3) we show that we can predict performance on an unseen language} to a reasonable extent and that this prediction can be improved drastically by including just 10 templates. These contributions trace a path from headline multilingual accuracy to in-depth analysis of why transfer succeeds or fails.

\section{Related work}
    \subsection{Multilingual Mathematical Reasoning}

\begin{table*}
    \centering
    \scriptsize
    \setlength{\tabcolsep}{1.8pt}
    \renewcommand{\arraystretch}{1.2}

    \begin{tabular}{lcccccccc}
        \toprule
    
        &&&&& \multicolumn{4}{c}{Validation}\\
        \cline{6-9}        
        \textbf{Dataset}
        & Languages & Symbolic & Code & Extensible
        &         Error Analysis
        & Translation & Localization
        & Validity\\
        \midrule

        \rowcolor{gray!15}\textsc{Multilingual GSM-symbolic} (ours) & 15 (100)
        & \checkmark& \checkmark& \checkmark& \checkmark & \checkmark
        & \checkmark
        & \checkmark\\

        \textbf{Previous work}& \\
        \hspace{0.2cm} MGSM \cite{Shi2022LanguageMA}
        & 11
        & & & & & \checkmark & (\checkmark)
        & \\

        \hspace{0.2cm} GSM-Symbolic \cite{Mirzadeh2024GSMSymbolicUT}
        & 1
        & \checkmark& && & & \\

        \hspace{0.2cm} GSM8K-Platinum
        \cite{vendrow2025largelanguagemodelbenchmarks}
        & 1
        & & & & \checkmark & 
        & \\

        \textbf{Parallel work}& \\
        \hspace{0.2cm} mGSM-Symbolic \cite{ranaldi-pucci-2025-multilingual}
        & 11
        & & && & \checkmark & 
        & \\

        \hspace{0.2cm} MGSM-Pro \cite{xu2026mgsmprosimplestrategyrobust}
        & 9
        & \checkmark& & & & \checkmark & (\checkmark)
        & \checkmark \\
\bottomrule
    \end{tabular}

    \caption{
        \textbf{Comparison of GSM-style mathematical reasoning benchmarks}.
        The parenthesis refer to languages that have not been human validated yet. For more see \ref{additional_languages}
        Our dataset combines multilingual templates, controllable variation and extensibility with extensive validation.
        Error analysis refer to the process of running a model and correcting ambiguous or incorrect questions.
        Extensible as defined by \citet{dynaword}, indicate that there is a process for adding to, extending or correcting the dataset. 
        Validity denote automated tests that ensures the quality of the generated examples. While MGSM and MGSM-Pro localizes languages they do not localize names, currency and metrics even though imperial units are rarely used in a non-English contexts.
    }
    \label{tab:gsm-dataset-comparison}
\end{table*}

Mathematical problems are often used to study reasoning capability in LLMs, because multi-step reasoning is an effective strategy for solving such problems, and because answers to these problems can be objectively verified \cite{wen2026reinforcement, cobbe2021gsm8k, lightman2024let}. Additionally, mathematics 
remains consistent across cultural contexts, compared to, e.g., social or medical reasoning which are influenced by the context in which they operate \cite{MCNAMARA201995, Napier2014CultureAH}. 
Grade school math is especially useful because the underlying mathematical operations are elementary, placing greater emphasis on whether models can construct the underlying reasoning procedure than their computational abilities. 
GSM8K \cite{cobbe2021gsm8k} and its variants \cite{Shi2022LanguageMA, vendrow2025largelanguagemodelbenchmarks} comprise a family of benchmarks consisting of grade school level arithmetic word problems requiring multi-step reasoning. GSM8K is limited by its reliance on a static dataset: each problem appears with fixed text and values. This makes it difficult to separate underlying reasoning performance from sensitivity to surface form \cite{Mirzadeh2024GSMSymbolicUT} and memorization enabled by potential contamination \cite{zhang2024carefulexaminationlargelanguage}. 
Several datasets have sought to address this and we provide an overview and comparison to our dataset in Table~\ref{tab:gsm-dataset-comparison} along with a brief introduction in the following. GSM-Symbolic \cite{Mirzadeh2024GSMSymbolicUT} turns GSM8K questions into templates, and demonstrates that surface-level substitutions of names and numbers significantly reduce performance. Efforts have also been made to extend evaluation to other languages. MGSM \cite{Shi2022LanguageMA} translates a 250-question subset of GSM8K into multiple languages. While MGSM is useful for comparing headline performance across languages, its language coverage and static nature are limitations. MGSM-Pro \cite{xu2026mgsmprosimplestrategyrobust} extends MGSM using templates and introduces symbolic and irrelevant context sets, but does not release the templates or sample generation code. 
mGSM-Symbolic \cite{ranaldi-pucci-2025-multilingual}, despite its name, does not utilize templates but translated instances derived from GSM-Symbolic and as such remains a static benchmark. Similarly, there is limited description of any validation of final samples.
Neither address known errors and shortcormings in GSM8K \cite{vendrow2025largelanguagemodelbenchmarks}.
Broader benchmarks such as Global MMLU \citep{singh2025globalmmluunderstandingaddressing} are unsuitable here, as multiple-choice formats admit choice-based shortcuts \cite{chandak2025answermatchingoutperformsmultiple} and culturally specific knowledge confounds the comparison.

\subsection{What determines transfer?}

MGSM investigated the impact of model scale on transfer, finding that relative transfer efficiency increases with model size. Furthermore, work on pretraining for reasoning models has demonstrated that increasing model size reduces relative transfer gaps, and, additionally, shows that target-language data had a high impact on performance \cite{barua2026long}. Other work has also examined the impact of language features on transfer. Experiments on transformer classifiers have shown significant correlation between target-language performance and pretraining token frequency as well as language similarity to English \cite{lauscher2020zeroherolimitationszeroshot,blevins-zettlemoyer-2022-language}. Similarly, \citet{chang-etal-2024-multilinguality} demonstrated that adding multilingual data improved non-English performance and that this improvement was based on the syntactic similarity of the additions.

This prior work suggests that both model and language factors impact transfer. However, these factors have rarely been jointly studied.

\section{Method}
    
\subsection{Dataset construction}

\textbf{Source Data and Template Process}:
We derive \textsc{Multilingual GSM-Symbolic} from GSM-Symbolic \cite{Mirzadeh2024GSMSymbolicUT}. As the authors do not provide a parser, we implement and release our own, resolving several inconsistencies and extending the syntax to allow for multilingual support.

\textbf{Language Selection}: We select languages from a diverse set of language families such that no single feature (e.g., resource level, family, or script) is conflated with another. 
To cover diversity across language families, we include Japanese (Japonic), Chinese (Sino-Tibetan), Russian (Slavic), Hindi (Indo-Aryan), French (Romance), Arabic (Semitic), and Estonian (Finno-Ugric). 
To examine varying resource levels, we construct a \textit{Germanic resource axis} spanning Icelandic, Danish, Dutch, German, and English (300k-1B speakers). 
Finally, to extend coverage of the resource axis beyond Germanic, we add Ukrainian, Italian, and Marathi. We examine the feature space further in Appendix~\ref{app:sec:feature-design-space}.

\textbf{Translation and Localization}:
All translations were performed by native speakers with technical expertise and higher education. The translators were 17 males and 7 females between 20-40 years old. Translators were allowed to use LLMs, which all chose to do, typically either GPT-5.4 or Claude Opus 4.8.
Each template includes a creation string, documenting the source language of the translation, the model used and the language-proficiency of the annotator (see Appendix~\ref{app:sec:language-anno-overview}). None of the models evaluated were used for the translation.
As problems in other languages containing English units do not reflect real-world use and have been shown to inflate measured performance \cite{azime2026bridgingculturegapframework}, we localise currencies, units, cultural references and numerals.
We thus define item-level equivalence at the level of the underlying reasoning structure and mathematical operations, which remain constant, rather than at the level of exact problem text, which introduces confounding translation artifacts (see also Appendix~\ref{app:sec:metric-verification}).

\textbf{Validation}:
Since prior work has identified ambiguities and errors in GSM8K \cite{vendrow2025largelanguagemodelbenchmarks}, we apply a series of verifications to templates. These checks were initially applied to English templates, to prevent errors from propagating, and on the translated templates catching introduced errors. After the initial translation and review, we perform an error analysis by generating 20 instances of each problem and manually review all templates that Claude Opus 4.8 (\texttt{xhigh}) answers incorrectly, reviewing reasoning traces to understand failure points.
eThis process is useful for catching unclear or incorrect templates\footnote{For example, one question asks: ``How much more likely is it that he rolls a number greater than 3 than that he rolls two even numbers in a row?''. Because both a relative interpretation (200 percent) and an absolute interpretation (25 percentage points) are valid, we clarify the question by adding ``[...] expressed as percentage points.''}. When applied to the translated templates the template errors were corrected by the native speakers of the respective language.
In addition, all templates are validated using an automated test-suite, that, among other things, ensures that rendered templates using default values match the original and that all template conditions are satisfied, addressing mismatch issues previously identified in GSM-Symbolic \cite{ivanova2025towardsmorerigorous}. These tests were continually updated as new kinds of errors were detected and corrected.

\subsection{Modeling Capability Transfer}
To estimate which language-level features govern cross-lingual transfer we model the performance on individual instances using a generalized linear mixed model (GLMM). Given that an instance is either solved or not, we work with a Bernoulli response and the matched design allows us to separate language effect from difficulty.

Let $y_{mlti} \in \{0, 1\}$ denote whether model $m$ solves instance $i$ of template $t$ in language $l$. We fit a crossed random-effects logistic model

\begin{equation}
\begin{aligned}
y_{mlti} &\sim \mathrm{Bernoulli}(\pi_{mlti}),
\quad
\mathrm{logit}(\pi_{mlti}) = \mathbf{x}_{ml}^{\top}\boldsymbol{\beta}
  + \sum_{g\in\mathcal{G}} u^{(g)}_{j_g(m,l,t)}, \\[4pt]
u^{(g)}_{j} &\overset{\text{iid}}{\sim} \mathcal{N}\!\left(0,\sigma_g^{2}\right),
\quad
\mathcal{G} = \{\text{family},\ \text{model},\ \text{variant},\
  \text{language},\ \text{template},\ \text{model}\times\text{language}\}
\end{aligned}
\label{eq:model}
\end{equation}

where $\mathbf{x}_{ml}$ is a vector of language (e.g. typological distance, resource level), model features (e.g. model size, reasoning) and their interactions. 
while $j_g(m,l,t)$ returns the level of grouping factor $g$ to
which observation $(m,l,t,i)$ belongs. On the model side these maps are nested:
a variant is a model evaluated with reasoning either enabled or disabled,
and each model belongs to exactly one family.

Each random effect absorbs variation left unexplained by the corresponding
features, so that the associated fixed effects are evaluated against the
appropriate source of variability.
The family term captures capability shared across lineage, such as 
pretraining corpus, tokenizer, and post-training recipe, and ensures that families
contributing many checkpoints do not enter as many independent
observations. 
The model term captures checkpoint-specific capability and the variant term seeks to differentiate model with reasoning enabled or not, this term is required to make the reasoning coefficient an estimate of how much reasoning helps \emph{across} base models.
The template and language terms captures difficulty not captured by the language features and ensure that the language features are not inflated. 

The last term model-by-language measures the cross-lingual transfer variation that remains after accounting for model strength and problem difficulty, and the reduction when the language features are added to the fixed effects quantifies how much of the transfer variation those features explain. We examine and discuss further model ablations in Appendix~\ref{app:sec:model-ablations}. The model is fit in R \cite{r2024r} using lme4 and lmerTest \cite{bates2015fitting, kuznetsova2017lmertest}.

\subsection{Features}
\textbf{Language Features}:
For our language features we use language resource level, typological distance from English and tokenizer fertility.
As our language resource proxy we used Common Crawl page count, as it has previously been used as a resource proxy and is widely used during pre-training \cite{bang-etal-2023-multitask, lai-etal-2023-chatgpt}. We apply log transform to reflect the heavy-tailed distribution of language resources \cite{joshiStateFateLinguistic2020}.

Typological distance from English is measured using cosine distance of 103 morphosyntactic features from URIEL \cite{littell2017uriel}, which include,  e.g., word order, adposition order, case marking.
Fertility is measured on the templates of the target language. To prevent collinearity with typological distance, fertility is normalized as the model deviation from the language mean as it is a relative measure of how well the model tokenizes the target language compared to its peers. 

\textbf{Model Features}:
For our model specific features we use size and reasoning due to their influence on performance and previous studies denoting their influence on transfer \cite{Shi2022LanguageMA}. Model size is log-transformed and reasoning is denoted as a boolean.

\textbf{Interaction terms}:
In addition to our main effects, we postulate that our model features interact with language features. This allow us to model, e.g., whether influence of resource level disproportionately impacts smaller models or whether reasoning simply improves performance without reducing the gap between languages.

\subsection{Model selection}

We restrict the suite to open-weight, instruction-tuned models \cite{chung2022scalinginstructionfinetunedlanguagemodels}, to allow for zero-shot prompting and ensure access to the tokenizer, inspectable reasoning traces, and, for some families, exact pre-training data.

Models are organised in a family $\times$ size grid, letting us estimate the effect of scale within a family while holding architecture, tokenizer, and pretraining mixture fixed; we therefore prioritize families spanning a wide range of sizes.

To examine the effect of reasoning, we include the Olmo 3 Think series \cite{olmo2026olmo3}, Qwen 3 and 3.5 \cite{yang2025qwen3technicalreport, qwen35blog}, and Granite 3.2 \cite{soule2025granite32}. To allow inspection of pretraining data, we include EuroLLM \cite{martins2024eurollmmultilinguallanguagemodels, ramos2026eurollm, martins2025eurollm}, Apertus \cite{apertus2025apertusdemocratizingopencompliant}, and OLMo 2 and 3 \cite{olmo20252olmo2furious, olmo2026olmo3}, which also span 50\%, 40\%, and 90\% English pretraining data, respectively. Finally, Qwen 2.5 \cite{qwen2025qwen25technicalreport} and Gemma 3 \cite{gemmateam2025gemma3technicalreport} broaden family coverage. Model ids and revisions are listed in Appendix~\ref{app:sec:model-ids}. We note that the goal is not to select the best-performing models, but those that best allow us to estimate the target variables\footnote{
Two models are removed; EuroLLM-1.7B as it consistently exceeded its own limit of 4,096 tokens (54\% of cases) and Qwen 3.5 0.8B (reasoning=on) as it consistently (95\% of cases) produces near-empty completions. Results for excluded models are in Appendix~\ref{app:sec:model-results-by-language}.
}.

\subsection{Evaluation}

Experiments were performed using Inspect AI \cite{aisi2024inspectai} and vLLM 0.22.1 \cite{kwon2023efficient}. Experiments were run with zero-shot prompting with human-translated versions of the English prompt: \textit{``Solve the following math problem step by step. Put your final numeric answer (without units, currency symbols, or percentage signs) in \textbackslash{}boxed\{\}, for example: \textbackslash{}boxed\{42\}"}. Responses without a parseable answer are scored incorrect, as reporting the answer as asked is part of the task.

\section{Results and Discussion}
    
\subsection{What Determines Capabilities?}

\begin{figure}[t]
\begin{minipage}[t]{0.59\linewidth}
    \centering
        \centering
        \includegraphics[width=\linewidth]{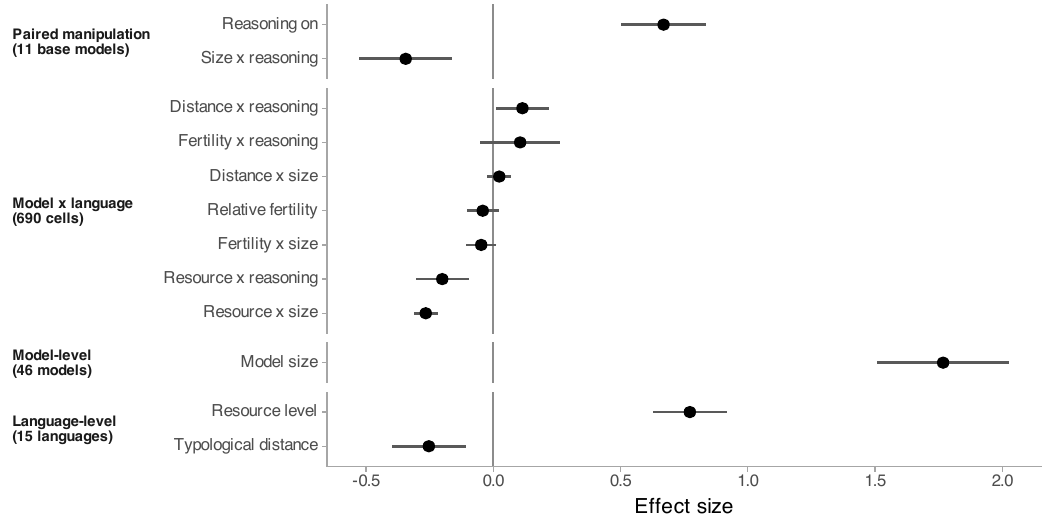}
        \caption{\textbf{Fixed effects} and their 95\% Wald interval. Effect sizes are normalized to log-odds per standard deviation (SD). The full fit can be seen in Appendix~\ref{app:sec:glmm-estimates}.}
        \label{fig:main-effects}
\end{minipage}\hfill
\begin{minipage}[t]{0.39\linewidth}
    \centering
    \includegraphics[width=\linewidth]{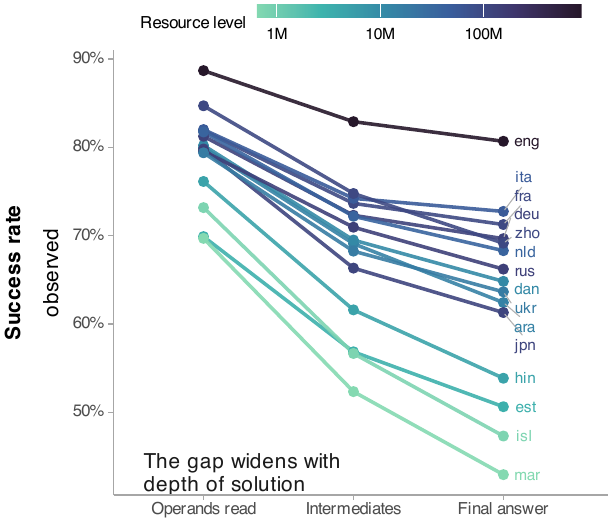}
    \caption{\textbf{Success-rate at depth of solutions}: Language level differences appear at later stages of processing rather than during the initial parsing.}
    \label{fig:ladder}
\end{minipage}
\end{figure}

Figure~\ref{fig:main-effects} shows the fixed effects of our fitted model. Three findings below complement upon those of previous work by extending previous results by estimating the effects jointly and extending them to a multilingual and symbolic setting. We state these briefly before moving on to the central contributions.

\begin{finding}
Larger models are more capable ($\beta_\text{size}=1.88$,
$p < 0.001$) and similarly so for reasoning ($\beta_\text{reasoning}=0.67$,
$p < 0.001$), though the effect reduces with scale ($\beta_{\text{size}\times\text{reasoning}}=-0.34$,
$p < 0.001$)
\end{finding}

These findings are consistent with the scaling experiment of \citet{Shi2022LanguageMA} and \citet{snell2025scaling}.

\begin{finding}
Resource level is the largest language effect. Higher-resource languages are substantially easier
($\beta_\text{resource}=0.77$, $p < 0.001$).
\end{finding}

These are similarly consistent with \citet{Shi2022LanguageMA}, \citet{blevins-zettlemoyer-2022-language} and \citet{barua2026long}.

\begin{finding}
Structurally distant languages transfer worse ($\beta_\text{distance}=-0.25$, $p = 0.001$).
\end{finding}

Which is consistent with \citet{chang-etal-2024-multilinguality}. However, this is strongly correlated with fertility ($r=0.8$). Examining uncorrelated relative fertility, we find no meaningful effect ($\beta_\text{rel. fertility}=-0.04$, $p = .19$).

\subsubsection{Interactions: How do we close the capability gaps?}

\begin{figure}
    \centering
        \centering
        \includegraphics[width=\linewidth]{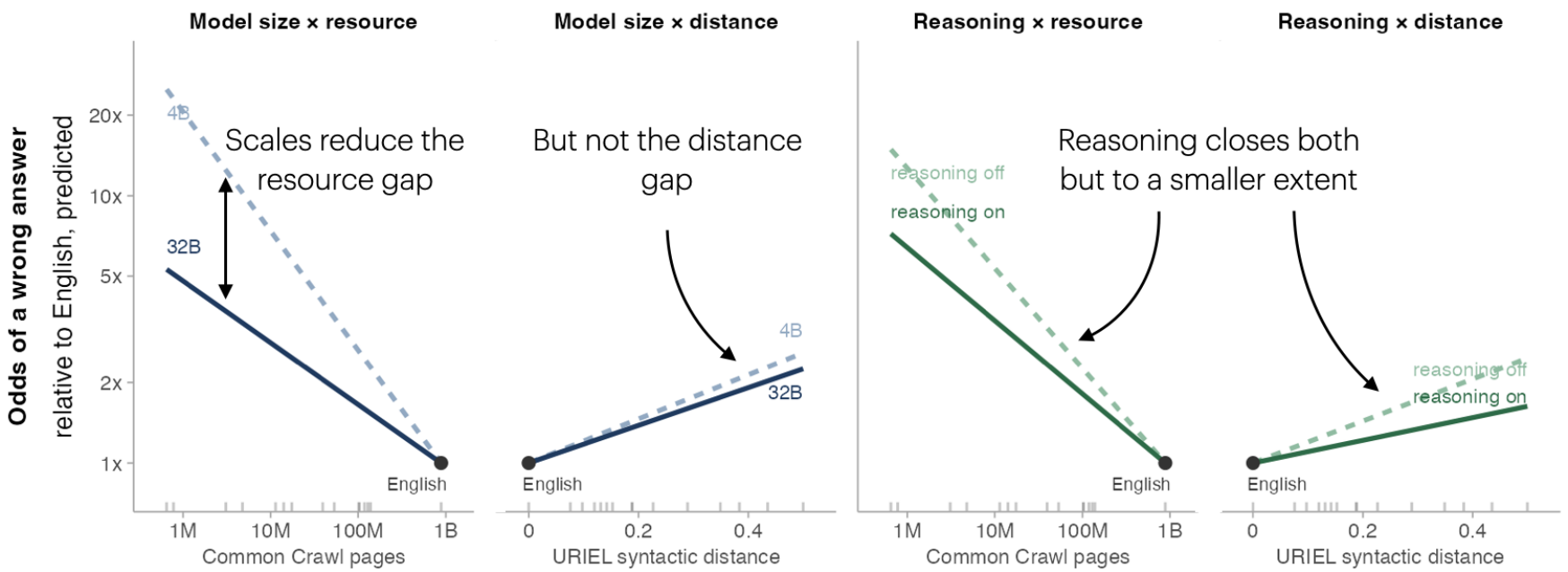}
        \caption{\textbf{Levers}: Model prediction for four interaction terms denoting two levers, showing the effect of reasoning and size across resource level and typological distance. Depending on the language you are working with, the method for improving capability transfer differs.}
        \label{fig:interaction-levers}
\end{figure}

While prior works have examined these factors predicting capability individually modeling these jointly produce two actionable levers shown in Figure~\ref{fig:interaction-levers}. We see that 

\begin{finding}
Model size and reasoning reduces the resources gap ($\beta_{\text{resource}\times\text{size}}=-0.27$, $p < 0.001$, $\beta_{\text{resource}\times\text{reasoning}}=-0.20$, $p < .001$)
\end{finding}

Reasoning has a small effect on distant languages ($\beta_{\text{distance}\times\text{reasoning}}=0.11$, $p = .03$), however, it should be noted that this effect doesn't reproduce under model ablations (see Appendix~\ref{app:sec:model-ablations}). Similarly, model size was not found to affect distance ($\beta_{\text{distance}\times\text{size}}=0.02$, $p = 0.32$).

\subsection{Predicting performance on an unseen language}
Beyond identifying best practices for model developers, modeling capability transfer also lets us estimate performance on unseen languages, which can save compute while still providing plausible performance estimates for languages without evaluation data.

We estimate how well we can predict performance on a language by holding out the given language and refitting eq.~\ref{eq:model} on the remaining and predicting the held-out language.

\begin{figure}[t]
\begin{minipage}[t]{0.49\linewidth}
    \centering
      \includegraphics[width=\linewidth]{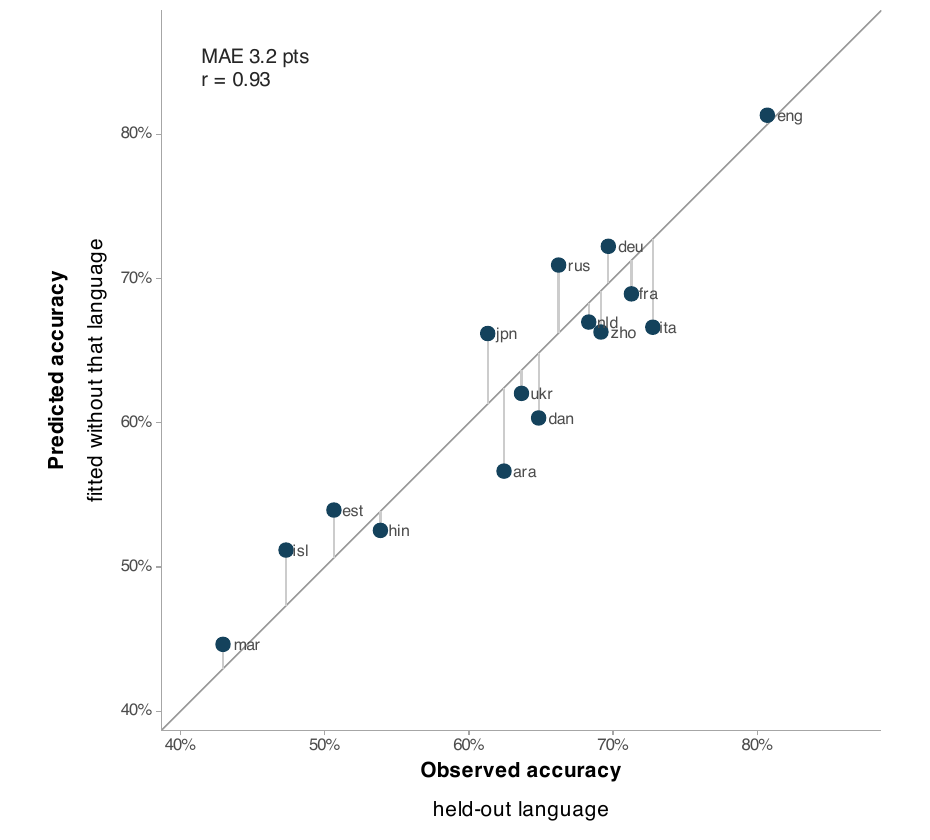}
    \caption{\textbf{Forecasting language performance}: Aggregate language performance with each language held out in turn.}
    \label{fig:predict-language}
\end{minipage}\hfill
\begin{minipage}[t]{0.49\linewidth}
    \centering
    \includegraphics[width=\linewidth]{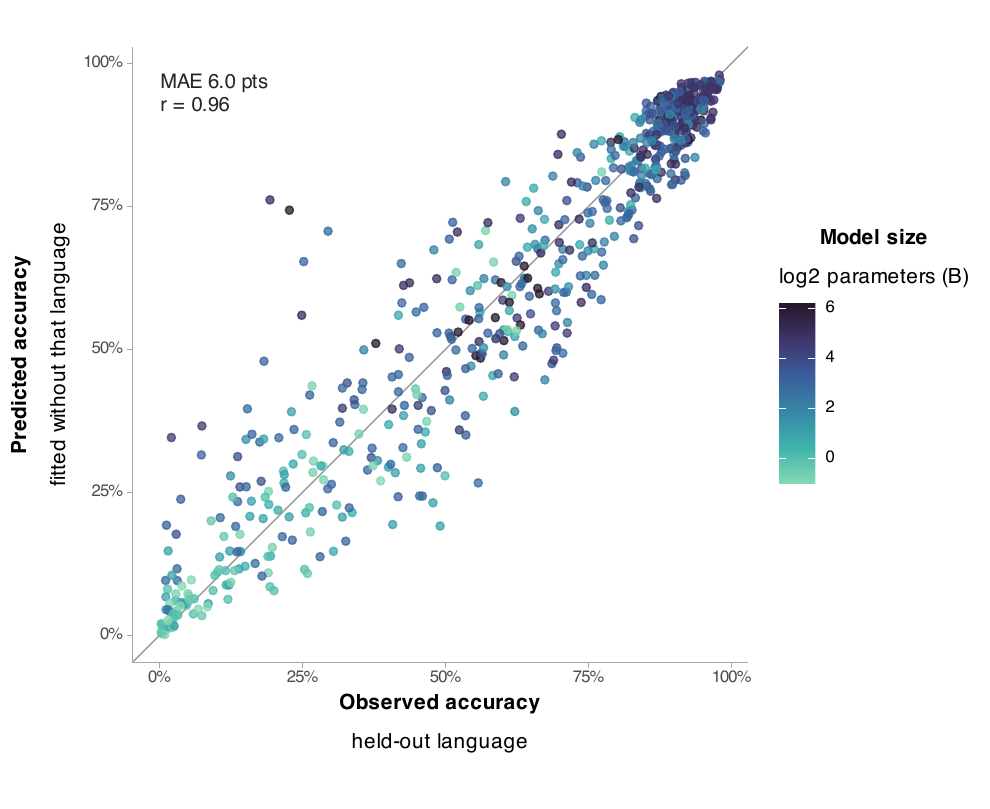}
    \caption{\textbf{Forecasting model performance} on a given language, showing a total of 768 (model, language) pairs. For a per-language decomposition, see Appendix~\ref{app:sec:prediction}.}
    \label{fig:predict-model}
\end{minipage}
\end{figure}

\begin{finding}
A language's average difficulty is predictable to 3.2pp (Pearson $r = 0.93$, Fig.~\ref{fig:predict-language}) and for a specific model in that language the error is 5.98pp (Pearson  $r = 0.96$, Fig.~\ref{fig:predict-model}), constituting a 63\% and 39\% reduction, respectively, over the naïve average baseline. Utilising just 10 templates, we can further reduce it to 2.38pp and 4.19pp, respectively (see Appendix~\ref{app:sec:prediction}).
\end{finding}

The largest failures in the held-out case are model-by-language idiosyncrasies, e.g. Apertus-70B scoring 23\% on English where its size and the language's features imply 75\%, and several large models fall well below prediction on Estonian (see Appendix~\ref{app:sec:prediction}).

Similarly, findings can also be obtained by comparing to a size-only baseline of Eq.~\ref{eq:model}, where we see that adding the language features removes 92\% of the between-language variance but only 23\% of the
model-by-language variance. As such, we can estimate language difficulty with a high degree of confidence, but which specific model falters in a target language remains hard to estimate. However, this can be remedied using only 10 templates.

\subsection{Symbolic variants are harder, but how much is unsystematic}
Similar to previous works \citep{Mirzadeh2024GSMSymbolicUT, xu2026mgsmprosimplestrategyrobust}
we see that symbolic variants are harder than the originals (see
Figure~\ref{fig:symbolic-penalty}). Across 705
(model, language) pairs, accuracy on symbolic instances is 3.14pp lower than on the original GSM8K questions, and the gap is unrelated to the model's accuracy level ($r = -0.02$). 
Fitting both splits jointly (Appendix~\ref{app:synth-vs-orig}) we observe a penalty of $-0.30$ in log-odds ($p < 0.001$). 
Additionally, we find that the penalty varies across languages ($\chi^2 = 21.6$, $p < 0.001$): it is smallest in English (1.7pp, 95\% CI $[-0.6, 4.0]$) and largest in Russian (5.5pp, 95\% CI $[4.6, 6.5]$), though the intervals for individual languages are wide. 
It is, however, orthogonal to the features that predict transfer (resource level $p = 0.68$, typological distance $p = 0.84$, see Appendix~\ref{app:synth-vs-orig}). 
Therefore, the features that explain how well a model does in a language do not predict how fragile it is there.


\subsection{Where in the solution does the gap appear?}
\label{sec:ladder}

One might argue that solving these multilingual problems is simply a matter of parsing the problem. We inspect this by utilizing that our template's gold solution carries the \texttt{<<lhs=rhs>>} calculator annotations and thus we can extract operand reads (\texttt{lhs}), intermediate computations (\texttt{rhs}), along with the final answer.

We see that multilingual performance isn't simply a question of parsing the problem
(operand reads), but is accumulated at every stage (see Figure~\ref{fig:ladder}). Fitting a
model similar to Equation~\ref{eq:model} at each stage (see Appendix~\ref{app:ladder}), we
see that this effect is mostly driven by resource level ($\beta_{operand}=0.32$, $p<.001$, $\beta_{intermediate}=0.45$, $p<.001$, and $\beta_{final}=0.77$, $p<.001$ log-odds per SD), and typological
distance
($\beta_{operand}=-0.08$, $p=.066$, $\beta_{intermediate}=-0.15$, $p=.001$, and $\beta_{final}=-0.25$, $p<.001$).
Similarly we also see that compliance with the prompt (using \verb|\\boxed{}| does vary by language, from $94.5\%$ in German to $77.1\%$ in Marathi, indicating that a models instruction following capabilities are also influenced by the language, supporting the notion that model capabilities degrade at every stage when processing low-resource languages.

\section*{Limitations}
    
\textbf{Size and coverage}: \textsc{Multilingual GSM-Symbolic} utilises 100 symbolic templates across 15 languages. While we have sought to represent a diverse set of languages and utilise templates to ensure diversity, it is likely that the utility would increase if more languages were added. To allow for such an addition, we have made sure to keep the dataset extensible, following \citet{dynaword}. In Appendix~\ref{app:sec:sample-size}, we explore the effect of varying sample size and see that already 50 samples reproduce the rank at 0.99, while at 100 samples a (model, language) prediction still has a standard error of 2.29pp, and it would require 400 templates to halve that. We similarly see that only minor gains can be obtained by further sampling.

\textbf{Saturation}: The problems in the dataset are relatively simple, requiring only grade-school maths, and as such saturation might influence our estimates. In Appendix~\ref{app:sec:saturation}, we re-fit Eq.~\ref{eq:model} at various performance thresholds. We see that our reported finding remains stable.

\textbf{On the generalisation of our results}: Our study focuses on cross-lingual transfer using grade-school mathematics word-problems as their results are comparable across different linguistic and cultural settings. However, the determinants of transfer and their effect sizes may differ across domains.

We limit our model selection to non-base models, so our estimates therefore characterise transfer in \textit{instruction-aligned} models. Similarly, to ensure comparable model sizes, we do not include Mixture of Experts models (MoEs) despite their increasing prevalence \cite{Cai_2025}. Future studies should explore the extent to which these findings generalise across domains, training stages, and model architectures.

\textbf{Feature specifications}: Due to a high degree of correlation between typological distance and fertility, we use relative fertility -- how well a model tokenizes a text compared with its peers. This means that typological distance encodes both the structural distance from English and (non-relative) fertility, so conclusions about typological distance should be interpreted with care.

Similarly, we acknowledge that language resource can’t be described on a single axis; it includes multiple components, such as the amount of data available, the number of speakers, and resources of those speakers (e.g. BNP/capita). Since many LLMs include Common Crawl or similarly crawled data, we believe it to be a reasonable proxy for the resource level of a given language \textit{within} the model. However, even to this extent, Common Crawl is likely imperfect, as the proportion of usable in-language text varies systematically with resource level. Lower-resource language corpora tend to contain a smaller proportion of usable text and are more likely to be misclassified \cite{kreutzer2022quality, caswell-etal-2020-language}. Examining the English-non-English gap in open-data model families in Eq.~\ref{eq:model} also confirms that model-specific numbers for the training data would likely improve fit, with Apertus having a negative average gap (-20.7), EuroLLM the smallest (7.0), and English-dominant Olmo 2 and 3 having the highest (41.7 and 25.0, respectively). However, it is hard to draw conclusions from these models, given that they are conflated with e.g. family, post-training recipe and recency.

\section*{Conclusion}
    We introduce \textsc{Multilingual GSM-Symbolic}, a verified, high-quality and extensible version of GSM-Symbolic covering 15 diverse languages. Using this dataset, we examine how capabilities transfer and show that this influences both model development and evaluation. For model developers, we estimate that the largest determinants of capability are ordered; \emph{model size} ($\beta=1.77$), \emph{language resource level} ($\beta=0.77$), \emph{reasoning} ($\beta=0.67$) and \emph{typological distance} ($\beta=-0.25$). Additionally, we show that both size and reasoning can help reduce the gap between low- and high-resource languages ($\beta=-0.25$, $\beta=-0.20$ respectively), while similar levers have little to no effect on typologically distant languages.

For model evaluators, we found that we can predict a model's performance on an unseen language within 5.98pp, and by including just 10 templates in the target language, we can reduce it to 4.19pp. Similarly, we found that the average difficulty of a language can be predicted down to 3.2pp (2.38pp using 10 templates), enabling reasonable estimates of performance with little or no downstream data.

\section*{Ethics statement}

\textbf{Human participants and compensation}:
The dataset was translated and validated by native speakers with technical expertise. 
Translators were informed of the purpose of the work, how their contributions would be released, and that their output would be published under an open license; participation was voluntary, and translators
could withdraw at any point. Translators were voluntary contributors and were invited to be co-authors, provided they met academic standards for co-authorship. Contributors who took a substantive role in the design, curation or writing of the work are listed as authors. All contributions are documented in Appendix~\ref{app:contributions}.

\textbf{Data provenance and licensing}:
Multilingual GSM-Symbolic is derived from GSM8K \citep{cobbe2021gsm8k} using the GSM-Symbolic
template formulation \citep{Mirzadeh2024GSMSymbolicUT} and is released under an MIT license, consistent with the license of the source data.
The underlying items contain no personally identifiable information.

\textbf{Representation and the risk of misreading the benchmark}:
Broad language coverage can suggest that multilingual evaluation is solved. Grade-school
arithmetic is a narrow, culturally-light slice of language use, chosen precisely because it
holds the underlying task constant; strong performance here is not evidence of general
capability or usability in a language. We encourage reporting per-language results rather
than a single multilingual average.

\textbf{Contamination and benchmark lifetime}:
Publicly released benchmarks are likely to enter pretraining corpora. The symbolic template
design mitigates, but does not eliminate, this risk: millions of unseen surface
variants can be generated from each template, so memorisation of a specific instance does
not transfer, though memorisation at the template level remains possible. We report the
generation procedure and release the package for generating variants so that fresh, previously unpublished
variants can be drawn at evaluation time.

\textbf{Environmental Impact}:
All experiments were run on NVIDIA Blackwell B200 GPU-hours on the UCloud interactive HPC system, managed by the eScience Center at the University of Southern Denmark. It is estimated to use 84\% renewable energy.
We evaluate only open-weight models and release per-model, per-sample logs so that the analyses in this paper can be reproduced and extended without re-running inference. A total of 180 GPU hours were used during inference. Additionally various APIs were used for translation, editing etc..

\section*{AI use statement}
Translators were permitted to use LLM assistance, and all chose to do so, followed by manual review and correction by the translator.
None of the models used for translation assistance are among the models evaluated in this paper, so the evaluation isn't biased towards its own translations.
Each template records the model, prompting language and the annotator's self-reported proficiency, so that the provenance of every item is inspectable. LLM assistance was likewise used by the authors for code and manuscript editing and reviewing; all reported results, analyses and claims were produced and verified by the authors.

\section*{Reproducibility Statement}
To support reproducibility, we share the code\footnote{\url{https://github.com/centre-for-humanities-computing/multilingual-gsm-symbolic}} and dataset\footnote{\url{https://huggingface.co/datasets/danish-foundation-models/multilingual-gsm-symbolic}}. To enable others to add new languages, extend the dataset, update the templates, correct errors, etc., we distribute both the dataset and the code as versioned artefacts to ensure reproducibility and traceability. Along with the code, we share a test suite to prevent digressions. Along with the dataset, we share exact logs for each model on each sample to enable further analysis.

\textbf{Integrations}:
As this dataset is the first of its kind for many languages, we ensure it is accessible by integrating it with EuroEval \cite{smart2023scandeval, smart2025encoder}, the largest European evaluation suite that is actively maintained, and the popular evaluation framework inspect.ai \cite{aisi2024inspectai}.

\section{Acknowledgments}
    
\subsection{Funding}
Yevhen Kostiuk, Gianluca Barmina, Mike Zhang, Zafar Hussain, and Kenneth Enevoldsen are funded by the Danish Foundation Models project (4378-00001B). Kenneth Enevoldsen, Zafar Hussain and Simon Enni are funded by the Aage and Johanne Louis-Hansens Foundation (25-1-17733), and the Augustinus Foundation (2025-0299). Kenneth Enevoldsen is additionally funded by Danish National Research Foundation (DNRF193) and the European Union, Horizon Europe (101178170). Lukas Galke Poech acknowledges support from the Novo Nordisk Foundation (NNF25OC0103204). Elisa Bassignana is supported by a research grant (VIL59826) from VILLUM FONDEN. 

Part of the computation done for this project was performed on the UCloud interactive HPC system, which is managed by the eScience Center at the University of Southern Denmark

\subsection{Contributors}
We would additionally also like to thanks the Swedish contributors notably Taras Kucherenko for coodination along with KB-Lab for providing hours. We would also like to thank Antonio Tällberg Ilestad, Viktoria Lundborg, Andreas Hedberg, and Johan Bisse Mattsson for assisting in validation.

\bibliography{custom}
\bibliographystyle{iclr2025_conference}

\newpage
\appendix

\section*{Appendix}
\label{sec:appendix}

\etocsetlocaltop{part}  
\etocsettocstyle{\subsection*{Appendix contents}}{}
\localtableofcontents
\newpage

\section{Ablation and Robustness}
\subsection{Alternative GLMM Specifications}
\label{app:sec:model-ablations}

Examining Equation~\ref{eq:model} one might reasonably argue for additional effect such as a interaction between resource and typoligical distance or adding script as a feature. We examine alternative specifications in the following section.

We found that an interaction term between resource and distance was nonsignificant ($\beta_{\text{resource}\times\text{distance}}=0.02$, $p = .61$) and similar was found when adding script
($\beta_{script}=-0.03$, $p = .90$). Adding script this also not notably affect the effect of typological distance barely moves when it is added (-0.29 to -0.28), so the distance effect is not a writing-system effect in disguise.

Similarly, one could reasonably argue for an inclusion of a template-by-language random effect as the current model assume template difficulty is language independent. Adding the this term we see that template difficulty is not language-invariant --- the
fitted component has $\mathrm{sd} = 0.92$, larger than the model-by-language term ($0.61$)
and more than three times the language term ($0.27$) and including it improves AIC by roughly $10^5$. The term however, does not influence our conclusions it: every significant coefficient keeps its sign and its significance, standard errors inflate by a uniform factor of $1.05$, and $\mathrm{sd}(\text{language})$ is unchanged at $0.27$, so the language-level estimates are unaffected. Refitting both the baseline and the full model with the term gives 91\% of between-language and 16\% of model-by-language variance explained, against 89\% and 20\% without it. 
The most notable changes are a reduction is that $\text{typological distance}\times\text{size}$ goes from $0.017$ to $-0.000$, reinforcing that scale does nothing for typological distance, while $\text{resource}\times\text{reasoning}$ weakens from $p = .004$ to $p = .03$. We report the initial implementation to avoid selecting a specification after seeing the results. WE, however, encourage future development in direction to include this term.

We imagine that many further exploration on the model fit could be made and hope that our released code allow for such experimentation.

\subsection{Machine vs human-translation} \label{additional_languages}

\begin{figure}
\begin{minipage}{0.48\linewidth}
\centering
\includegraphics[width=\linewidth]{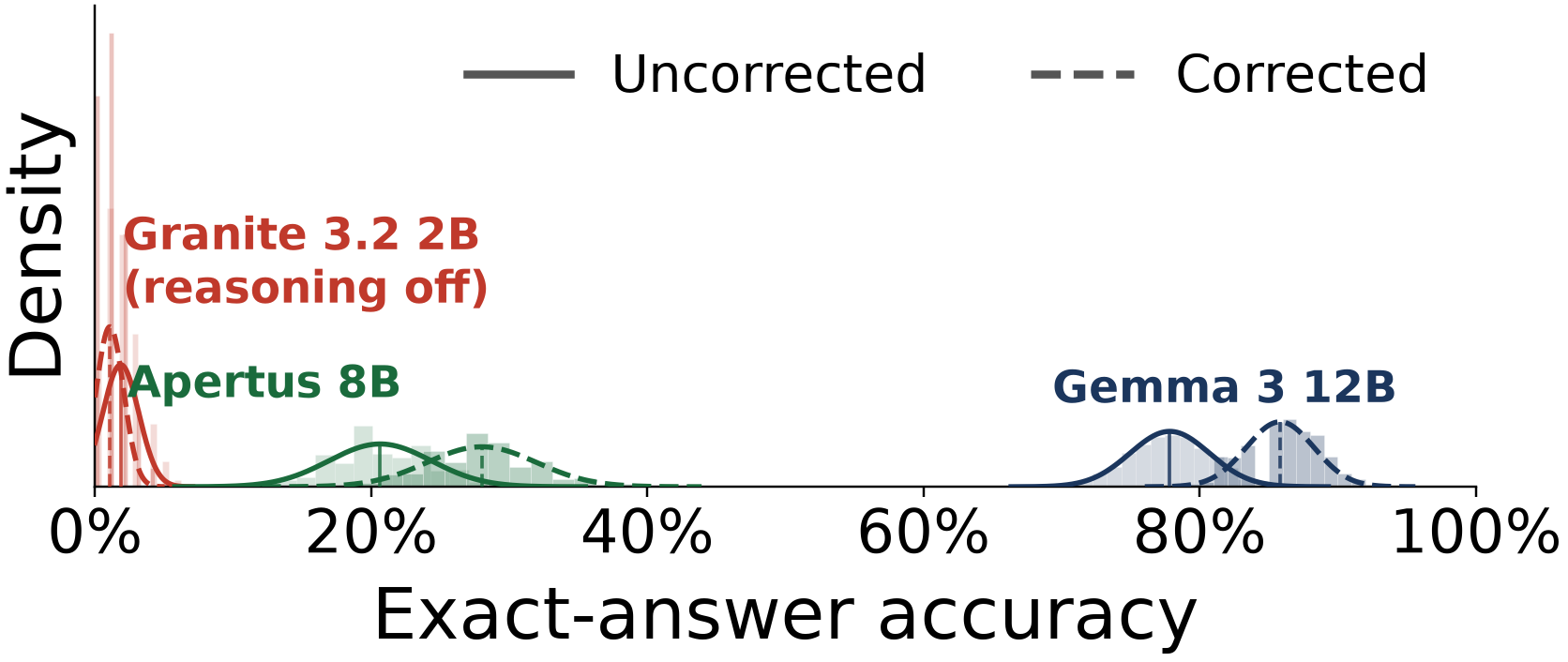}
\caption{\textbf{Icelandic validated vs unvalidated}: Accuracy on Icelandic before and after validation across three models. To inspect it across all model see Appendix~\ref{app:sec:icelandic-verification}}
\label{fig} 
\end{minipage}\hfill
\begin{minipage}{0.48\linewidth}
\centering
\includegraphics[width=\linewidth]{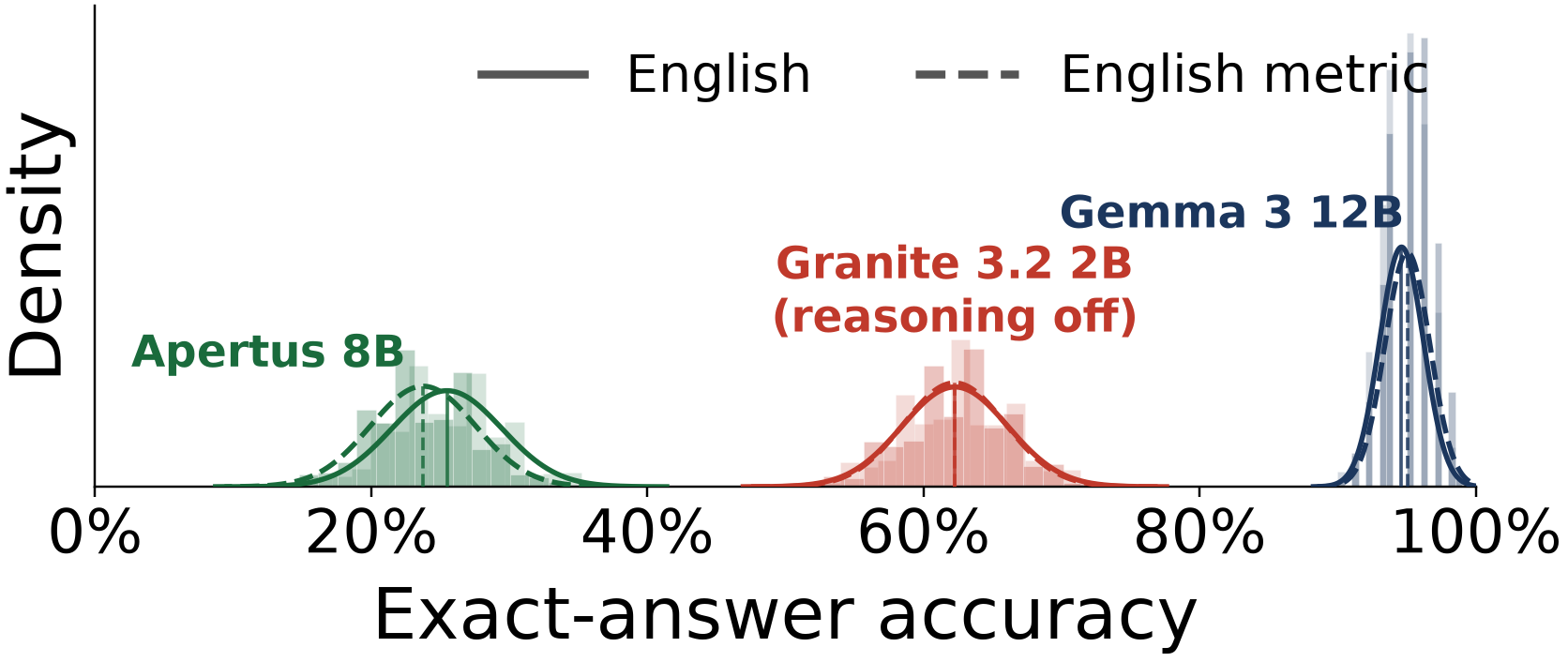}
\caption{\textbf{Imperial vs Metric}: Accuracy on English utilizing imperial and metric units. To inspect it across all model see Appendix~\ref{app:sec:metric-verification}}
\label{fig}
\end{minipage}
\end{figure}

While this study utilizes only human-verified templates, it is of interest if conclusions utilizing only the translated templates would have provided equivalent results. To test this we applied a TOST equivalence test with a margin of 1pp on a machine-translated and human-verified templates across the full set of models. For our language we choose Icelandic as it is low-resource with less than 400.000 native speakers. We found that the means scores were not  equivalent (p>0.999) within the margin. This motivates the decision to use only human-validated languages in the analysis. 
To allow for further validation we provide the initial machine-translation of 100 popular languages, but note that these should be used for evaluation with care. For the full list of models see Appendix~\ref{app:sec:model-ids}.

\subsection{English Imperial vs English Metric}
\label{app:sec:standard-v-metric}

To understand if multilingual performance gaps are due to non-English templates using metric units as opposed to the imperial units in English templates, we construct a metric version of our English dataset. We apply a TOST equivalence test with a margin of 1pp on the English templates using imperial units and English templates using metric units across the full set of models. We found that the mean scores were equivalent (p <0.01) within the margin. For the full list of models see Appendix~\ref{app:sec:model-ids}. 

This supports our approach of localizing units by providing evidence that performance gaps are not simply due to the use of metric rather than imperial units. Similarly there is no evidence to suggest that the use of metric eases model computation.

\subsection{Sample Size}
\label{app:sec:sample-size}

The benchmark comprises 100 templates, each instantiated 20 times. In this section we explore varying these parameters.

\begin{figure}
\centering
\includegraphics[width=\linewidth]{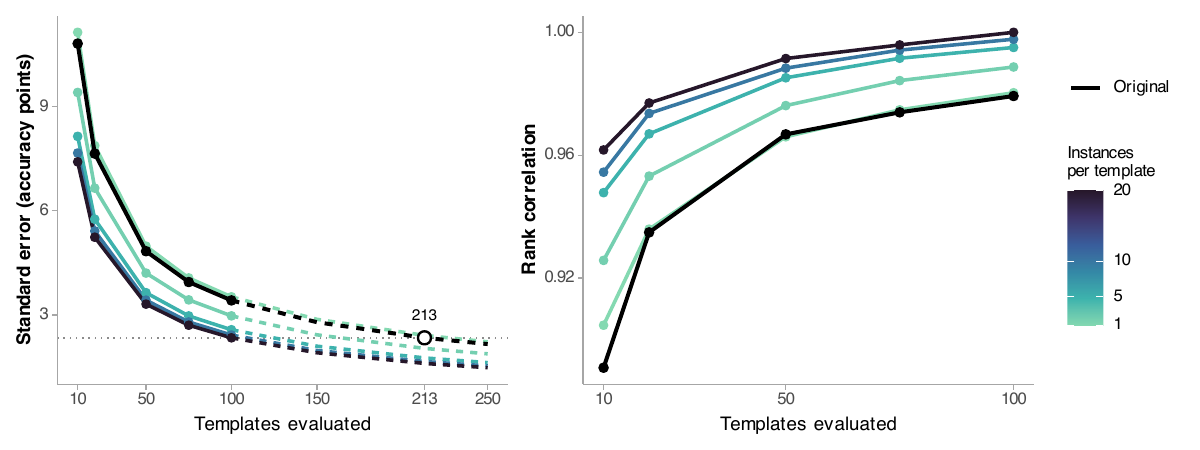}
\caption{\textbf{How much data is needed}. Left: the standard error of a single (model, language) accuracy at a given template budget, which falls as $1/\sqrt{k}$ and is still $2.29$ points at the full 100 templates. Right: the rank correlation between the model ordering at that budget and the ordering on all 100 templates, which reaches $0.96$ by 10 symbolic templates. Rankings settle long before the scores do, and the symbolic split is more precise than the original at every budget.}
\label{fig:sample-size}
\end{figure}

\textbf{How many templates?} We resample $k$ of the 100 templates and ask two questions of the result: whether the ordering of models is preserved, and how much sampling error remains in a model's reported accuracy. The first is a rank correlation against the full 100-template ordering; the second is the standard error of a (model, language) accuracy at a budget of $k$ templates, computed as the template-to-template standard deviation over $\sqrt{k}$. Both are shown in Figure~\ref{fig:sample-size} and in Table~\ref{tab:sample-size}.

\begin{table}
\centering
\begin{tabular}{lrrrr}
\toprule
& \multicolumn{2}{c}{Rank correlation} & \multicolumn{2}{c}{SE (accuracy points)} \\
\cmidrule(lr){2-3}\cmidrule(lr){4-5}
Templates & Original & Symbolic & Original & Symbolic \\
\midrule
10  & $0.914$ & $0.964$ & $10.60$ & $7.26$ \\
20  & $0.951$ & $0.977$ & $7.49$  & $5.13$ \\
50  & $0.983$ & $0.991$ & $4.74$  & $3.24$ \\
75  & $0.993$ & $0.997$ & $3.87$  & $2.65$ \\
100 & $1.000$ & $1.000$ & $3.35$  & $2.29$ \\
\bottomrule
\end{tabular}
\caption{Effect of the template budget on model ranking and on the precision of a
single (model, language) score. Rank correlation is against the full 100-template
ordering and reaches 1 there by construction; the standard errors do not.}
\label{tab:sample-size}
\end{table}

Model rankings are stable well below the full set: 10 symbolic templates already reproduce the full ordering at $\rho = 0.96$, and 50 at $0.99$.

Reported accuracies are a different matter. At the full 100 templates a single (model, language) score still carries a standard error of $2.29$ accuracy points, and because the error falls as $1/\sqrt{k}$, halving it would require roughly 400 templates.

\textbf{Do symbolic variants help?} At every budget the symbolic split is more precise by a factor of $1.46$ in
standard error. Examining the extrapolation in Figure~\ref{fig:sample-size} we see that it would take it would take about 213 original templates to meet the symbolic value at 100 templates.

\textbf{How many instances per template?} The instance budgets in Figure~\ref{fig:sample-size} converge with sharply
diminishing returns: at the full template set, going from 5 instances to 10 removes $6\%$ of the standard error and 10 to 20 only $3\%$, while no number of additional instances, however large, could remove more than a further $3.4\%$.
The reason is that the across-template variance splits as $\mathrm{Var}(\hat p) = \sigma^2_{\text{template}} + \mathbb{E}[q(1-q)]/n$, where only the second term falls with $n$. 
At $n = 20$ the template-level term is already $93\%$ of the total, so the $n \to \infty$ floor sits just below the precision we already have. The binding constraint is therefore the number of templates, not instances. This is a property of the current generator, which varies names, values and surface context; structural variation in the manner of \citet{Mirzadeh2024GSMSymbolicUT} would move variance into the instance term and make further instances worth generating.

\section{Contributions}

We list contributions according to the Contributor Roles Taxonomy (CRediT)\footnote{https://credit.niso.org/}

\textbf{Kenneth Enevoldsen}: Conceptualization, Data curation, Formal analysis, Funding acquisition, Investigation, Methodology, Project administration, Resources, Software, Supervision, Validation, Visualization, Writing – original draft, Writing – review \& editing

\textbf{Riley Herchert}: Data curation, Formal analysis, Investigation, Methodology, Software, Validation, Visualization, Writing – original draft, Writing – review \& editing

\textbf{Sofie Mosegaard}: Data curation, Investigation, Resources, Software, Writing – review \& editing

\textbf{Dan Sattrup Smart}: Data curation, Formal analysis, Methodology, Resources, Writing – review \& editing

\textbf{Simon Enni}: Data curation, Investigation, Resources, Software, Writing – review \& editing

\textbf{Isaac Chung}: Data curation, Resources, Writing – original draft, Writing – review \& editing

\textbf{Sofie Bruun}: Data curation, Resources, Writing – review \& editing

\textbf{Ayush Sunil Munot}: Data curation, Resources, Writing – review \& editing

\textbf{Max Müller-Eberstein}: Data curation, Resources, Writing – review \& editing

\textbf{Adnan El-Assadi}: Data curation, Resources, Writing – review \& editing

\textbf{Elisa Bassignana}: Data curation, Resources, Writing – review \& editing

\textbf{Gianluca Barmina}: Data curation, Resources, Writing – review \& editing

\textbf{Hafsteinn Einarsson}: Data curation, Resources, Writing – review \& editing

\textbf{Iben Nyholm Debess}: Data curation, Resources, Writing – review \& editing

\textbf{Linda Freienthal}: Data curation, Resources, Writing – review \& editing

\textbf{Lukas Galke Poech}: Data curation, Resources, Writing – review \& editing

\textbf{Mike Zhang}: Data curation, Resources, Writing – review \& editing

\textbf{Nicolas Legrand}: Data curation, Resources, Writing – review \& editing

\textbf{Vladimir Salnikov}: Data curation, Resources, Writing – review \& editing

\textbf{Yevhen Kostiuk}: Data curation, Resources, Writing – review \& editing

\textbf{Zafar Hussain}: Data curation, Resources, Writing – review \& editing

\textbf{Sagandeep Kaur}: Data curation, Resources, Writing – review \& editing

\textbf{Agnes Toftgård}: Data curation, Resources, Writing – review \& editing

\textbf{Marie Mattson}: Data curation, Resources, Writing – review \& editing

\textbf{Faton Rekathati}: Data curation, Resources, Writing – review \& editing

\textbf{Kevin Kelly}: Data curation, Resources, Writing – review \& editing

\textbf{Taras Kucherenko}: Project administration

\textbf{Kristoffer Nielbo}: Funding, Writing – review \& editing




\section{GLMM Estimates}

\label{app:sec:glmm-estimates}

Table~\ref{tab:model-fit} reports the fitted estimates of Equation~\ref{eq:model} and Table~\ref{tab:standardisation} the constants needed to return them to raw units. 
The model was fitted using in \texttt{lme4} 1.1.37 \citep{bates2015fitting} under R 4.5.1, using the \texttt{bobyqa} optimiser.
It converged without warnings on $N = 69{,}000$ cells of 20 instances each ($\mathrm{AIC} = 422{,}410$, $\mathrm{BIC} = 422{,}584$, $\log L = -211{,}186$).

Fixed effects are in log-odds per standard deviation of the predictor and are therefore comparable across predictors; \texttt{reasoning} is a two-level factor with \texttt{off} as the reference.
Because resource level and model size are log-transformed before standardising, one standard deviation of each is a fixed number of doublings.

\begin{table}
\centering
\small
\begin{tabular}{lrrrr}
\toprule
 & $\hat\beta$ & SE & $z$ & $p$ \\
\midrule
\multicolumn{5}{l}{\emph{Language features}} \\
\quad Resource level               & $0.773$  & $0.074$ & $10.41$  & $<2\times10^{-16}$ \\
\quad Typological distance         & $-0.253$ & $0.074$ & $-3.41$  & $0.0006$ \\
\quad Relative fertility (within)  & $-0.041$ & $0.032$ & $-1.30$  & $0.195$ \\
\addlinespace
\multicolumn{5}{l}{\emph{Model features}} \\
\quad Model size                   & $1.768$  & $0.133$ & $13.34$  & $<2\times10^{-16}$ \\
\quad Reasoning (on)               & $0.669$  & $0.085$ & $7.90$   & $2.9\times10^{-15}$ \\
\addlinespace
\multicolumn{5}{l}{\emph{Feature $\times$ size}} \\
\quad Resource $\times$ size       & $-0.265$ & $0.024$ & $-11.01$ & $<2\times10^{-16}$ \\
\quad Distance $\times$ size       & $0.024$  & $0.024$ & $1.00$   & $0.319$ \\
\quad Fertility $\times$ size      & $-0.047$ & $0.030$ & $-1.55$  & $0.121$ \\
\addlinespace
\multicolumn{5}{l}{\emph{Feature $\times$ reasoning}} \\
\quad Resource $\times$ reasoning  & $-0.200$ & $0.053$ & $-3.76$  & $0.0002$ \\
\quad Distance $\times$ reasoning  & $0.115$  & $0.053$ & $2.17$   & $0.030$ \\
\quad Fertility $\times$ reasoning & $0.106$  & $0.080$ & $1.33$   & $0.183$ \\
\quad Size $\times$ reasoning      & $-0.344$ & $0.093$ & $-3.71$  & $0.0002$ \\
\addlinespace
\quad Intercept                    & $0.015$  & $0.446$ & $0.03$   & $0.973$ \\
\midrule
\multicolumn{5}{l}{\emph{Random effects}\quad (levels, $\mathrm{sd}$)} \\
\quad Family                       & \multicolumn{4}{l}{9\quad $1.307$} \\
\quad Template                     & \multicolumn{4}{l}{100\quad $1.209$} \\
\quad Base model within family     & \multicolumn{4}{l}{35\quad $0.727$} \\
\quad Model $\times$ language      & \multicolumn{4}{l}{690\quad $0.578$} \\
\quad Language                     & \multicolumn{4}{l}{15\quad $0.250$} \\
\quad Model                        & \multicolumn{4}{l}{46\quad $0.126$} \\
\bottomrule
\end{tabular}
\caption{\textbf{Fitted estimates} of Equation~\ref{eq:model}. Fixed effects in log-odds per
standard deviation of the predictor; random effects ordered by magnitude, with the
number of levels of each grouping factor.}
\label{tab:model-fit}
\end{table}

\begin{table}
\centering
\begin{tabular}{llrr}
\toprule
Term & Raw quantity & Mean & SD \\
\midrule
Resource level       & $\log_{10}$ Common Crawl pages             & $7.396$ & $0.862$ \\
Typological distance & URIEL \texttt{syntax\_knn} cosine distance & $0.219$ & $0.139$ \\
Relative fertility   & within-language deviation                  & $0.000$ & $0.472$ \\
Model size           & $\log_{2}$ parameters (billions)           & $2.791$ & $1.865$ \\
\bottomrule
\end{tabular}
\caption{\textbf{Standardisation constants}}
\label{tab:standardisation}
\end{table}
\section{Synthetic vs Original Variants}
\label{app:synth-vs-orig}

\begin{figure}
    \centering
    \includegraphics[width=0.60\linewidth]{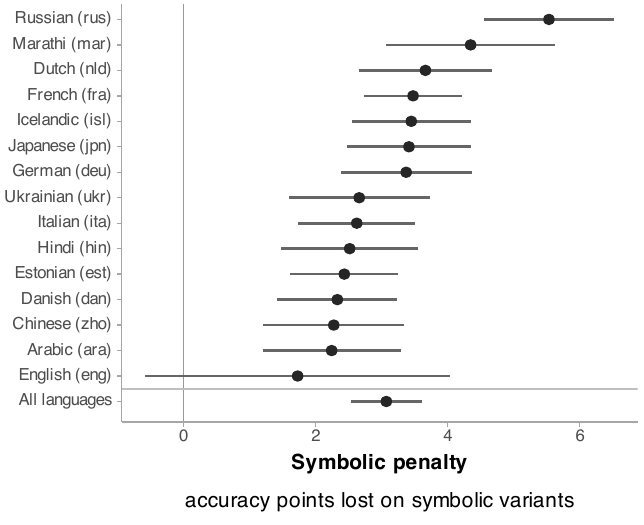}
    \caption{\textbf{Symbolic penalty}: Mean penalty across languages along with a 95\% CI.}
    \label{fig:symbolic-penalty}
\end{figure}

Figure~\ref{fig:symbolic-penalty} shows the mean penalty across languages measured as the difference in the synthetic sets and the originals.

\begin{figure}
        \centering
        \includegraphics[width=0.75\linewidth]{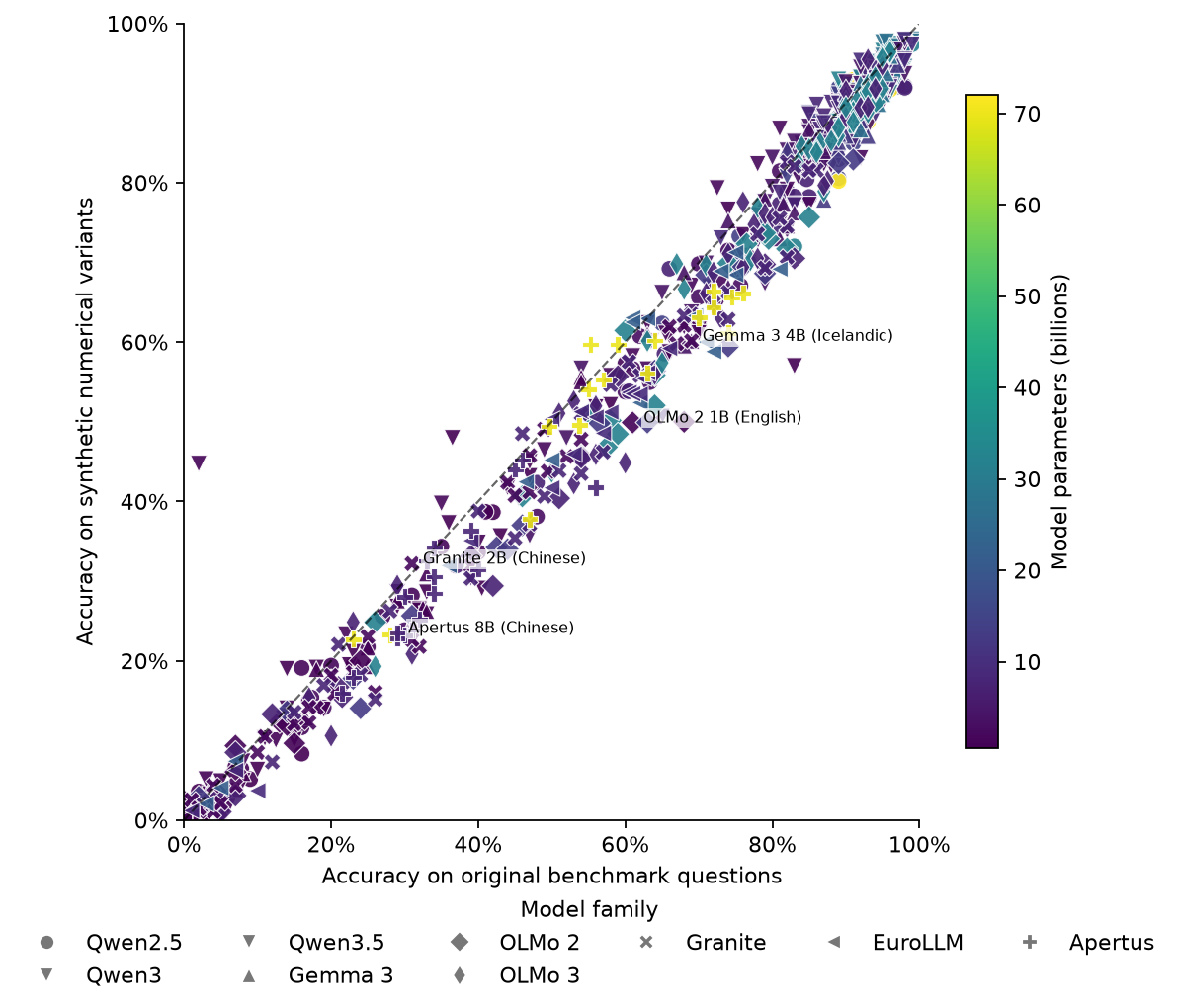}
        \caption{\textbf{Original set accuracy vs synthetic set accuracy}: Each dot denote a (model, language) pair. Accuracy is the mean accuracy on either the synthetic or original set.}
        \label{app:fig:original-vs-synthetic}
\end{figure}

Figure~\ref{app:fig:original-vs-synthetic} complement the main analysis and shows performance discrepancy on the original questions against synthetic variants.

\subsection{Modeling the Symbolic Penalty}
\label{app:synth-vs-orig}
Whether symbolic variants are harder, and whether the penalty differs across languages, is
estimated over both splits jointly rather than from a paired test on per-cell differences.
A paired test would treat the 48 models evaluated in a language as that many independent
observations of that language's fragility, inflating the precision of any per-language
claim for the same reason the language term is needed in Equation~\ref{eq:model}.

Let $z_{s} = 1$ if an instance comes from the symbolic split and $0$ if it comes from the
original questions. Extending Equation~\ref{eq:model},

\begin{equation}
\begin{aligned}
y_{mltsi} &\sim \mathrm{Bernoulli}(\pi_{mltsi}), \\[2pt]
\mathrm{logit}(\pi_{mltsi}) &= \mathbf{x}_{l}^{\top}\boldsymbol{\beta}
  \;+\; z_{s}\!\left(\delta + \mathbf{x}_{l}^{\top}\boldsymbol{\delta}\right)
  \;+\; \sum_{g\in\mathcal{G}} u^{(g)}_{j_g(m,l,t)}
  \;+\; z_{s}\, v_{l}, \\[4pt]
\begin{pmatrix} u^{(\text{language})}_{l} \\[2pt] v_{l} \end{pmatrix}
  &\overset{\text{iid}}{\sim} \mathcal{N}\!\left(\mathbf{0}, \boldsymbol{\Sigma}\right)
\end{aligned}
\label{eq:symbolic}
\end{equation}

where $\mathbf{x}_{l}$ holds the two language features and $\mathcal{G}$ is as in
Equation~\ref{eq:model}. Model size and reasoning need no fixed terms here: the model,
base-model and family random effects already absorb capability, and the question is only
how the split effect behaves.

Each new term answers one question. The scalar $\delta$ is the average symbolic penalty.
The vector $\boldsymbol{\delta}$ asks whether that penalty is predicted by the language
features. The language-specific slope $v_{l}$ measures how much it varies across languages
beyond those features, so $\tau^2 = \mathrm{Var}(v_l)$ is the quantity of interest: a
likelihood-ratio test of $\tau = 0$ against a model with $v_{l}$ removed tests whether
fragility differs across languages at all. Correlating $v_{l}$ with the language intercept
lets harder languages be more or less fragile rather than assuming the two unrelated.

We fit over all $153{,}600$ (model $\times$ language $\times$ template $\times$ split)
cells. The splits differ in instances per cell --- one per template for the original
questions, twenty for the symbolic variants --- which the binomial response accommodates
directly, the original arm simply carrying less information per cell. We obtain
$\hat{\delta} = -0.297$ ($p < 10^{-15}$) and $\hat{\tau} = 0.059$ log-odds
($\chi^{2} = 18.4$, $\mathrm{df} = 2$, $p < 0.001$), with both elements of
$\boldsymbol{\delta}$ null (resource $p = 0.64$, typological distance $p = 0.85$). 
\section{Reasoning vs Non Reasoning}

\subsection{Capability Transfer Gap}
\begin{figure}[H]
    \centering
    \includegraphics[width=.8\linewidth]{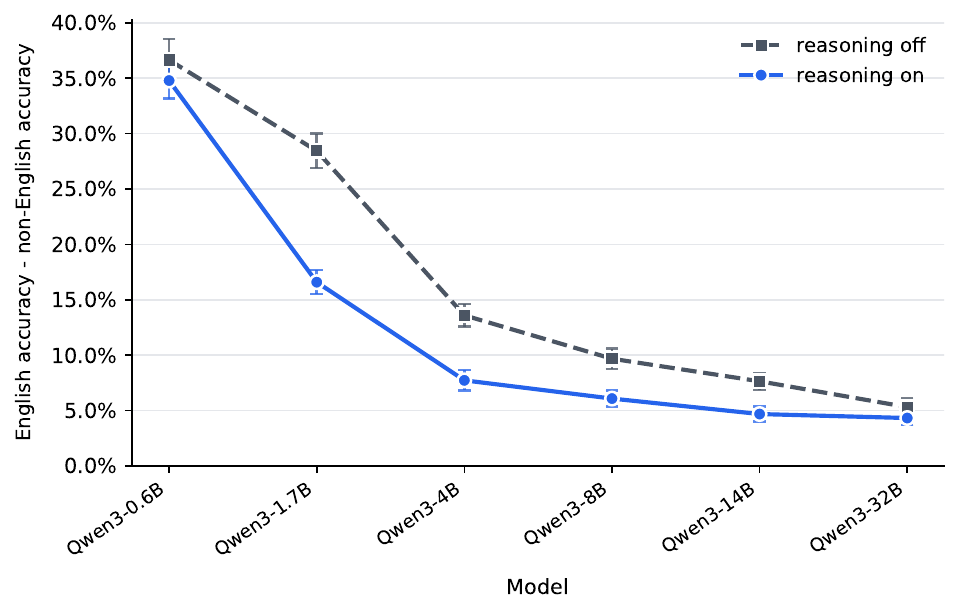}
    \caption{\textbf{Transfer gap with reasoning enabled vs disabled} across the Qwen models.}
    \label{app:fig:transfer-gap}
\end{figure}

Figure \ref{app:fig:transfer-gap} shows the transfer gap for Qwen3 across size with and without reasoning enabled. Across all model sizes, enabling reasoning reduced the transfer gap, consistent with the conclusion of our model.

\subsection{Under compute constraint}

\begin{figure}
    \centering
    \includegraphics[width=\linewidth]{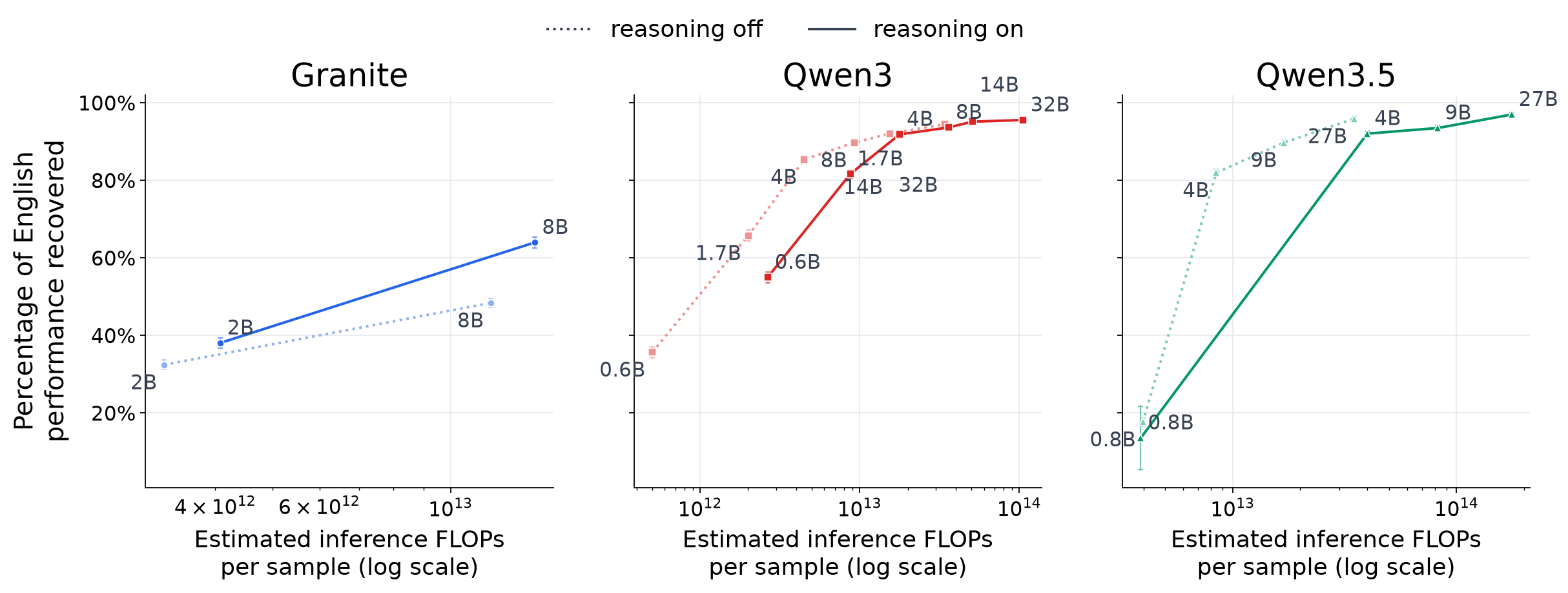}
    \caption{\textbf{Reasoning on/off at a given compute budget}: We include only models that allow enabling or disabling reasoning.}
    \label{app:fig:reasoning-at-a-budget}
\end{figure}

In Figure \ref{app:fig:reasoning-at-a-budget} we compare how the relative gap between English and non-English language changes across compute constraint at inference time. We examine this only in models that allow reasoning to be toggled. We show that it can often be favorable to prefer larger models over reasoning models if one wishes to reduce the language gap, but that enabling reasoning allows for the largest possible absolute reduction. 

\section{Leave-one-language-out Prediction}
\label{app:sec:prediction}

Each of the languages is held out in turn and Equation~\ref{eq:model} refitted on the remaining. 
Standardisation is recomputed on each training fold and applied to the held-out language, so its own mean and spread never enter the fit.
Prediction uses the fixed effects together with the family, model, variant and template random effects, all of which are estimated from the other languages; 
the language and model-by-language terms are set to zero as they can't be known for a language that has not been evaluated.

The naïve baseline is computed by estimating the model's mean accuracy over the training set.

\begin{figure}
    \centering
    \includegraphics[width=1\linewidth]{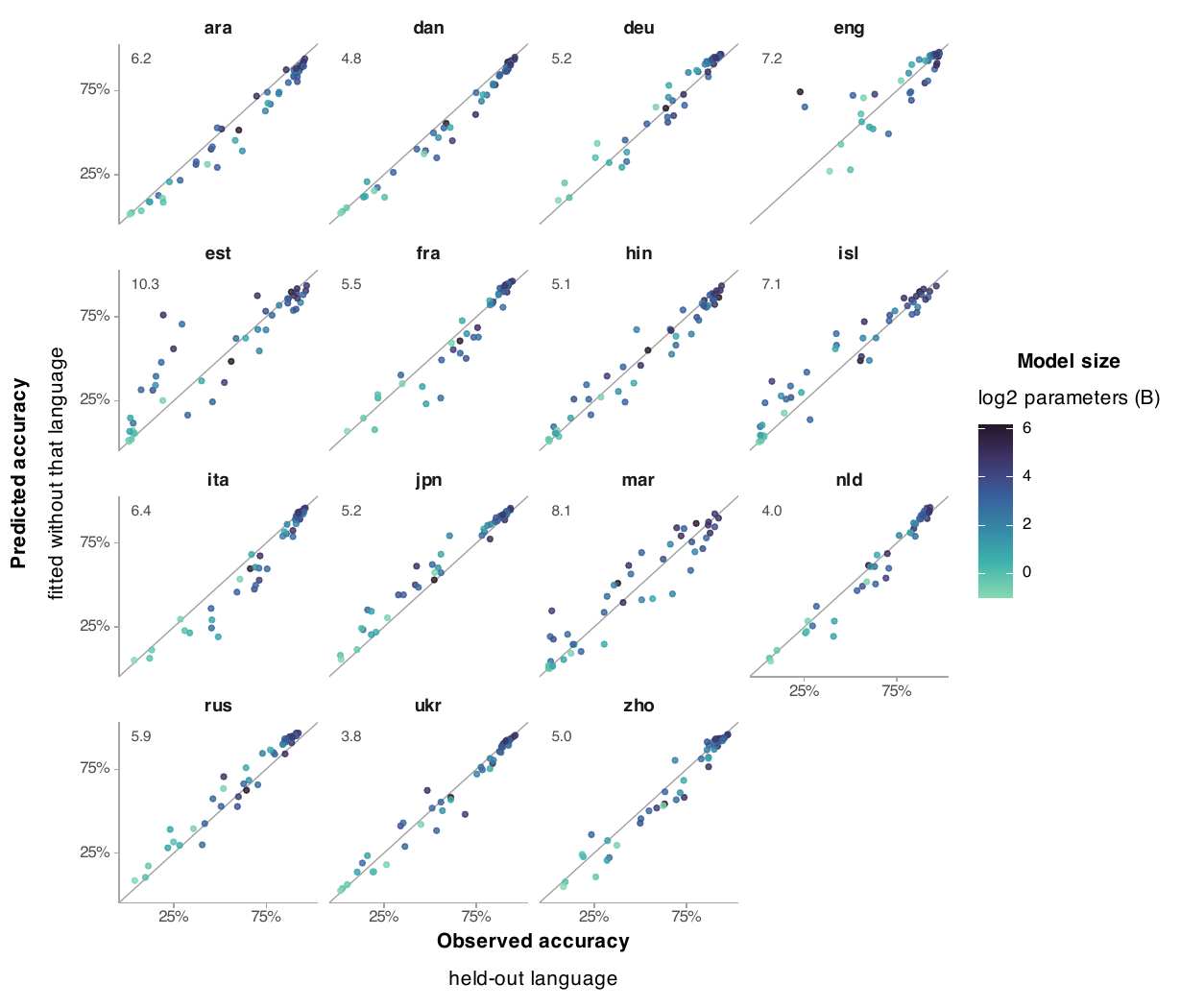}
    \caption{\textbf{Forecasting model performance} on a given language, showing (model, language) pairs grouped by language.}
    \label{fig:predict-by-language}
\end{figure}

Figure~\ref{fig:predict-by-language} show Figure~\ref{fig:predict-model} decomposed by language.

For the experiment utilizing existing templates we utilize 20 samples from each template and explore across $k\in\{0, 1, 2, 5, 10, 20\}$ templates, with the $k$ templates sampled at random and each budget repeated over three independent draws. Results are shown in Table~\ref{app:tab:predict-unseen-at-k}.

\begin{table}[t]
\centering
\begin{tabular}{lcc}
\toprule
$k$ & Model in language & Language mean \\
\midrule
0  & $5.98$ $[5.22, 6.86]$ & $3.17$ $[2.32, 4.06]$ \\
1  & $8.38$ $[7.15, 9.69]$ & $6.66$ $[5.05, 8.41]$ \\
2  & $7.02$ $[6.26, 7.94]$ & $4.35$ $[3.26, 5.38]$ \\
5  & $5.63$ $[4.88, 6.48]$ & $3.29$ $[2.26, 4.33]$ \\
10 & $4.19$ $[3.75, 4.69]$ & $2.38$ $[1.78, 3.08]$ \\
20 & $2.73$ $[2.50, 2.96]$ & $1.44$ $[1.11, 1.78]$ \\
\bottomrule
\end{tabular}
\caption{\textbf{Effect of including a few templates} when predicting on an otherwise
held-out language. Mean absolute error in accuracy points over three draws per language,
with 95\% bootstrap intervals over the 15 languages.}
\label{app:tab:predict-unseen-at-k}
\end{table}

\section{Resource cost in model size}

\begin{figure}
    \centering
    \includegraphics[width=0.75\linewidth]{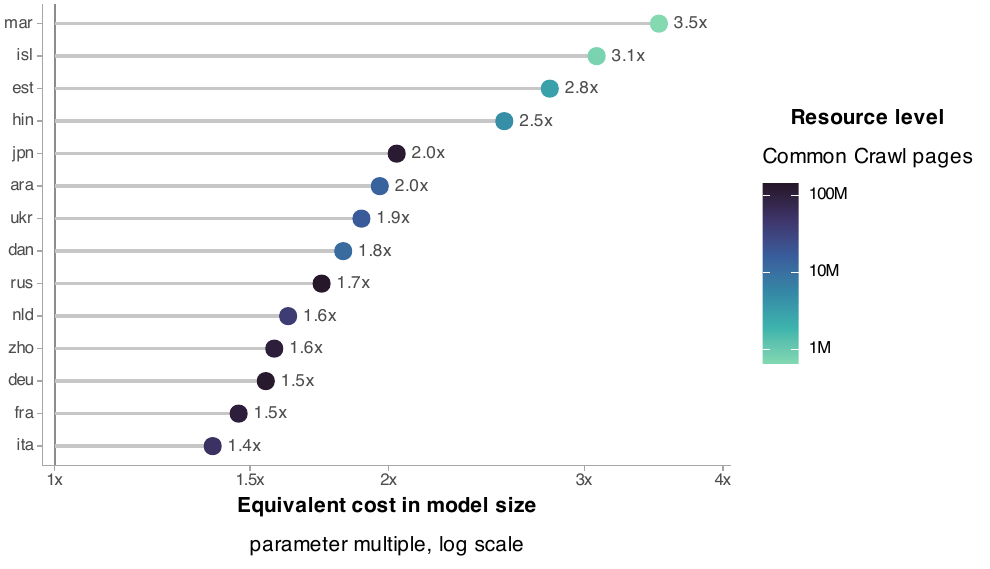}
    \caption{What does it cost to be a low-resource language in parameters?}
    \label{app:fig:resource-in-model-cost}
\end{figure}

Given the model specified in Eq.~\ref{eq:model} and the log scale of the variables it is possible examine how much it costs to be a low-resource language in terms of model size. 
This is done by dividing by their standard deviations such that the two can be compared, we see a .99 log-odds per doubling of parameters and .25 log-odds per doubling of Common Crawl pages, therefore doubling the model size is worth about four doublings of language data (~15x the data).
We examine the consequences of this pr. language in Figure~\ref{app:fig:resource-in-model-cost}. We see that Marathi has to use a models 3 times larger to obtain equivalent performance estimates.

\section{Feature design space}
\label{app:sec:feature-design-space}

\begin{figure}
    \centering
    \includegraphics[width=0.75\linewidth]{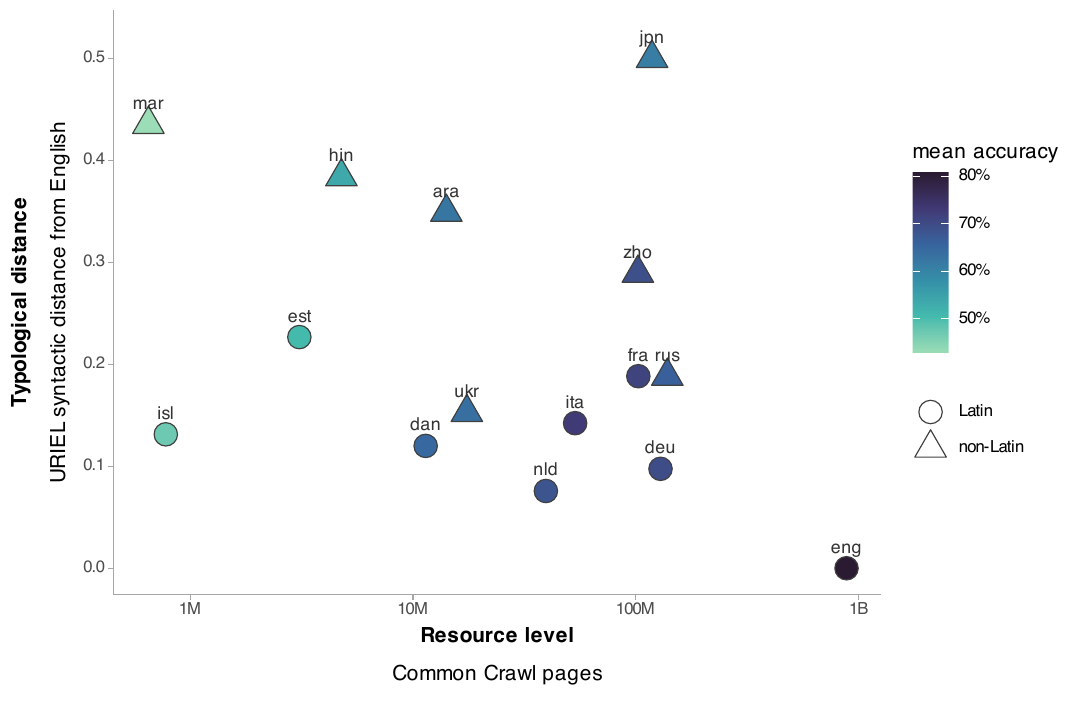}
    \caption{\textbf{Feature design space}: We cover both high and low resource languages at various typological distances from English.}
    \label{app:fig:design-space}
\end{figure}

In Figure~\ref{app:fig:design-space} we show that our selected languages cover the feature design space appropriately, covering both the axis of high and low resource languages as well as the typological distance axis.

\section{Solution gap analysis}
\label{app:ladder}

Understanding whether incorrect responses arise from failures to correctly extract information in the prompt or from errors correctly applying reasoning procedure is important for understanding capability transfer gaps. If the model fail primarily due to language extraction then language understanding may be the limiting factor while if it correctly extract numbers but fail at later stages this reflects instability in reasoning across languages.

\subsection{Outcome extraction}
As our templates carry GSM-style \texttt{<<lhs=rhs>>} calculator annotations we can extract operand reads (\texttt{lhs}), and intermediate computations (\texttt{rhs}) . We then count how many of these gold values occur anywhere in the model's response. Final answers are assessed separately.

The extractor normalizes Unicode characters, converting forms such as full-width digits to standard equivalents. Digit-based numbers written using Devanagari and Arabic-Indic numerals are converted to numeric values. Fractions, including  Japanese forms such as, \texttt{\begin{CJK}{UTF8}{gbsn}分の\end{CJK}}, are converted to decimal values. Comma and decimal usage is localized: \texttt{dan}, \texttt{deu}, \texttt{fao}, \texttt{fra}, \texttt{isl}, \texttt{ita}, \texttt{nld}, \texttt{nob}, and \texttt{por} use periods as thousands separators and commas as decimal points while other languages use periods as decimal separators and commas for grouping. Numbers written as words are not extracted.

The mean operand count per item ranges 5.31-5.45 across the 16 languages and mean intermediate
count 3.09-3.17. Only 2.1\% of the total (1{,}584 of
76{,}800) carry no annotations and are evenly distributed across languages. These are excluded when fitting the intermediate model.

\subsection{GLMM Estimates} 

\begin{figure}
    \centering
    \includegraphics[width=0.75\linewidth]{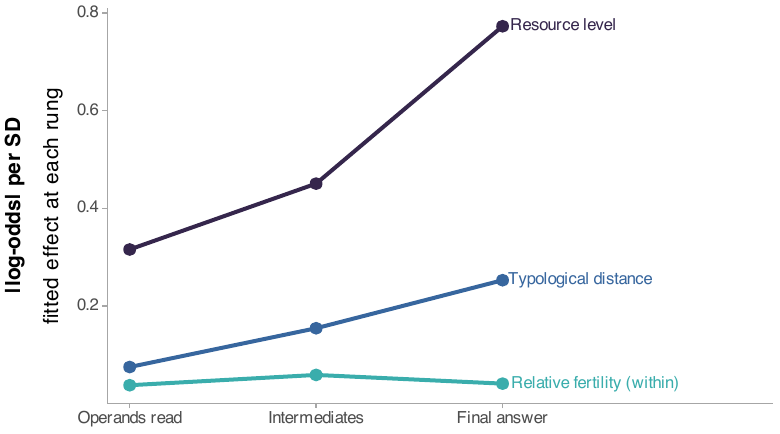}
    \caption{Fitted effects at each stage.}
    \label{app:fig:ladder}
\end{figure}

Each stage is fitted with the specification of Equation~\ref{eq:model} excluding reasoning and it related random intercepts controlling for model variants (with and without reasoning). The resulting estimated effects can be observed in Table~\ref{app:tab:ladder} and visualized in Figure~\ref{app:fig:ladder}.

\begin{table}[h]
\centering
\begin{tabular}{lrrr}
\toprule
 & Operand reads & Intermediates & Final answer \\
\midrule
Resource level              & $0.299^{***}$  & $0.420^{***}$  & $0.712^{***}$ \\
Typological distance        & $-0.091^{*}$   & $-0.172^{***}$ & $-0.288^{***}$ \\
Relative fertility (within) & $-0.035$       & $-0.067^{***}$ & $-0.071^{*}$ \\
Model size                  & $0.493^{***}$  & $0.934^{***}$  & $1.881^{***}$ \\
Resource $\times$ size      & $-0.070^{***}$ & $-0.120^{***}$ & $-0.246^{***}$ \\
\midrule
$\mathrm{sd}(\text{language})$       & 0.136 & 0.163 & 0.272 \\
$\mathrm{sd}(\text{model:language})$ & 0.390 & 0.390 & 0.580 \\
$\mathrm{sd}(\text{family})$         & 0.001 & 0.541 & 1.328 \\
\bottomrule
\end{tabular}
\caption{Fixed effects in log-odds per SD and variance components at each stage.
$^{*}p<0.05$, $^{**}p<0.01$, $^{***}p<0.001$.}
\label{app:tab:ladder}
\end{table}

\subsection{Is the size effect a ceiling artifact?}
\label{app:sec:saturation}

A negative $\text{resource} \times \text{size}$ interaction is what one would obtain mechanically if high-resource languages saturate while low-resource ones retain headroom.
Across our data 25\% of (model, language) cells exceed 90\% accuracy, so the concern is real, and because the interaction is offered as an actionable lever it should be shown not to depend on the ceiling.

We refit Equation~\ref{eq:model} on subsets that remove saturation. The primary restriction keeps only the 20 models scoring below 90\% in English; no cell in that subset exceeds 90\%, while the full 0.5--70B range of model sizes is retained (sd of $\log_2$ parameters 1.95, against 1.89 in the full data), so scale remains identified over the same span. We also
report a laxer threshold, and --- for completeness --- the cell-level filter, noting that dropping observations on the basis of their own outcome biases the estimate downward and so cannot be the primary evidence.

\begin{table}[h]
\centering
\begin{tabular}{lrrrr}
\toprule
 & Full & English $<95\%$ & English $<90\%$ & Cells $\leq 90\%$ \\
Models & 46 & 29 & 20 & 45 \\
Cells above 90\% & 171 & 12 & 0 & 0 \\
\midrule
Resource $\times$ size   & $-0.265^{***}$ & $-0.205^{***}$ & $-0.188^{***}$ & $-0.245^{***}$ \\
Distance $\times$ size   & $0.024$        & $0.005$        & $-0.012$       & $0.017$ \\
\addlinespace
Resource level           & $0.773^{***}$  & $0.941^{***}$  & $1.030^{***}$  & $0.853^{***}$ \\
Typological distance     & $-0.253^{***}$ & $-0.277^{***}$ & $-0.311^{***}$ & $-0.259^{**}$ \\
Model size               & $1.768^{***}$  & $1.837^{***}$  & $1.319^{**}$   & $1.779^{***}$ \\
\bottomrule
\end{tabular}
\caption{The size interaction under progressively stricter removal of saturated cells.
$^{*}p<0.05$, $^{**}p<0.01$, $^{***}p<0.001$.}
\label{tab:saturation}
\end{table}

The interaction survives. With saturation eliminated entirely it is $-0.188$
($p = 1.4\times10^{-5}$), and the attenuation across the four columns is orderly:
$-0.265$, $-0.245$, $-0.205$, $-0.188$ as the ceiling is progressively removed. We read
this as roughly a quarter to a third of the full-sample estimate being ceiling-inflated,
with the remainder a genuine effect. Equally, $\text{distance} \times \text{size}$ is null
in every specification, so the asymmetry between the two features --- the finding the
section rests on --- does not depend on the ceiling either.

One caveat bounds what this subset can test. Reasoning-capable models are
disproportionately the strong ones, so the 90\% restriction leaves only 3 base models run
both ways, against 11 in the full data. The reasoning interactions are therefore not
identified in this subset and we do not interpret them here; their estimates move toward
zero, but with three paired models that is an absence of evidence rather than evidence of
absence.

\section{Model Identifiers and Revisions}
\label{app:sec:model-ids}

\begin{table}[H]
\centering
\caption{Models repository identifiers on Hugging Face and revision commit hashes used in the evaluation.}
\label{tab:model-ids}

\footnotesize
\setlength{\tabcolsep}{4pt}
\renewcommand{\arraystretch}{1.02}

\begin{tabularx}{\textwidth}{
    @{}
    >{\raggedright\arraybackslash}X l
    @{\hspace{1.5em}}
    >{\raggedright\arraybackslash}X l
    @{}
}
\toprule
\textbf{Model ID} & \textbf{Rev} & \textbf{Model ID} & \textbf{Rev} \\
\midrule

\multicolumn{4}{l}{\textbf{Qwen2.5 Instruct}} \\[0.15em]
\nolinkurl{Qwen/Qwen2.5-0.5B-Instruct} & \texttt{7ae5576} & \nolinkurl{Qwen/Qwen2.5-1.5B-Instruct} & \texttt{989aa79} \\
\nolinkurl{Qwen/Qwen2.5-3B-Instruct}   & \texttt{aa8e725} & \nolinkurl{Qwen/Qwen2.5-7B-Instruct}   & \texttt{a09a354} \\
\nolinkurl{Qwen/Qwen2.5-14B-Instruct}  & \texttt{cf98f3b} & \nolinkurl{Qwen/Qwen2.5-32B-Instruct}  & \texttt{5ede1c9} \\
\nolinkurl{Qwen/Qwen2.5-72B-Instruct}  & \texttt{495f393} &                                         &                   \\[0.4em]

\multicolumn{4}{l}{\textbf{Qwen3}} \\[0.15em]
\nolinkurl{Qwen/Qwen3-0.6B}            & \texttt{c1899de} & \nolinkurl{Qwen/Qwen3-1.7B}            & \texttt{70d244c} \\
\nolinkurl{Qwen/Qwen3-4B}              & \texttt{1cfa9a7} & \nolinkurl{Qwen/Qwen3-8B}              & \texttt{b968826} \\
\nolinkurl{Qwen/Qwen3-14B}             & \texttt{40c0698} & \nolinkurl{Qwen/Qwen3-32B}             & \texttt{9216db5} \\[0.4em]

\multicolumn{4}{l}{\textbf{Qwen3.5}} \\[0.15em]
\nolinkurl{Qwen/Qwen3.5-0.8B}          & \texttt{2fc0636} & \nolinkurl{Qwen/Qwen3.5-4B}            & \texttt{851bf6e} \\
\nolinkurl{Qwen/Qwen3.5-9B}            & \texttt{c202236} & \nolinkurl{Qwen/Qwen3.5-27B}           & \texttt{fc05dae} \\[0.4em]

\multicolumn{4}{l}{\textbf{OLMo 2 Instruct}} \\[0.15em]
\nolinkurl{allenai/OLMo-2-0425-1B-Instruct}  & \texttt{48d788e} & \nolinkurl{allenai/OLMo-2-1124-7B-Instruct}  & \texttt{470b1fb} \\
\nolinkurl{allenai/OLMo-2-1124-13B-Instruct} & \texttt{3a5c85b} & \nolinkurl{allenai/OLMo-2-0325-32B-Instruct} & \texttt{b960243} \\[0.4em]

\multicolumn{4}{l}{\textbf{OLMo 3}} \\[0.15em]
\nolinkurl{allenai/Olmo-3-7B-Think}       & \texttt{d97e442} & \nolinkurl{allenai/Olmo-3-7B-Instruct}    & \texttt{6e5971d} \\
\nolinkurl{allenai/Olmo-3-32B-Think}      & \texttt{f2edda1} & \nolinkurl{allenai/Olmo-3.1-32B-Instruct} & \texttt{ac0587e} \\[0.4em]

\multicolumn{4}{l}{\textbf{Granite 3.2}} \\[0.15em]
\nolinkurl{ibm-granite/granite-3.2-2b-instruct} & \texttt{641593c} & \nolinkurl{ibm-granite/granite-3.2-8b-instruct} & \texttt{610d8c6} \\[0.4em]

\multicolumn{4}{l}{\textbf{Gemma 3 IT}} \\[0.15em]
\nolinkurl{google/gemma-3-1b-it}  & \texttt{dcc83ea} & \nolinkurl{google/gemma-3-4b-it}  & \texttt{093f9f3} \\
\nolinkurl{google/gemma-3-12b-it} & \texttt{96b6f1e} & \nolinkurl{google/gemma-3-27b-it} & \texttt{005ad34} \\[0.4em]

\multicolumn{4}{l}{\textbf{Apertus}} \\[0.15em]
\nolinkurl{swiss-ai/Apertus-8B-Instruct-2509} & \texttt{b946d40} & \nolinkurl{swiss-ai/Apertus-70B-Instruct-2509} & \texttt{0f27676} \\[0.4em]

\multicolumn{4}{l}{\textbf{OpenEuroLLM}} \\[0.15em]
\nolinkurl{utter-project/EuroLLM-1.7B-Instruct}    & \texttt{a25c7fa} & \nolinkurl{utter-project/EuroLLM-9B-Instruct-2512} & \texttt{def8245} \\
\nolinkurl{utter-project/EuroLLM-22B-Instruct-2512} & \texttt{8c6ed5c} &                                                    &                  \\

\bottomrule
\end{tabularx}
\end{table}
\newpage
\section{Language overview}

\subsection{Included languages}

In Table \ref{app:tab:langauge-overview} we include an overview of included languages, including both the verified and translated variants. In the Human verified column, \checkmark\ indicates completed verification, while (\checkmark) denotes languages in the process of validation. The analysis in the paper utilizes only the verified subsets; languages marked with parenthesized check marks are not included in the analysis.

\begin{table}
\centering

\scriptsize
\setlength{\tabcolsep}{2.2pt}
\renewcommand{\arraystretch}{0.88}
\begin{tabularx}{\textwidth}{
  @{}
  >{\raggedright\arraybackslash}X c c c
  @{\hspace{0.8em}}
  >{\raggedright\arraybackslash}X c c c
  @{}
}
\toprule
\textbf{Language} &
\makecell{\textbf{Human}\\\textbf{verified}} &
\textbf{EU} &
\makecell{\textbf{Ethno-}\\\textbf{logue}} &
\textbf{Language} &
\makecell{\textbf{Human}\\\textbf{verified}} &
\textbf{EU} &
\makecell{\textbf{Ethno-}\\\textbf{logue}} \\
\midrule

Afrikaans & & & \checkmark &
Lithuanian & & \checkmark & \\

Amharic & & & \checkmark &
Magahi & & & \checkmark \\

Arabic & \checkmark & & &
Maithili & & & \checkmark \\

Assamese & & & \checkmark &
Malayalam & & & \checkmark \\

Bajjika & & & \checkmark &
Maltese & & \checkmark & \\

Bamanankan & & & \checkmark &
Marathi & \checkmark & & \\

Bavarian & & & \checkmark &
Moore & & & \checkmark \\

Bengali & & & \checkmark &
Nepali & & & \checkmark \\

Bhojpuri & & & \checkmark &
Nigerian Fulfulde & & & \checkmark \\

Bulgarian & & \checkmark & &
Nigerian Pidgin & & & \checkmark \\

Burmese & & & \checkmark &
Northern Kurdish & & & \checkmark \\

Cantonese & & & \checkmark &
Northern Pashto & & & \checkmark \\

Cebuano & & & \checkmark &
Northern Sotho & & & \checkmark \\

Central Malay & & & \checkmark &
Northern Uzbek & & & \checkmark \\

Chhattisgarhi & & & \checkmark &
Norwegian Bokm\r{a}l & & \checkmark & \\

Chichewa & & & \checkmark &
Odia & & & \checkmark \\

Chinese & \checkmark & & &
Polish & & \checkmark & \\

Chittagonian & & & \checkmark &
Portuguese & & \checkmark & \\

Croatian & & \checkmark & &
Romanian & & \checkmark & \\

Czech & & \checkmark & &
Rundi & & & \checkmark \\

Danish & \checkmark & \checkmark & &
Russian & \checkmark & & \\

Dutch & \checkmark & \checkmark & &
Sadri & & & \checkmark \\

English & & \checkmark & &
Saraiki & & & \checkmark \\

Estonian & \checkmark & \checkmark & &
Setswana & & & \checkmark \\

Faroese & (\checkmark) & & &
Shona & & & \checkmark \\

Finnish & & \checkmark & &
Sindhi & & & \checkmark \\

French & \checkmark & \checkmark & &
Sinhala & & & \checkmark \\

Ganda & & & \checkmark &
Slovak & & \checkmark & \\

German & \checkmark & \checkmark & &
Slovenian & & \checkmark & \\

Greek & & \checkmark & &
Somali & & & \checkmark \\

Gujarati & & & \checkmark &
South Azerbaijani & & & \checkmark \\

Haitian Creole & & & \checkmark &
Spanish & & \checkmark & \\

Hausa & & & \checkmark &
Sundanese & & & \checkmark \\

Hindi & \checkmark & & &
Swahili & & & \checkmark \\

Hungarian & & \checkmark & &
Swedish & (\checkmark) & \checkmark & \\

Icelandic & \checkmark & & &
Tagalog & & & \checkmark \\

Igbo & & & \checkmark &
Tamil & & & \checkmark \\

Indonesian & & & \checkmark &
Telugu & & & \checkmark \\

Irish & & \checkmark & &
Thai & & & \checkmark \\

Italian & \checkmark & \checkmark & &
Turkish & & & \checkmark \\

Japanese & \checkmark & & &
Ukrainian & \checkmark & & \\

Javanese & & & \checkmark &
Urdu & (\checkmark) & & \checkmark \\

Jula & & & \checkmark &
Uyghur & & & \checkmark \\

Kannada & & & \checkmark &
Vietnamese & & & \checkmark \\

Kazakh & & & \checkmark &
West Central Oromo & & & \checkmark \\

Khmer & & & \checkmark &
Western Persian & & & \checkmark \\

Kinyarwanda & & & \checkmark &
Western Punjabi & (\checkmark) & & \checkmark \\

Kituba & & & \checkmark &
Wolof & & & \checkmark \\

Korean & & & \checkmark &
Xhosa & & & \checkmark \\

Latvian & & \checkmark & &
Yoruba & & & \checkmark \\

Lingala & & & \checkmark &
Zulu & & & \checkmark \\

\bottomrule
\end{tabularx}
\caption{Overview of included languages: We include a total of 102 languages. Besides the human verified languages we fill in languages from the Ethnologue top 200, prioritizing one main language in cases of a family of multiple languages and full coverage of the official European languages. For example, we select only standard Arabic and not Egyptian Arabic or Algerian Arabic. Languages with check marks in parenthesis are in the process of validation and will to added to the analysis later.}
\label{app:tab:langauge-overview}
\end{table}

\subsection{Language Annotation overview}
\label{app:sec:language-anno-overview}

Table~\ref{tab:language-validation} details the annotation process of each language.

\begin{table*}
\centering
\small
\setlength{\tabcolsep}{4pt}
\begin{tabular}{llccll}
\hline
Language & Human review & Comp. validated & Error analysis & Initial translation model & Source lang. \\
\hline
ara & a native speaker & $\checkmark$ & $\checkmark$ & anthropic/claude-opus-4-8 & eng \\
dan & three native speakers & $\checkmark$ & $\checkmark$ & gpt-5.4 & eng \\
deu & two native speakers & $\checkmark$ & $\checkmark$ & gpt-5.4 & dan \\
eng & a native speaker & $\checkmark$ & $\checkmark$ &  &  \\
est & a native speaker & $\checkmark$ & $\checkmark$ & gpt-5.4-nano & eng\_metric \\
fra & a native speaker & $\checkmark$ & $\checkmark$ & anthropic/claude-opus-4-8 & dan \\
hin & a native speaker & $\checkmark$ & $\checkmark$ & gpt-5.4 & eng \\
ita & two native speakers & $\checkmark$ & $\checkmark$ & gpt-5.4 & dan \\
isl & two native speakers & $\checkmark$ & $\checkmark$ & gpt-5.4 & dan \\
jpn & a native speaker & $\checkmark$ & $\checkmark$ & gpt-5.4 & eng \\
mar & a native speaker & $\checkmark$ & $\checkmark$ & gpt-5.4 & eng \\
nld & a native speaker & $\checkmark$ & $\checkmark$ & gpt-5.4; gpt-5.4-nano & dan \\
rus & a native speaker & $\checkmark$ & $\checkmark$ & gpt-5.4 & dan \\
ukr & a native speaker & $\checkmark$ & $\checkmark$ & gpt-5.4 & dan \\
zho & a native speaker & $\checkmark$ & $\checkmark$ & openai/gpt-5.4-2026-03-05 & eng \\
\hline
\end{tabular}
\caption{Human-validated languages and reviews in progress, alongside the English originals. Computational validation is enforced by CI. Error analysis is marked only when complete for all active templates. Note that the numbers does not match with the number of translators as some translated more than one language (bilinguals) and some are in the process evaluation.}
\label{tab:language-validation}
\end{table*}


\section{Machine vs Human-translation Full Comparison}
\label{app:sec:icelandic-verification}
Below we present the full set of distributions for the models evaluated on the machine-translated and verified Icelandic subsets. See Appendix~\ref{additional_languages} for the analysis.

\begin{figure}
    \centering
    \includegraphics[width=0.90\textwidth]{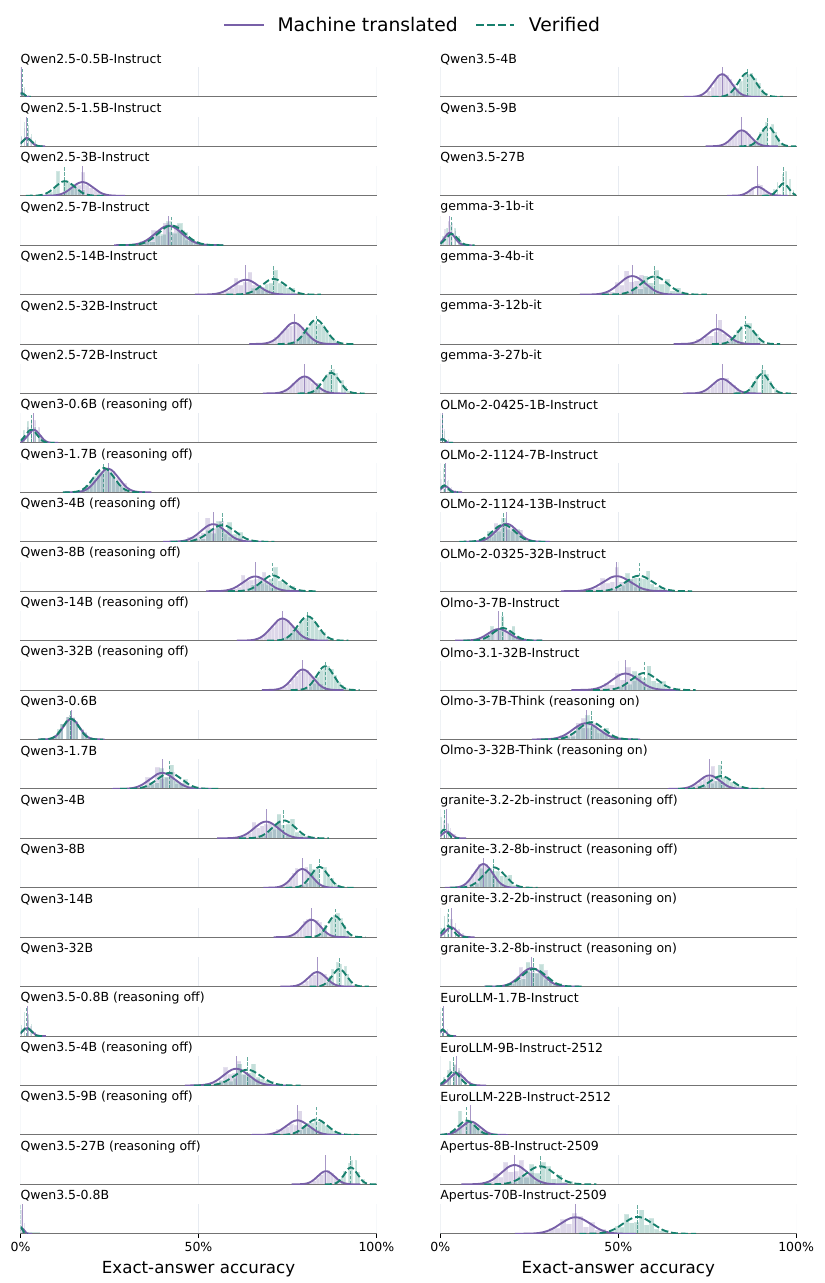}
    \caption{Distribution of model performances on the machine-translated and verified Icelandic subsets.}
    \label{fig:icelandic-full-graph}
\end{figure}
\clearpage

\section{English Standard vs English Metric Full Comparison}
\label{app:sec:metric-verification}
Below we present the full set of distributions for the models evaluated on the English Standard and English Metric subsets. See Appendix~\ref{app:sec:standard-v-metric} for the analysis.

\begin{figure}
    \centering
    \includegraphics[width=0.90\textwidth]{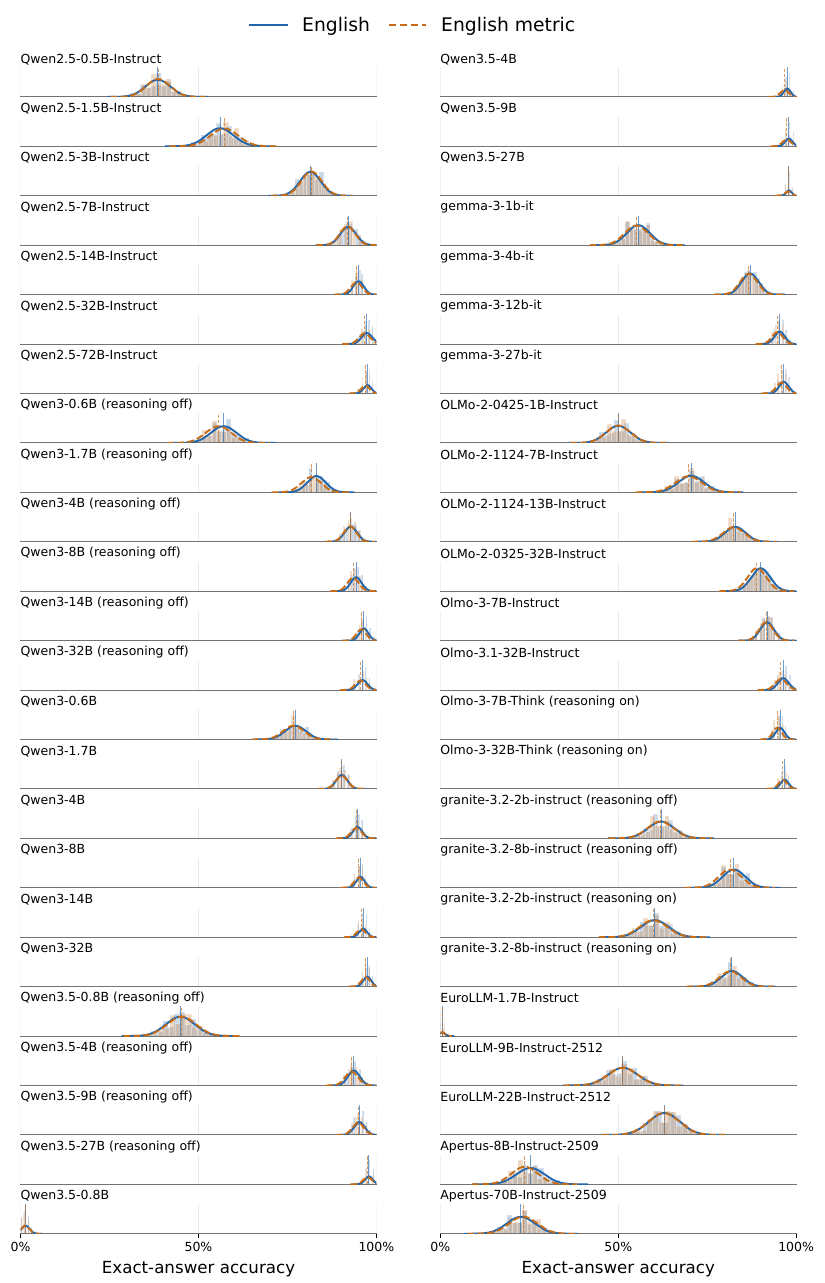}
    \caption{Distribution of model performances on the English and metric-converted English subsets.}
    \label{fig:eng-metric-full}
\end{figure}
\clearpage

\newpage
\section{Model Results}

This section reports the models results both as a simple overview (Table~\ref{tab:leaderboard}) and a detailed per-language table (Table~\ref{app:tab:original-vs-synthetic-accuracy}).

\subsection{Overview}

Table~\ref{tab:leaderboard} reports every run in the suite, including the two models that were included (marked with
$^{*}$).

\begin{table}[H]
\centering
\small
\resizebox{\textwidth}{!}{
\begin{tabular}{llrrrrrrl}
\toprule
Model & Reas. & Params & Orig. & Symb. & English & Non-Eng. & Gap & Worst \\
\midrule
Qwen3.5-27B & on & 27.0 & 96.3 & 94.9 & 98.0 & 94.7 & 3.3 & 90.2 (mar) \\
Qwen3.5-27B & off & 27.0 & 95.4 & 93.7 & 97.9 & 93.5 & 4.4 & 86.9 (mar) \\
Qwen3-32B & on & 32.0 & 93.9 & 93.1 & 97.5 & 92.8 & 4.7 & 89.5 (isl) \\
Qwen3-14B & on & 14.0 & 93.0 & 91.7 & 96.2 & 91.4 & 4.8 & 88.4 (isl) \\
Qwen3.5-9B & on & 9.0 & 93.8 & 91.7 & 98.0 & 91.2 & 6.7 & 73.6 (mar) \\
Qwen2.5-72B-Instruct & off & 72.0 & 93.6 & 91.2 & 97.5 & 90.7 & 6.8 & 80.2 (mar) \\
gemma-3-27b-it & off & 27.0 & 93.2 & 91.1 & 96.5 & 90.7 & 5.8 & 86.7 (mar) \\
Qwen3-32B & off & 32.0 & 92.5 & 90.9 & 96.1 & 90.5 & 5.6 & 85.5 (isl) \\
Qwen3-8B & on & 8.0 & 91.1 & 89.8 & 95.5 & 89.4 & 6.1 & 84.0 (isl) \\
Qwen3.5-4B & on & 4.0 & 91.8 & 89.8 & 97.5 & 89.3 & 8.2 & 66.2 (mar) \\
Qwen3-14B & off & 14.0 & 90.4 & 89.4 & 96.5 & 88.9 & 7.6 & 80.8 (isl) \\
Qwen2.5-32B-Instruct & off & 32.0 & 91.7 & 89.1 & 97.4 & 88.5 & 8.9 & 72.0 (mar) \\
gemma-3-12b-it & off & 12.0 & 90.6 & 88.0 & 95.4 & 87.5 & 7.9 & 77.8 (mar) \\
Qwen3-4B & on & 4.0 & 88.9 & 87.6 & 94.8 & 87.1 & 7.6 & 74.0 (isl) \\
Olmo-3-32B-Think & on & 32.0 & 89.3 & 86.9 & 96.8 & 86.2 & 10.5 & 69.7 (mar) \\
Qwen3.5-9B & off & 9.0 & 89.1 & 86.1 & 95.2 & 85.4 & 9.8 & 50.6 (mar) \\
Qwen3-8B & off & 8.0 & 87.5 & 85.1 & 94.3 & 84.4 & 9.9 & 70.9 (isl) \\
Qwen2.5-14B-Instruct & off & 14.0 & 87.3 & 84.6 & 95.0 & 83.8 & 11.1 & 62.4 (mar) \\
Qwen3-4B & off & 4.0 & 82.6 & 80.7 & 92.8 & 79.8 & 13.1 & 56.7 (isl) \\
Qwen3.5-4B & off & 4.0 & 81.8 & 78.0 & 93.7 & 76.9 & 16.8 & 35.7 (mar) \\
Qwen3-1.7B & on & 1.7 & 78.1 & 75.4 & 90.3 & 74.4 & 16.0 & 41.8 (isl) \\
Olmo-3.1-32B-Instruct & off & 32.0 & 77.3 & 74.7 & 96.4 & 73.2 & 23.2 & 19.3 (est) \\
Qwen2.5-7B-Instruct & off & 7.0 & 77.6 & 73.6 & 92.1 & 72.3 & 19.8 & 31.9 (mar) \\
Olmo-3-7B-Think & on & 7.0 & 77.5 & 73.5 & 95.5 & 72.0 & 23.5 & 29.4 (est) \\
gemma-3-4b-it & off & 4.0 & 77.1 & 72.3 & 87.1 & 71.2 & 15.8 & 50.7 (mar) \\
OLMo-2-0325-32B-Instruct & off & 32.0 & 65.9 & 59.6 & 90.1 & 57.5 & 32.7 & 24.9 (est) \\
Qwen3-1.7B & off & 1.7 & 61.9 & 57.4 & 83.1 & 55.6 & 27.5 & 23.4 (isl) \\
Apertus-70B-Instruct-2509 & off & 70.0 & 61.7 & 56.1 & 22.7 & 58.5 & -35.8 & 22.7 (eng) \\
granite-3.2-8b-instruct & on & 8.0 & 60.9 & 55.3 & 82.0 & 53.4 & 28.6 & 18.2 (est) \\
Qwen2.5-3B-Instruct & off & 3.0 & 57.6 & 52.7 & 81.5 & 50.7 & 30.8 & 12.4 (isl) \\
Olmo-3-7B-Instruct & off & 7.0 & 57.4 & 51.9 & 91.8 & 49.0 & 42.8 & 10.6 (mar) \\
EuroLLM-22B-Instruct-2512 & off & 22.0 & 55.8 & 51.0 & 63.0 & 50.1 & 12.9 & 2.1 (mar) \\
Qwen3-0.6B & on & 0.6 & 50.3 & 46.2 & 77.3 & 44.0 & 33.3 & 12.4 (mar) \\
granite-3.2-8b-instruct & off & 8.0 & 50.4 & 44.3 & 82.6 & 41.5 & 41.1 & 7.3 (est) \\
EuroLLM-9B-Instruct-2512 & off & 9.0 & 49.2 & 43.9 & 51.2 & 43.4 & 7.9 & 1.2 (mar) \\
OLMo-2-1124-13B-Instruct & off & 13.0 & 51.1 & 43.4 & 82.9 & 40.6 & 42.3 & 2.9 (mar) \\
Apertus-8B-Instruct-2509 & off & 8.0 & 34.9 & 30.4 & 25.2 & 30.8 & -5.6 & 16.2 (jpn) \\
OLMo-2-1124-7B-Instruct & off & 7.0 & 32.1 & 26.1 & 70.5 & 22.9 & 47.6 & 1.1 (isl) \\
granite-3.2-2b-instruct & on & 2.0 & 30.1 & 25.9 & 60.2 & 23.4 & 36.8 & 1.5 (est) \\
granite-3.2-2b-instruct & off & 2.0 & 26.6 & 23.8 & 62.1 & 21.1 & 41.0 & 1.1 (est) \\
Qwen3-0.6B & off & 0.6 & 28.0 & 23.5 & 57.0 & 21.1 & 35.9 & 2.4 (mar) \\
Qwen2.5-1.5B-Instruct & off & 1.5 & 21.3 & 19.6 & 56.0 & 17.0 & 39.0 & 1.5 (mar) \\
gemma-3-1b-it & off & 1.0 & 22.7 & 19.5 & 55.6 & 17.0 & 38.6 & 0.3 (mar) \\
Qwen3.5-0.8B & off & 0.8 & 10.0 & 10.7 & 44.8 & 8.2 & 36.5 & 0.2 (mar) \\
OLMo-2-0425-1B-Instruct & off & 1.0 & 9.9 & 8.6 & 49.9 & 5.7 & 44.2 & 0.4 (mar) \\
Qwen2.5-0.5B-Instruct & off & 0.5 & 6.5 & 5.4 & 38.6 & 3.0 & 35.7 & 0.4 (isl) \\
EuroLLM-1.7B-Instruct$^{*}$ & off & 1.7 & 0.3 & 0.4 & 0.6 & 0.4 & 0.2 & 0.1 (ara) \\
Qwen3.5-0.8B$^{*}$ & on & 0.8 & 0.3 & 0.3 & 1.4 & 0.2 & 1.2 & 0.0 (est) \\
\bottomrule
\end{tabular}
}
\caption{Exact-answer accuracy on Multilingual GSM-Symbolic, averaged over the 15 benchmark languages and ordered by symbolic accuracy. \emph{Gap} is English minus the mean of the other 14 languages. \emph{Worst} is the lowest-scoring language on the symbolic split. Per-language scores carry roughly $\pm 2.3$ points of sampling error on the symbolic split and $\pm 3.4$ on the original; the 15-language averages about $\pm 1.4$ (Appendix~\ref{app:sec:sample-size}). $^{*}$Reported here but excluded from the model in Equation~\ref{eq:model}.}
\label{tab:leaderboard}
\end{table}

\subsection{Model results by language}
\label{app:sec:model-results-by-language}

In Table~\ref{app:tab:original-vs-synthetic-accuracy} we present the results of each model on each of the validated languages. To allow for further analysis we also distribute the response logs along with the dataset.

\begingroup
\tiny
\setlength{\tabcolsep}{1.25pt}
\renewcommand{\arraystretch}{0.9}
\begin{longtable}{p{0.145\linewidth}lrr@{\hspace{4pt}}p{0.145\linewidth}lrr}
\caption{Original and synthetic exact-answer accuracy by model and language for all 48 models. Bracketed values are 95\% confidence intervals (CI), computed using Wilson score intervals over scored examples within each split.}\label{app:tab:original-vs-synthetic-accuracy} \\
\toprule
Model & Language & \shortstack{Original\\Accuracy} & \shortstack{Synthetic\\Accuracy} & Model & Language & \shortstack{Original\\Accuracy} & \shortstack{Synthetic\\Accuracy} \\
\midrule
\endfirsthead
\multicolumn{8}{c}{\tablename\ \thetable{} -- continued from previous page} \\
\toprule
Model & Language & \shortstack{Original\\Accuracy} & \shortstack{Synthetic\\Accuracy} & Model & Language & \shortstack{Original\\Accuracy} & \shortstack{Synthetic\\Accuracy} \\
\midrule
\endhead
\midrule
\multicolumn{8}{r}{Continued on next page} \\
\endfoot
\bottomrule
\endlastfoot
Qwen2.5-0.5B-Instruct & Chinese & 16.0\% [10.1-24.4] & 8.3\% [7.2-9.6] & Qwen3.5-4B & Chinese & 95.0\% [88.8-97.8] & 94.6\% [93.5-95.5] \\
 & Hindi & 1.0\% [0.2-5.4] & 0.6\% [0.3-1.0] &  & Hindi & 85.0\% [76.7-90.7] & 83.4\% [81.7-84.9] \\
 & English & 42.0\% [32.8-51.8] & 38.6\% [36.5-40.8] &  & English & 98.0\% [93.0-99.4] & 97.5\% [96.7-98.1] \\
 & English metric & 41.0\% [31.9-50.8] & 38.7\% [36.6-40.9] &  & English metric & 97.0\% [91.5-99.0] & 96.7\% [95.8-97.4] \\
 & Arabic & 3.0\% [1.0-8.5] & 1.1\% [0.7-1.7] &  & Arabic & 92.0\% [85.0-95.9] & 92.4\% [91.2-93.5] \\
 & Japanese & 2.2\% [1.2-4.2] & 1.7\% [1.4-2.0] &  & Japanese & 95.0\% [91.0-97.3] & 91.5\% [90.6-92.3] \\
 & Russian & 4.0\% [1.6-9.8] & 3.9\% [3.1-4.8] &  & Russian & 93.0\% [86.3-96.6] & 89.3\% [87.9-90.6] \\
 & German & 7.8\% [5.8-10.5] & 6.5\% [5.9-7.1] &  & German & 95.0\% [91.0-97.3] & 92.2\% [91.4-93.0] \\
 & Marathi & 5.0\% [2.2-11.2] & 0.9\% [0.6-1.4] &  & Marathi & 65.0\% [55.3-73.6] & 66.2\% [64.1-68.2] \\
 & French & 9.0\% [4.8-16.2] & 5.1\% [4.3-6.2] &  & French & 95.0\% [88.8-97.8] & 91.3\% [90.0-92.5] \\
 & Italian & 2.0\% [0.6-7.0] & 3.6\% [2.9-4.5] &  & Italian & 95.0\% [88.8-97.8] & 95.0\% [94.0-95.9] \\
 & Ukrainian & 3.0\% [1.0-8.5] & 1.5\% [1.0-2.1] &  & Ukrainian & 93.0\% [86.3-96.6] & 92.5\% [91.2-93.5] \\
 & Dutch & 7.0\% [3.4-13.7] & 6.8\% [5.7-7.9] &  & Dutch & 94.0\% [87.5-97.2] & 90.5\% [89.1-91.7] \\
 & Danish & 1.6\% [0.8-3.1] & 1.8\% [1.5-2.2] &  & Danish & 93.5\% [89.2-96.2] & 92.3\% [91.4-93.1] \\
 & Estonian & 0.8\% [0.3-2.2] & 0.8\% [0.6-1.1] &  & Estonian & 96.0\% [90.2-98.4] & 93.6\% [92.4-94.6] \\
 & Icelandic & 0.0\% [0.0-3.7] & 0.4\% [0.2-0.8] &  & Icelandic & 92.0\% [85.0-95.9] & 86.4\% [84.8-87.8] \\
\addlinespace[2pt]
Qwen2.5-1.5B-Instruct & Chinese & 16.0\% [10.1-24.4] & 19.1\% [17.4-20.9] & Qwen3.5-9B & Chinese & 97.0\% [91.5-99.0] & 95.2\% [94.2-96.1] \\
 & Hindi & 7.0\% [3.4-13.7] & 3.9\% [3.1-4.8] &  & Hindi & 87.0\% [79.0-92.2] & 87.5\% [86.0-88.9] \\
 & English & 59.0\% [49.2-68.1] & 56.0\% [53.8-58.2] &  & English & 98.0\% [93.0-99.4] & 98.0\% [97.2-98.5] \\
 & English metric & 60.0\% [50.2-69.1] & 57.3\% [55.1-59.5] &  & English metric & 99.0\% [94.6-99.8] & 97.4\% [96.6-98.0] \\
 & Arabic & 16.0\% [10.1-24.4] & 11.7\% [10.3-13.1] &  & Arabic & 92.0\% [85.0-95.9] & 95.3\% [94.3-96.1] \\
 & Japanese & 23.0\% [18.6-28.1] & 19.5\% [18.5-20.5] &  & Japanese & 95.0\% [91.0-97.3] & 91.8\% [90.9-92.6] \\
 & Russian & 31.0\% [22.8-40.6] & 28.2\% [26.3-30.2] &  & Russian & 95.0\% [88.8-97.8] & 90.9\% [89.6-92.1] \\
 & German & 35.0\% [29.8-40.6] & 34.3\% [33.1-35.5] &  & German & 95.5\% [91.7-97.6] & 93.6\% [92.8-94.3] \\
 & Marathi & 2.0\% [0.6-7.0] & 1.5\% [1.0-2.1] &  & Marathi & 81.0\% [72.2-87.5] & 73.6\% [71.6-75.5] \\
 & French & 25.0\% [17.5-34.3] & 21.8\% [20.0-23.6] &  & French & 96.0\% [90.2-98.4] & 92.2\% [90.9-93.3] \\
 & Italian & 40.0\% [30.9-49.8] & 33.7\% [31.6-35.8] &  & Italian & 98.0\% [93.0-99.4] & 95.5\% [94.4-96.3] \\
 & Ukrainian & 20.0\% [13.3-28.9] & 19.4\% [17.7-21.2] &  & Ukrainian & 94.0\% [87.5-97.2] & 92.8\% [91.5-93.8] \\
 & Dutch & 27.0\% [19.3-36.4] & 25.6\% [23.7-27.5] &  & Dutch & 94.0\% [87.5-97.2] & 89.9\% [88.5-91.1] \\
 & Danish & 17.0\% [13.2-21.7] & 15.0\% [14.1-15.9] &  & Danish & 95.5\% [91.7-97.6] & 94.2\% [93.5-94.9] \\
 & Estonian & 2.3\% [1.1-4.7] & 2.4\% [2.0-2.8] &  & Estonian & 96.0\% [90.2-98.4] & 95.0\% [94.0-95.9] \\
 & Icelandic & 3.0\% [1.0-8.5] & 1.8\% [1.3-2.5] &  & Icelandic & 96.0\% [90.2-98.4] & 92.0\% [90.7-93.1] \\
\addlinespace[2pt]
Qwen2.5-3B-Instruct & Chinese & 74.0\% [64.6-81.6] & 71.5\% [69.5-73.4] & Qwen3.5-27B & Chinese & 98.0\% [93.0-99.4] & 97.3\% [96.5-97.9] \\
 & Hindi & 48.0\% [38.5-57.7] & 38.1\% [36.0-40.2] &  & Hindi & 95.0\% [88.8-97.8] & 94.0\% [92.8-94.9] \\
 & English & 81.0\% [72.2-87.5] & 81.5\% [79.7-83.1] &  & English & 99.0\% [94.6-99.8] & 98.0\% [97.3-98.5] \\
 & English metric & 83.0\% [74.5-89.1] & 81.8\% [80.0-83.4] &  & English metric & 99.0\% [94.6-99.8] & 97.8\% [97.0-98.3] \\
 & Arabic & 70.0\% [60.4-78.1] & 62.1\% [60.0-64.2] &  & Arabic & 95.0\% [88.8-97.8] & 96.0\% [95.0-96.7] \\
 & Japanese & 63.3\% [57.7-68.6] & 55.0\% [53.7-56.2] &  & Japanese & 96.0\% [92.3-98.0] & 94.3\% [93.5-95.0] \\
 & Russian & 70.0\% [60.4-78.1] & 65.6\% [63.5-67.7] &  & Russian & 96.0\% [90.2-98.4] & 92.5\% [91.2-93.5] \\
 & German & 75.7\% [70.5-80.2] & 67.1\% [65.9-68.3] &  & German & 96.5\% [93.0-98.3] & 94.2\% [93.4-94.8] \\
 & Marathi & 19.0\% [12.5-27.8] & 14.1\% [12.7-15.7] &  & Marathi & 91.0\% [83.8-95.2] & 90.2\% [88.9-91.5] \\
 & French & 70.0\% [60.4-78.1] & 69.8\% [67.8-71.8] &  & French & 98.0\% [93.0-99.4] & 94.5\% [93.5-95.5] \\
 & Italian & 75.0\% [65.7-82.5] & 69.2\% [67.1-71.1] &  & Italian & 96.0\% [90.2-98.4] & 96.2\% [95.3-97.0] \\
 & Ukrainian & 61.0\% [51.2-70.0] & 56.6\% [54.5-58.8] &  & Ukrainian & 97.0\% [91.5-99.0] & 96.0\% [95.1-96.8] \\
 & Dutch & 62.0\% [52.2-70.9] & 60.8\% [58.6-62.9] &  & Dutch & 97.0\% [91.5-99.0] & 93.0\% [91.8-94.0] \\
 & Danish & 60.0\% [54.4-65.4] & 53.8\% [52.5-55.0] &  & Danish & 97.5\% [94.3-98.9] & 96.4\% [95.7-96.9] \\
 & Estonian & 17.3\% [13.5-22.0] & 15.4\% [14.5-16.3] &  & Estonian & 98.0\% [93.0-99.4] & 96.9\% [96.0-97.6] \\
 & Icelandic & 15.0\% [9.3-23.3] & 12.4\% [11.0-13.9] &  & Icelandic & 98.0\% [93.0-99.4] & 96.4\% [95.5-97.1] \\
\addlinespace[2pt]
Qwen2.5-7B-Instruct & Chinese & 89.0\% [81.4-93.7] & 83.0\% [81.2-84.5] & gemma-3-1b-it & Chinese & 23.0\% [15.8-32.2] & 18.4\% [16.8-20.2] \\
 & Hindi & 66.0\% [56.3-74.5] & 69.2\% [67.1-71.2] &  & Hindi & 8.0\% [4.1-15.0] & 5.9\% [4.9-7.0] \\
 & English & 98.0\% [93.0-99.4] & 92.1\% [90.8-93.2] &  & English & 54.0\% [44.3-63.4] & 55.6\% [53.4-57.8] \\
 & English metric & 98.0\% [93.0-99.4] & 91.9\% [90.6-93.0] &  & English metric & 54.0\% [44.3-63.4] & 55.0\% [52.9-57.2] \\
 & Arabic & 81.0\% [72.2-87.5] & 77.3\% [75.4-79.1] &  & Arabic & 25.0\% [17.5-34.3] & 19.3\% [17.6-21.1] \\
 & Japanese & 85.0\% [80.5-88.6] & 78.2\% [77.1-79.2] &  & Japanese & 16.5\% [12.0-22.3] & 12.9\% [11.9-13.9] \\
 & Russian & 82.0\% [73.3-88.3] & 79.0\% [77.1-80.7] &  & Russian & 31.0\% [22.8-40.6] & 24.9\% [23.1-26.8] \\
 & German & 82.7\% [78.0-86.5] & 84.0\% [83.0-84.9] &  & German & 29.0\% [23.2-35.6] & 27.6\% [26.2-29.0] \\
 & Marathi & 37.0\% [28.2-46.8] & 31.9\% [29.9-34.0] &  & Marathi & 3.0\% [1.0-8.5] & 0.3\% [0.1-0.7] \\
 & French & 88.0\% [80.2-93.0] & 83.7\% [82.0-85.3] &  & French & 25.0\% [17.5-34.3] & 21.6\% [19.9-23.5] \\
 & Italian & 87.0\% [79.0-92.2] & 86.0\% [84.4-87.5] &  & Italian & 33.0\% [24.6-42.7] & 30.9\% [29.0-33.0] \\
 & Ukrainian & 83.0\% [74.5-89.1] & 78.2\% [76.4-80.0] &  & Ukrainian & 18.0\% [11.7-26.7] & 18.9\% [17.3-20.7] \\
 & Dutch & 88.0\% [80.2-93.0] & 82.8\% [81.1-84.4] &  & Dutch & 33.0\% [24.6-42.7] & 26.2\% [24.3-28.2] \\
 & Danish & 84.7\% [80.2-88.3] & 80.4\% [79.4-81.4] &  & Danish & 32.5\% [26.4-39.3] & 26.6\% [25.3-28.0] \\
 & Estonian & 65.3\% [59.8-70.5] & 58.5\% [57.3-59.8] &  & Estonian & 7.0\% [3.4-13.7] & 3.9\% [3.1-4.8] \\
 & Icelandic & 48.0\% [38.5-57.7] & 42.4\% [40.2-44.5] &  & Icelandic & 3.0\% [1.0-8.5] & 3.2\% [2.5-4.1] \\
\addlinespace[2pt]
Qwen2.5-14B-Instruct & Chinese & 90.0\% [82.6-94.5] & 90.0\% [88.6-91.2] & gemma-3-4b-it & Chinese & 68.0\% [58.3-76.3] & 68.8\% [66.7-70.7] \\
 & Hindi & 89.0\% [81.4-93.7] & 80.5\% [78.8-82.2] &  & Hindi & 74.0\% [64.6-81.6] & 66.9\% [64.8-68.9] \\
 & English & 97.0\% [91.5-99.0] & 95.0\% [93.9-95.8] &  & English & 88.0\% [80.2-93.0] & 87.1\% [85.6-88.5] \\
 & English metric & 96.0\% [90.2-98.4] & 94.4\% [93.3-95.3] &  & English metric & 85.0\% [76.7-90.7] & 86.7\% [85.1-88.1] \\
 & Arabic & 89.0\% [81.4-93.7] & 91.9\% [90.6-93.0] &  & Arabic & 82.0\% [73.3-88.3] & 74.6\% [72.6-76.5] \\
 & Japanese & 91.0\% [87.2-93.7] & 85.0\% [84.0-85.8] &  & Japanese & 63.0\% [56.1-69.4] & 61.0\% [59.5-62.5] \\
 & Russian & 89.0\% [81.4-93.7] & 85.6\% [84.0-87.1] &  & Russian & 81.0\% [72.2-87.5] & 73.0\% [71.0-74.9] \\
 & German & 93.3\% [89.9-95.6] & 88.8\% [87.9-89.5] &  & German & 81.5\% [75.5-86.3] & 77.3\% [76.0-78.6] \\
 & Marathi & 65.0\% [55.3-73.6] & 62.4\% [60.2-64.4] &  & Marathi & 64.0\% [54.2-72.7] & 50.7\% [48.5-52.9] \\
 & French & 93.0\% [86.3-96.6] & 89.7\% [88.3-91.0] &  & French & 85.0\% [76.7-90.7] & 82.5\% [80.7-84.1] \\
 & Italian & 97.0\% [91.5-99.0] & 92.0\% [90.8-93.2] &  & Italian & 87.0\% [79.0-92.2] & 84.0\% [82.3-85.5] \\
 & Ukrainian & 92.0\% [85.0-95.9] & 87.5\% [86.0-88.9] &  & Ukrainian & 74.0\% [64.6-81.6] & 75.2\% [73.3-77.1] \\
 & Dutch & 89.0\% [81.4-93.7] & 88.0\% [86.5-89.4] &  & Dutch & 83.0\% [74.5-89.1] & 76.2\% [74.3-78.1] \\
 & Danish & 89.0\% [85.0-92.1] & 88.2\% [87.4-89.0] &  & Danish & 79.0\% [72.8-84.1] & 78.0\% [76.7-79.2] \\
 & Estonian & 79.0\% [74.0-83.2] & 75.0\% [73.9-76.1] &  & Estonian & 77.0\% [67.8-84.2] & 71.3\% [69.3-73.2] \\
 & Icelandic & 75.0\% [65.7-82.5] & 71.2\% [69.2-73.1] &  & Icelandic & 69.0\% [59.4-77.2] & 60.2\% [58.0-62.3] \\
\addlinespace[2pt]
Qwen2.5-32B-Instruct & Chinese & 94.0\% [87.5-97.2] & 94.0\% [92.9-95.0] & gemma-3-12b-it & Chinese & 89.0\% [81.4-93.7] & 86.1\% [84.5-87.5] \\
 & Hindi & 93.0\% [86.3-96.6] & 90.2\% [88.8-91.4] &  & Hindi & 89.0\% [81.4-93.7] & 84.2\% [82.6-85.8] \\
 & English & 99.0\% [94.6-99.8] & 97.4\% [96.6-98.0] &  & English & 96.0\% [90.2-98.4] & 95.4\% [94.4-96.2] \\
 & English metric & 97.0\% [91.5-99.0] & 96.7\% [95.8-97.4] &  & English metric & 97.0\% [91.5-99.0] & 94.8\% [93.7-95.6] \\
 & Arabic & 95.0\% [88.8-97.8] & 92.9\% [91.7-93.9] &  & Arabic & 94.0\% [87.5-97.2] & 91.9\% [90.6-93.0] \\
 & Japanese & 93.0\% [89.5-95.4] & 89.4\% [88.6-90.1] &  & Japanese & 90.0\% [85.1-93.4] & 85.9\% [84.8-86.9] \\
 & Russian & 91.0\% [83.8-95.2] & 87.1\% [85.5-88.5] &  & Russian & 89.0\% [81.4-93.7] & 85.1\% [83.5-86.6] \\
 & German & 92.3\% [88.8-94.8] & 90.7\% [90.0-91.4] &  & German & 90.0\% [85.1-93.4] & 89.0\% [88.0-90.0] \\
 & Marathi & 83.0\% [74.5-89.1] & 72.0\% [70.0-73.9] &  & Marathi & 87.0\% [79.0-92.2] & 77.8\% [76.0-79.6] \\
 & French & 94.0\% [87.5-97.2] & 90.5\% [89.2-91.8] &  & French & 92.0\% [85.0-95.9] & 87.8\% [86.2-89.1] \\
 & Italian & 94.0\% [87.5-97.2] & 92.7\% [91.4-93.7] &  & Italian & 90.0\% [82.6-94.5] & 91.3\% [90.0-92.5] \\
 & Ukrainian & 92.0\% [85.0-95.9] & 89.5\% [88.0-90.7] &  & Ukrainian & 91.0\% [83.8-95.2] & 88.5\% [87.1-89.9] \\
 & Dutch & 91.0\% [83.8-95.2] & 90.4\% [89.0-91.6] &  & Dutch & 92.0\% [85.0-95.9] & 89.1\% [87.7-90.4] \\
 & Danish & 93.0\% [89.5-95.4] & 91.9\% [91.2-92.5] &  & Danish & 91.5\% [86.8-94.6] & 91.5\% [90.5-92.3] \\
 & Estonian & 90.0\% [86.1-92.9] & 87.0\% [86.1-87.8] &  & Estonian & 92.0\% [85.0-95.9] & 91.2\% [89.9-92.4] \\
 & Icelandic & 87.0\% [79.0-92.2] & 83.0\% [81.3-84.6] &  & Icelandic & 93.0\% [86.3-96.6] & 85.9\% [84.3-87.3] \\
\addlinespace[2pt]
Qwen2.5-72B-Instruct & Chinese & 95.0\% [88.8-97.8] & 94.6\% [93.5-95.5] & gemma-3-27b-it & Chinese & 92.0\% [85.0-95.9] & 90.6\% [89.3-91.8] \\
 & Hindi & 93.0\% [86.3-96.6] & 92.5\% [91.2-93.5] &  & Hindi & 94.0\% [87.5-97.2] & 90.1\% [88.8-91.4] \\
 & English & 98.0\% [93.0-99.4] & 97.5\% [96.7-98.1] &  & English & 95.0\% [88.8-97.8] & 96.5\% [95.7-97.3] \\
 & English metric & 98.0\% [93.0-99.4] & 96.9\% [96.0-97.6] &  & English metric & 96.0\% [90.2-98.4] & 95.9\% [94.9-96.7] \\
 & Arabic & 94.0\% [87.5-97.2] & 94.7\% [93.6-95.6] &  & Arabic & 93.0\% [86.3-96.6] & 91.2\% [89.9-92.4] \\
 & Japanese & 96.0\% [92.3-98.0] & 91.7\% [90.8-92.5] &  & Japanese & 94.5\% [90.4-96.9] & 89.8\% [88.9-90.7] \\
 & Russian & 93.0\% [86.3-96.6] & 87.6\% [86.1-89.0] &  & Russian & 92.0\% [85.0-95.9] & 89.0\% [87.6-90.3] \\
 & German & 95.5\% [91.7-97.6] & 92.4\% [91.5-93.2] &  & German & 95.0\% [91.0-97.3] & 91.1\% [90.2-91.9] \\
 & Marathi & 89.0\% [81.4-93.7] & 80.2\% [78.4-81.9] &  & Marathi & 92.0\% [85.0-95.9] & 86.7\% [85.1-88.1] \\
 & French & 95.0\% [88.8-97.8] & 92.2\% [90.9-93.3] &  & French & 94.0\% [87.5-97.2] & 90.4\% [89.0-91.6] \\
 & Italian & 94.0\% [87.5-97.2] & 93.8\% [92.7-94.8] &  & Italian & 92.0\% [85.0-95.9] & 92.5\% [91.3-93.6] \\
 & Ukrainian & 91.0\% [83.8-95.2] & 91.5\% [90.2-92.6] &  & Ukrainian & 92.0\% [85.0-95.9] & 90.8\% [89.5-92.0] \\
 & Dutch & 95.0\% [88.8-97.8] & 92.5\% [91.3-93.6] &  & Dutch & 94.0\% [87.5-97.2] & 92.3\% [91.1-93.4] \\
 & Danish & 94.5\% [90.4-96.9] & 92.6\% [91.7-93.3] &  & Danish & 95.5\% [91.7-97.6] & 93.7\% [92.9-94.4] \\
 & Estonian & 90.5\% [85.6-93.8] & 88.6\% [87.6-89.6] &  & Estonian & 94.0\% [87.5-97.2] & 92.0\% [90.7-93.1] \\
 & Icelandic & 92.0\% [85.0-95.9] & 87.4\% [85.8-88.7] &  & Icelandic & 90.0\% [82.6-94.5] & 90.5\% [89.2-91.8] \\
\addlinespace[2pt]
Qwen3-0.6B (reasoning off) & Chinese & 36.0\% [27.3-45.8] & 37.3\% [35.2-39.4] & OLMo-2-0425-1B-Instruct & Chinese & 7.0\% [3.4-13.7] & 9.3\% [8.2-10.7] \\
 & Hindi & 8.0\% [4.1-15.0] & 5.0\% [4.1-6.0] &  & Hindi & 0.0\% [0.0-3.7] & 0.8\% [0.5-1.2] \\
 & English & 83.0\% [74.5-89.1] & 57.0\% [54.8-59.2] &  & English & 61.0\% [51.2-70.0] & 49.9\% [47.7-52.1] \\
 & English metric & 59.0\% [49.2-68.1] & 55.6\% [53.4-57.8] &  & English metric & 68.0\% [58.3-76.3] & 50.0\% [47.8-52.2] \\
 & Arabic & 14.0\% [8.5-22.1] & 19.0\% [17.3-20.8] &  & Arabic & 4.0\% [1.6-9.8] & 2.1\% [1.6-2.9] \\
 & Japanese & 32.0\% [25.9-38.8] & 26.6\% [25.2-27.9] &  & Japanese & 0.5\% [0.1-2.8] & 1.9\% [1.5-2.4] \\
 & Russian & 43.0\% [33.7-52.8] & 35.6\% [33.6-37.8] &  & Russian & 15.0\% [9.3-23.3] & 9.7\% [8.4-11.0] \\
 & German & 40.5\% [33.9-47.4] & 29.1\% [27.7-30.5] &  & German & 13.0\% [9.0-18.4] & 12.3\% [11.4-13.4] \\
 & Marathi & 5.0\% [2.2-11.2] & 2.4\% [1.8-3.1] &  & Marathi & 1.0\% [0.2-5.4] & 0.4\% [0.2-0.9] \\
 & French & 40.0\% [30.9-49.8] & 34.8\% [32.8-37.0] &  & French & 24.0\% [16.7-33.2] & 20.0\% [18.3-21.8] \\
 & Italian & 33.0\% [24.6-42.7] & 28.6\% [26.7-30.6] &  & Italian & 14.0\% [8.5-22.1] & 11.9\% [10.6-13.4] \\
 & Ukrainian & 32.0\% [23.7-41.7] & 26.4\% [24.5-28.3] &  & Ukrainian & 0.0\% [0.0-3.7] & 2.4\% [1.8-3.2] \\
 & Dutch & 29.0\% [21.0-38.5] & 26.8\% [24.9-28.8] &  & Dutch & 7.0\% [3.4-13.7] & 6.0\% [5.1-7.2] \\
 & Danish & 23.0\% [17.7-29.3] & 21.1\% [19.9-22.4] &  & Danish & 3.0\% [1.4-6.4] & 2.1\% [1.7-2.6] \\
 & Estonian & 3.5\% [1.7-7.0] & 3.3\% [2.8-3.9] &  & Estonian & 0.0\% [0.0-3.7] & 0.6\% [0.3-1.0] \\
 & Icelandic & 6.0\% [2.8-12.5] & 3.0\% [2.3-3.8] &  & Icelandic & 1.0\% [0.2-5.4] & 0.6\% [0.3-1.0] \\
\addlinespace[2pt]
Qwen3-1.7B (reasoning off) & Chinese & 76.0\% [66.8-83.3] & 73.5\% [71.5-75.3] & OLMo-2-1124-7B-Instruct & Chinese & 39.0\% [30.0-48.8] & 33.1\% [31.1-35.2] \\
 & Hindi & 49.0\% [39.4-58.7] & 46.5\% [44.3-48.6] &  & Hindi & 7.0\% [3.4-13.7] & 3.1\% [2.4-4.0] \\
 & English & 92.0\% [85.0-95.9] & 83.1\% [81.4-84.7] &  & English & 83.0\% [74.5-89.1] & 70.5\% [68.5-72.5] \\
 & English metric & 83.0\% [74.5-89.1] & 81.7\% [79.9-83.3] &  & English metric & 79.0\% [70.0-85.8] & 69.8\% [67.7-71.7] \\
 & Arabic & 61.0\% [51.2-70.0] & 58.2\% [56.1-60.4] &  & Arabic & 21.0\% [14.2-30.0] & 16.7\% [15.1-18.4] \\
 & Japanese & 58.0\% [51.1-64.6] & 55.8\% [54.2-57.3] &  & Japanese & 21.5\% [16.4-27.7] & 15.4\% [14.4-16.6] \\
 & Russian & 72.0\% [62.5-79.9] & 64.1\% [62.0-66.2] &  & Russian & 51.0\% [41.3-60.6] & 40.4\% [38.3-42.6] \\
 & German & 71.5\% [64.9-77.3] & 66.9\% [65.5-68.4] &  & German & 54.0\% [47.1-60.8] & 45.7\% [44.1-47.2] \\
 & Marathi & 39.0\% [30.0-48.8] & 30.4\% [28.4-32.5] &  & Marathi & 2.0\% [0.6-7.0] & 1.5\% [1.0-2.1] \\
 & French & 73.0\% [63.6-80.7] & 67.3\% [65.2-69.3] &  & French & 59.0\% [49.2-68.1] & 55.7\% [53.5-57.9] \\
 & Italian & 69.0\% [59.4-77.2] & 67.1\% [65.0-69.1] &  & Italian & 54.0\% [44.3-63.4] & 45.5\% [43.3-47.6] \\
 & Ukrainian & 70.0\% [60.4-78.1] & 61.2\% [59.0-63.3] &  & Ukrainian & 12.0\% [7.0-19.8] & 13.3\% [11.9-14.9] \\
 & Dutch & 72.0\% [62.5-79.9] & 64.3\% [62.2-66.4] &  & Dutch & 42.0\% [32.8-51.8] & 29.4\% [27.4-31.4] \\
 & Danish & 65.5\% [58.7-71.7] & 61.1\% [59.6-62.6] &  & Danish & 28.0\% [22.2-34.6] & 21.5\% [20.3-22.8] \\
 & Estonian & 45.5\% [38.7-52.4] & 40.4\% [38.9-42.0] &  & Estonian & 5.0\% [2.2-11.2] & 3.0\% [2.4-3.9] \\
 & Icelandic & 22.0\% [15.0-31.1] & 23.4\% [21.5-25.3] &  & Icelandic & 5.0\% [2.2-11.2] & 1.1\% [0.7-1.6] \\
\addlinespace[2pt]
Qwen3-4B (reasoning off) & Chinese & 87.0\% [79.0-92.2] & 86.1\% [84.5-87.5] & OLMo-2-1124-13B-Instruct & Chinese & 63.0\% [53.2-71.8] & 49.9\% [47.7-52.1] \\
 & Hindi & 79.0\% [70.0-85.8] & 77.6\% [75.7-79.4] &  & Hindi & 24.0\% [16.7-33.2] & 14.1\% [12.6-15.6] \\
 & English & 95.0\% [88.8-97.8] & 92.8\% [91.6-93.9] &  & English & 91.0\% [83.8-95.2] & 82.9\% [81.2-84.5] \\
 & English metric & 90.0\% [82.6-94.5] & 92.6\% [91.4-93.7] &  & English metric & 89.0\% [81.4-93.7] & 82.5\% [80.7-84.1] \\
 & Arabic & 86.0\% [77.9-91.5] & 81.8\% [80.0-83.4] &  & Arabic & 46.0\% [36.6-55.7] & 37.0\% [34.9-39.1] \\
 & Japanese & 83.5\% [77.7-88.0] & 80.8\% [79.6-82.0] &  & Japanese & 42.5\% [35.9-49.4] & 34.2\% [32.8-35.7] \\
 & Russian & 86.0\% [77.9-91.5] & 84.0\% [82.3-85.5] &  & Russian & 74.0\% [64.6-81.6] & 59.5\% [57.3-61.6] \\
 & German & 88.5\% [83.3-92.2] & 84.7\% [83.5-85.7] &  & German & 71.0\% [64.4-76.8] & 68.4\% [67.0-69.8] \\
 & Marathi & 79.0\% [70.0-85.8] & 67.3\% [65.3-69.4] &  & Marathi & 2.0\% [0.6-7.0] & 2.9\% [2.3-3.7] \\
 & French & 85.0\% [76.7-90.7] & 87.1\% [85.5-88.5] &  & French & 76.0\% [66.8-83.3] & 69.5\% [67.4-71.4] \\
 & Italian & 85.0\% [76.7-90.7] & 86.4\% [84.8-87.8] &  & Italian & 79.0\% [70.0-85.8] & 68.6\% [66.5-70.6] \\
 & Ukrainian & 87.0\% [79.0-92.2] & 84.2\% [82.6-85.8] &  & Ukrainian & 44.0\% [34.7-53.8] & 34.0\% [32.0-36.1] \\
 & Dutch & 83.0\% [74.5-89.1] & 85.2\% [83.5-86.6] &  & Dutch & 64.0\% [54.2-72.7] & 56.4\% [54.2-58.6] \\
 & Danish & 88.5\% [83.3-92.2] & 84.6\% [83.4-85.7] &  & Danish & 56.5\% [49.6-63.2] & 49.8\% [48.3-51.3] \\
 & Estonian & 71.0\% [64.4-76.8] & 69.9\% [68.4-71.3] &  & Estonian & 14.0\% [8.5-22.1] & 13.7\% [12.2-15.2] \\
 & Icelandic & 54.0\% [44.3-63.4] & 56.7\% [54.5-58.9] &  & Icelandic & 23.0\% [15.8-32.2] & 17.8\% [16.1-19.5] \\
\addlinespace[2pt]
Qwen3-8B (reasoning off) & Chinese & 88.0\% [80.2-93.0] & 87.1\% [85.5-88.5] & OLMo-2-0325-32B-Instruct & Chinese & 79.0\% [70.0-85.8] & 73.7\% [71.7-75.6] \\
 & Hindi & 84.0\% [75.6-89.9] & 81.7\% [79.9-83.3] &  & Hindi & 58.0\% [48.2-67.2] & 50.1\% [48.0-52.3] \\
 & English & 95.0\% [88.8-97.8] & 94.3\% [93.2-95.2] &  & English & 94.0\% [87.5-97.2] & 90.1\% [88.8-91.4] \\
 & English metric & 93.0\% [86.3-96.6] & 93.5\% [92.3-94.5] &  & English metric & 92.0\% [85.0-95.9] & 89.0\% [87.6-90.3] \\
 & Arabic & 87.0\% [79.0-92.2] & 87.0\% [85.5-88.4] &  & Arabic & 55.0\% [45.2-64.4] & 50.9\% [48.7-53.1] \\
 & Japanese & 92.0\% [87.4-95.0] & 87.2\% [86.2-88.2] &  & Japanese & 57.5\% [50.6-64.1] & 46.8\% [45.3-48.4] \\
 & Russian & 91.0\% [83.8-95.2] & 85.1\% [83.5-86.6] &  & Russian & 64.0\% [54.2-72.7] & 52.0\% [49.9-54.2] \\
 & German & 88.5\% [83.3-92.2] & 87.3\% [86.2-88.3] &  & German & 76.5\% [70.2-81.8] & 72.3\% [70.9-73.7] \\
 & Marathi & 83.0\% [74.5-89.1] & 77.2\% [75.3-79.0] &  & Marathi & 46.0\% [36.6-55.7] & 40.6\% [38.5-42.8] \\
 & French & 89.0\% [81.4-93.7] & 89.3\% [87.9-90.6] &  & French & 85.0\% [76.7-90.7] & 75.6\% [73.7-77.5] \\
 & Italian & 90.0\% [82.6-94.5] & 90.8\% [89.4-91.9] &  & Italian & 82.0\% [73.3-88.3] & 71.7\% [69.7-73.6] \\
 & Ukrainian & 87.0\% [79.0-92.2] & 88.3\% [86.9-89.7] &  & Ukrainian & 59.0\% [49.2-68.1] & 48.4\% [46.3-50.6] \\
 & Dutch & 90.0\% [82.6-94.5] & 87.2\% [85.6-88.5] &  & Dutch & 74.0\% [64.6-81.6] & 69.9\% [67.9-71.9] \\
 & Danish & 91.5\% [86.8-94.6] & 87.7\% [86.7-88.7] &  & Danish & 79.5\% [73.4-84.5] & 73.0\% [71.6-74.3] \\
 & Estonian & 81.0\% [75.0-85.8] & 78.8\% [77.5-80.0] &  & Estonian & 26.0\% [18.4-35.4] & 24.9\% [23.0-26.8] \\
 & Icelandic & 76.0\% [66.8-83.3] & 70.9\% [68.8-72.8] &  & Icelandic & 64.0\% [54.2-72.7] & 55.9\% [53.7-58.0] \\
\addlinespace[2pt]
Qwen3-14B (reasoning off) & Chinese & 89.0\% [81.4-93.7] & 91.1\% [89.8-92.3] & Olmo-3-7B-Instruct & Chinese & 76.0\% [66.8-83.3] & 69.4\% [67.3-71.4] \\
 & Hindi & 86.0\% [77.9-91.5] & 85.7\% [84.1-87.2] &  & Hindi & 31.0\% [22.8-40.6] & 20.8\% [19.1-22.7] \\
 & English & 96.0\% [90.2-98.4] & 96.5\% [95.5-97.2] &  & English & 94.0\% [87.5-97.2] & 91.8\% [90.5-92.9] \\
 & English metric & 95.0\% [88.8-97.8] & 95.8\% [94.8-96.5] &  & English metric & 90.0\% [82.6-94.5] & 91.5\% [90.2-92.6] \\
 & Arabic & 91.0\% [83.8-95.2] & 91.1\% [89.8-92.3] &  & Arabic & 56.0\% [46.2-65.3] & 45.9\% [43.7-48.1] \\
 & Japanese & 90.0\% [85.1-93.4] & 88.1\% [87.1-89.1] &  & Japanese & 64.0\% [57.1-70.3] & 56.0\% [54.4-57.5] \\
 & Russian & 94.0\% [87.5-97.2] & 89.2\% [87.8-90.5] &  & Russian & 78.0\% [68.9-85.0] & 70.5\% [68.5-72.5] \\
 & German & 93.0\% [88.6-95.8] & 91.3\% [90.4-92.1] &  & German & 72.0\% [65.4-77.8] & 68.8\% [67.3-70.2] \\
 & Marathi & 86.0\% [77.9-91.5] & 83.8\% [82.1-85.3] &  & Marathi & 20.0\% [13.3-28.9] & 10.6\% [9.3-12.0] \\
 & French & 95.0\% [88.8-97.8] & 90.8\% [89.5-92.0] &  & French & 79.0\% [70.0-85.8] & 76.1\% [74.2-77.9] \\
 & Italian & 94.0\% [87.5-97.2] & 92.5\% [91.3-93.6] &  & Italian & 74.0\% [64.6-81.6] & 71.1\% [69.1-73.0] \\
 & Ukrainian & 95.0\% [88.8-97.8] & 92.3\% [91.0-93.4] &  & Ukrainian & 51.0\% [41.3-60.6] & 51.0\% [48.9-53.2] \\
 & Dutch & 89.0\% [81.4-93.7] & 89.2\% [87.8-90.5] &  & Dutch & 73.0\% [63.6-80.7] & 63.0\% [60.9-65.1] \\
 & Danish & 95.5\% [91.7-97.6] & 92.5\% [91.6-93.3] &  & Danish & 56.0\% [49.1-62.7] & 52.0\% [50.4-53.5] \\
 & Estonian & 85.5\% [80.0-89.7] & 86.1\% [85.0-87.1] &  & Estonian & 17.0\% [10.9-25.5] & 15.3\% [13.8-17.0] \\
 & Icelandic & 82.0\% [73.3-88.3] & 80.8\% [79.0-82.4] &  & Icelandic & 20.0\% [13.3-28.9] & 17.5\% [15.9-19.2] \\
\addlinespace[2pt]
Qwen3-32B (reasoning off) & Chinese & 95.0\% [88.8-97.8] & 92.5\% [91.3-93.6] & Olmo-3.1-32B-Instruct & Chinese & 89.0\% [81.4-93.7] & 86.9\% [85.4-88.3] \\
 & Hindi & 89.0\% [81.4-93.7] & 89.3\% [87.9-90.6] &  & Hindi & 68.0\% [58.3-76.3] & 66.6\% [64.6-68.7] \\
 & English & 97.0\% [91.5-99.0] & 96.1\% [95.2-96.9] &  & English & 96.0\% [90.2-98.4] & 96.4\% [95.5-97.1] \\
 & English metric & 96.0\% [90.2-98.4] & 95.5\% [94.5-96.3] &  & English metric & 95.0\% [88.8-97.8] & 95.7\% [94.7-96.5] \\
 & Arabic & 94.0\% [87.5-97.2] & 94.0\% [92.8-94.9] &  & Arabic & 67.0\% [57.3-75.4] & 69.8\% [67.8-71.8] \\
 & Japanese & 93.5\% [89.2-96.2] & 91.3\% [90.4-92.2] &  & Japanese & 83.5\% [77.7-88.0] & 82.5\% [81.3-83.6] \\
 & Russian & 92.0\% [85.0-95.9] & 88.5\% [87.1-89.9] &  & Russian & 88.0\% [80.2-93.0] & 85.2\% [83.6-86.7] \\
 & German & 94.0\% [89.8-96.5] & 90.6\% [89.6-91.4] &  & German & 91.0\% [86.2-94.2] & 87.6\% [86.5-88.6] \\
 & Marathi & 88.0\% [80.2-93.0] & 87.7\% [86.2-89.1] &  & Marathi & 50.0\% [40.4-59.6] & 43.7\% [41.5-45.9] \\
 & French & 93.0\% [86.3-96.6] & 90.7\% [89.3-91.9] &  & French & 86.0\% [77.9-91.5] & 84.0\% [82.4-85.6] \\
 & Italian & 96.0\% [90.2-98.4] & 92.2\% [90.9-93.3] &  & Italian & 90.0\% [82.6-94.5] & 89.5\% [88.0-90.7] \\
 & Ukrainian & 95.0\% [88.8-97.8] & 92.7\% [91.4-93.7] &  & Ukrainian & 86.0\% [77.9-91.5] & 83.8\% [82.1-85.3] \\
 & Dutch & 94.0\% [87.5-97.2] & 91.2\% [89.9-92.4] &  & Dutch & 85.0\% [76.7-90.7] & 84.2\% [82.5-85.7] \\
 & Danish & 94.0\% [89.8-96.5] & 92.8\% [91.9-93.5] &  & Danish & 86.5\% [81.1-90.6] & 84.9\% [83.8-86.0] \\
 & Estonian & 91.0\% [83.8-95.2] & 90.0\% [88.6-91.2] &  & Estonian & 26.0\% [18.4-35.4] & 19.3\% [17.6-21.1] \\
 & Icelandic & 86.0\% [77.9-91.5] & 85.5\% [83.9-87.0] &  & Icelandic & 65.0\% [55.3-73.6] & 57.4\% [55.2-59.6] \\
\addlinespace[2pt]
Qwen3-0.6B & Chinese & 63.0\% [53.2-71.8] & 62.4\% [60.3-64.5] & Olmo-3-7B-Think (reasoning on) & Chinese & 88.0\% [80.2-93.0] & 86.9\% [85.3-88.3] \\
 & Hindi & 29.0\% [21.0-38.5] & 28.6\% [26.7-30.7] &  & Hindi & 71.0\% [61.5-79.0] & 66.1\% [64.0-68.1] \\
 & English & 83.0\% [74.5-89.1] & 77.3\% [75.4-79.1] &  & English & 93.0\% [86.3-96.6] & 95.5\% [94.4-96.3] \\
 & English metric & 74.0\% [64.6-81.6] & 76.6\% [74.7-78.5] &  & English metric & 95.0\% [88.8-97.8] & 94.7\% [93.6-95.6] \\
 & Arabic & 47.0\% [37.5-56.7] & 43.2\% [41.0-45.4] &  & Arabic & 80.0\% [71.1-86.7] & 75.6\% [73.7-77.5] \\
 & Japanese & 58.0\% [51.1-64.6] & 52.1\% [50.6-53.7] &  & Japanese & 87.5\% [82.2-91.4] & 80.2\% [78.9-81.4] \\
 & Russian & 56.0\% [46.2-65.3] & 51.8\% [49.7-54.0] &  & Russian & 87.0\% [79.0-92.2] & 79.7\% [77.8-81.4] \\
 & German & 63.5\% [56.6-69.9] & 60.1\% [58.6-61.6] &  & German & 90.0\% [85.1-93.4] & 88.2\% [87.2-89.2] \\
 & Marathi & 15.0\% [9.3-23.3] & 12.4\% [11.1-14.0] &  & Marathi & 60.0\% [50.2-69.1] & 44.9\% [42.7-47.0] \\
 & French & 70.0\% [60.4-78.1] & 61.6\% [59.4-63.7] &  & French & 92.0\% [85.0-95.9] & 89.4\% [88.0-90.7] \\
 & Italian & 66.0\% [56.3-74.5] & 60.8\% [58.6-62.9] &  & Italian & 93.0\% [86.3-96.6] & 89.5\% [88.1-90.8] \\
 & Ukrainian & 47.0\% [37.5-56.7] & 45.0\% [42.8-47.2] &  & Ukrainian & 76.0\% [66.8-83.3] & 77.5\% [75.7-79.3] \\
 & Dutch & 66.0\% [56.3-74.5] & 58.8\% [56.6-60.9] &  & Dutch & 86.0\% [77.9-91.5] & 85.0\% [83.4-86.5] \\
 & Danish & 36.5\% [30.1-43.4] & 48.0\% [46.4-49.5] &  & Danish & 77.5\% [71.2-82.7] & 74.8\% [73.4-76.1] \\
 & Estonian & 22.5\% [17.3-28.8] & 19.5\% [18.3-20.7] &  & Estonian & 29.0\% [21.0-38.5] & 29.4\% [27.5-31.5] \\
 & Icelandic & 14.0\% [8.5-22.1] & 14.1\% [12.6-15.6] &  & Icelandic & 53.0\% [43.3-62.5] & 42.2\% [40.1-44.4] \\
\addlinespace[2pt]
Qwen3-1.7B & Chinese & 81.0\% [72.2-87.5] & 86.8\% [85.2-88.2] & Olmo-3-32B-Think (reasoning on) & Chinese & 95.0\% [88.8-97.8] & 91.8\% [90.5-92.9] \\
 & Hindi & 75.0\% [65.7-82.5] & 69.2\% [67.2-71.2] &  & Hindi & 84.0\% [75.6-89.9] & 84.5\% [82.9-86.1] \\
 & English & 92.0\% [85.0-95.9] & 90.3\% [89.0-91.6] &  & English & 96.0\% [90.2-98.4] & 96.8\% [95.9-97.4] \\
 & English metric & 89.0\% [81.4-93.7] & 90.2\% [88.9-91.5] &  & English metric & 96.0\% [90.2-98.4] & 96.2\% [95.3-97.0] \\
 & Arabic & 78.0\% [68.9-85.0] & 75.9\% [74.0-77.8] &  & Arabic & 89.0\% [81.4-93.7] & 85.9\% [84.3-87.3] \\
 & Japanese & 82.0\% [76.1-86.7] & 79.1\% [77.9-80.4] &  & Japanese & 91.5\% [86.8-94.6] & 90.3\% [89.3-91.2] \\
 & Russian & 85.0\% [76.7-90.7] & 77.2\% [75.4-79.0] &  & Russian & 93.0\% [86.3-96.6] & 88.8\% [87.3-90.1] \\
 & German & 85.0\% [79.4-89.3] & 82.5\% [81.3-83.7] &  & German & 92.5\% [88.0-95.4] & 91.0\% [90.0-91.8] \\
 & Marathi & 63.0\% [53.2-71.8] & 56.6\% [54.4-58.8] &  & Marathi & 71.0\% [61.5-79.0] & 69.7\% [67.6-71.6] \\
 & French & 88.0\% [80.2-93.0] & 82.2\% [80.4-83.8] &  & French & 94.0\% [87.5-97.2] & 91.6\% [90.3-92.7] \\
 & Italian & 84.0\% [75.6-89.9] & 84.9\% [83.3-86.4] &  & Italian & 94.0\% [87.5-97.2] & 93.2\% [92.1-94.3] \\
 & Ukrainian & 78.0\% [68.9-85.0] & 82.3\% [80.6-84.0] &  & Ukrainian & 91.0\% [83.8-95.2] & 88.9\% [87.4-90.2] \\
 & Dutch & 87.0\% [79.0-92.2] & 82.2\% [80.5-83.8] &  & Dutch & 93.0\% [86.3-96.6] & 91.5\% [90.1-92.6] \\
 & Danish & 72.5\% [65.9-78.2] & 79.3\% [78.0-80.6] &  & Danish & 93.0\% [88.6-95.8] & 92.3\% [91.4-93.1] \\
 & Estonian & 72.5\% [65.9-78.2] & 64.3\% [62.8-65.8] &  & Estonian & 77.0\% [67.8-84.2] & 70.2\% [68.2-72.2] \\
 & Icelandic & 44.0\% [34.7-53.8] & 41.8\% [39.6-43.9] &  & Icelandic & 87.0\% [79.0-92.2] & 78.9\% [77.1-80.6] \\
\addlinespace[2pt]
Qwen3-4B & Chinese & 89.0\% [81.4-93.7] & 89.2\% [87.8-90.5] & granite-3.2-2b-instruct (reasoning off) & Chinese & 39.0\% [30.0-48.8] & 31.9\% [29.9-34.0] \\
 & Hindi & 85.0\% [76.7-90.7] & 86.6\% [85.0-88.0] &  & Hindi & 4.0\% [1.6-9.8] & 4.3\% [3.5-5.3] \\
 & English & 95.0\% [88.8-97.8] & 94.8\% [93.7-95.6] &  & English & 68.0\% [58.3-76.3] & 62.1\% [60.0-64.2] \\
 & English metric & 96.0\% [90.2-98.4] & 94.4\% [93.3-95.3] &  & English metric & 66.0\% [56.3-74.5] & 62.0\% [59.8-64.1] \\
 & Arabic & 90.0\% [82.6-94.5] & 89.9\% [88.5-91.1] &  & Arabic & 15.0\% [9.3-23.3] & 12.1\% [10.7-13.6] \\
 & Japanese & 93.7\% [90.3-95.9] & 90.6\% [89.8-91.3] &  & Japanese & 20.0\% [15.0-26.1] & 18.2\% [17.0-19.4] \\
 & Russian & 91.0\% [83.8-95.2] & 88.2\% [86.7-89.5] &  & Russian & 32.0\% [23.7-41.7] & 21.8\% [20.0-23.7] \\
 & German & 89.3\% [85.3-92.3] & 90.1\% [89.3-90.8] &  & German & 44.0\% [37.3-50.9] & 42.4\% [40.9-43.9] \\
 & Marathi & 80.0\% [71.1-86.7] & 79.5\% [77.7-81.2] &  & Marathi & 1.0\% [0.2-5.4] & 2.5\% [1.9-3.3] \\
 & French & 92.0\% [85.0-95.9] & 90.5\% [89.1-91.7] &  & French & 54.0\% [44.3-63.4] & 47.8\% [45.6-50.0] \\
 & Italian & 91.0\% [83.8-95.2] & 91.3\% [90.0-92.5] &  & Italian & 49.0\% [39.4-58.7] & 49.0\% [46.9-51.2] \\
 & Ukrainian & 92.0\% [85.0-95.9] & 89.3\% [87.9-90.6] &  & Ukrainian & 11.0\% [6.3-18.6] & 10.5\% [9.2-11.9] \\
 & Dutch & 92.0\% [85.0-95.9] & 89.7\% [88.3-91.0] &  & Dutch & 45.0\% [35.6-54.8] & 40.8\% [38.6-42.9] \\
 & Danish & 91.0\% [87.2-93.7] & 91.0\% [90.1-91.9] &  & Danish & 18.0\% [13.3-23.9] & 14.1\% [13.1-15.2] \\
 & Estonian & 82.0\% [77.3-85.9] & 82.4\% [81.2-83.5] &  & Estonian & 4.0\% [1.6-9.8] & 1.1\% [0.7-1.7] \\
 & Icelandic & 82.0\% [73.3-88.3] & 74.0\% [72.0-75.9] &  & Icelandic & 2.0\% [0.6-7.0] & 1.1\% [0.7-1.7] \\
\addlinespace[2pt]
Qwen3-8B & Chinese & 90.0\% [82.6-94.5] & 90.3\% [88.9-91.5] & granite-3.2-8b-instruct (reasoning off) & Chinese & 58.0\% [48.2-67.2] & 54.6\% [52.5-56.8] \\
 & Hindi & 86.0\% [77.9-91.5] & 89.8\% [88.4-91.1] &  & Hindi & 21.0\% [14.2-30.0] & 22.1\% [20.3-23.9] \\
 & English & 95.0\% [88.8-97.8] & 95.5\% [94.6-96.4] &  & English & 82.0\% [73.3-88.3] & 82.6\% [80.9-84.2] \\
 & English metric & 95.0\% [88.8-97.8] & 95.0\% [94.0-95.9] &  & English metric & 86.0\% [77.9-91.5] & 81.5\% [79.8-83.2] \\
 & Arabic & 88.0\% [80.2-93.0] & 90.0\% [88.7-91.3] &  & Arabic & 47.0\% [37.5-56.7] & 37.1\% [35.1-39.3] \\
 & Japanese & 94.5\% [90.4-96.9] & 90.5\% [89.6-91.4] &  & Japanese & 54.0\% [47.1-60.8] & 43.5\% [42.0-45.0] \\
 & Russian & 92.0\% [85.0-95.9] & 87.9\% [86.4-89.3] &  & Russian & 57.0\% [47.2-66.3] & 46.2\% [44.0-48.4] \\
 & German & 92.5\% [88.0-95.4] & 91.1\% [90.2-91.9] &  & German & 72.0\% [65.4-77.8] & 64.9\% [63.4-66.3] \\
 & Marathi & 87.0\% [79.0-92.2] & 85.7\% [84.1-87.2] &  & Marathi & 15.0\% [9.3-23.3] & 13.6\% [12.1-15.1] \\
 & French & 93.0\% [86.3-96.6] & 90.3\% [88.9-91.5] &  & French & 74.0\% [64.6-81.6] & 66.2\% [64.1-68.3] \\
 & Italian & 96.0\% [90.2-98.4] & 92.5\% [91.3-93.6] &  & Italian & 79.0\% [70.0-85.8] & 69.3\% [67.2-71.3] \\
 & Ukrainian & 93.0\% [86.3-96.6] & 92.2\% [91.0-93.3] &  & Ukrainian & 45.0\% [35.6-54.8] & 35.4\% [33.3-37.5] \\
 & Dutch & 92.0\% [85.0-95.9] & 90.2\% [88.9-91.5] &  & Dutch & 69.0\% [59.4-77.2] & 63.5\% [61.4-65.6] \\
 & Danish & 92.5\% [88.0-95.4] & 92.1\% [91.2-92.9] &  & Danish & 51.0\% [44.1-57.8] & 43.9\% [42.3-45.4] \\
 & Estonian & 86.5\% [81.1-90.6] & 86.7\% [85.6-87.7] &  & Estonian & 12.0\% [7.0-19.8] & 7.3\% [6.2-8.5] \\
 & Icelandic & 88.0\% [80.2-93.0] & 84.0\% [82.4-85.6] &  & Icelandic & 26.0\% [18.4-35.4] & 15.2\% [13.6-16.8] \\
\addlinespace[2pt]
Qwen3-14B & Chinese & 95.0\% [88.8-97.8] & 94.2\% [93.0-95.1] & granite-3.2-2b-instruct (reasoning on) & Chinese & 31.0\% [22.8-40.6] & 32.2\% [30.2-34.3] \\
 & Hindi & 85.0\% [76.7-90.7] & 88.8\% [87.3-90.1] &  & Hindi & 17.0\% [10.9-25.5] & 12.2\% [10.9-13.8] \\
 & English & 96.0\% [90.2-98.4] & 96.2\% [95.3-97.0] &  & English & 69.0\% [59.4-77.2] & 60.2\% [58.0-62.3] \\
 & English metric & 97.0\% [91.5-99.0] & 95.7\% [94.7-96.5] &  & English metric & 67.0\% [57.3-75.4] & 59.8\% [57.6-61.9] \\
 & Arabic & 92.0\% [85.0-95.9] & 93.1\% [91.9-94.1] &  & Arabic & 31.0\% [22.8-40.6] & 22.7\% [20.9-24.5] \\
 & Japanese & 94.5\% [90.4-96.9] & 91.3\% [90.4-92.2] &  & Japanese & 23.0\% [17.7-29.3] & 18.5\% [17.3-19.7] \\
 & Russian & 93.0\% [86.3-96.6] & 89.8\% [88.3-91.0] &  & Russian & 25.0\% [17.5-34.3] & 23.1\% [21.3-24.9] \\
 & German & 93.0\% [88.6-95.8] & 92.0\% [91.2-92.8] &  & German & 49.5\% [42.6-56.4] & 43.8\% [42.2-45.3] \\
 & Marathi & 93.0\% [86.3-96.6] & 90.2\% [88.9-91.5] &  & Marathi & 10.0\% [5.5-17.4] & 8.5\% [7.4-9.8] \\
 & French & 94.0\% [87.5-97.2] & 91.0\% [89.6-92.1] &  & French & 47.0\% [37.5-56.7] & 45.8\% [43.6-47.9] \\
 & Italian & 95.0\% [88.8-97.8] & 93.2\% [92.0-94.2] &  & Italian & 52.0\% [42.3-61.5] & 45.6\% [43.5-47.8] \\
 & Ukrainian & 94.0\% [87.5-97.2] & 92.2\% [90.9-93.2] &  & Ukrainian & 26.0\% [18.4-35.4] & 16.1\% [14.5-17.7] \\
 & Dutch & 93.0\% [86.3-96.6] & 92.0\% [90.7-93.1] &  & Dutch & 47.0\% [37.5-56.7] & 41.2\% [39.1-43.4] \\
 & Danish & 96.5\% [93.0-98.3] & 93.8\% [93.0-94.5] &  & Danish & 20.5\% [15.5-26.6] & 15.9\% [14.8-17.1] \\
 & Estonian & 91.0\% [86.2-94.2] & 91.5\% [90.6-92.4] &  & Estonian & 2.0\% [0.6-7.0] & 1.5\% [1.1-2.1] \\
 & Icelandic & 93.0\% [86.3-96.6] & 88.4\% [86.9-89.7] &  & Icelandic & 5.0\% [2.2-11.2] & 2.1\% [1.6-2.9] \\
\addlinespace[2pt]
Qwen3-32B & Chinese & 95.0\% [88.8-97.8] & 95.5\% [94.6-96.4] & granite-3.2-8b-instruct (reasoning on) & Chinese & 68.0\% [58.3-76.3] & 63.3\% [61.2-65.4] \\
 & Hindi & 92.0\% [85.0-95.9] & 93.2\% [92.0-94.2] &  & Hindi & 49.0\% [39.4-58.7] & 40.6\% [38.5-42.8] \\
 & English & 97.0\% [91.5-99.0] & 97.5\% [96.7-98.1] &  & English & 83.0\% [74.5-89.1] & 82.0\% [80.3-83.7] \\
 & English metric & 96.0\% [90.2-98.4] & 97.0\% [96.2-97.7] &  & English metric & 85.0\% [76.7-90.7] & 81.5\% [79.8-83.2] \\
 & Arabic & 95.0\% [88.8-97.8] & 95.2\% [94.1-96.0] &  & Arabic & 46.0\% [36.6-55.7] & 48.5\% [46.3-50.7] \\
 & Japanese & 94.5\% [90.4-96.9] & 92.2\% [91.4-93.0] &  & Japanese & 54.0\% [47.1-60.8] & 51.0\% [49.5-52.5] \\
 & Russian & 95.0\% [88.8-97.8] & 91.3\% [90.0-92.5] &  & Russian & 74.0\% [64.6-81.6] & 62.8\% [60.7-64.9] \\
 & German & 94.5\% [90.4-96.9] & 92.6\% [91.7-93.4] &  & German & 82.0\% [76.1-86.7] & 74.6\% [73.2-75.9] \\
 & Marathi & 90.0\% [82.6-94.5] & 92.3\% [91.0-93.4] &  & Marathi & 39.0\% [30.0-48.8] & 30.3\% [28.4-32.4] \\
 & French & 94.0\% [87.5-97.2] & 91.7\% [90.4-92.8] &  & French & 78.0\% [68.9-85.0] & 73.6\% [71.6-75.4] \\
 & Italian & 97.0\% [91.5-99.0] & 94.7\% [93.6-95.6] &  & Italian & 81.0\% [72.2-87.5] & 75.5\% [73.6-77.4] \\
 & Ukrainian & 96.0\% [90.2-98.4] & 93.8\% [92.7-94.8] &  & Ukrainian & 62.0\% [52.2-70.9] & 55.8\% [53.6-58.0] \\
 & Dutch & 95.0\% [88.8-97.8] & 92.7\% [91.4-93.7] &  & Dutch & 82.0\% [73.3-88.3] & 70.7\% [68.6-72.6] \\
 & Danish & 95.5\% [91.7-97.6] & 94.1\% [93.3-94.8] &  & Danish & 60.5\% [53.6-67.0] & 57.5\% [56.0-59.0] \\
 & Estonian & 89.5\% [84.5-93.0] & 91.7\% [90.4-92.8] &  & Estonian & 24.0\% [16.7-33.2] & 18.2\% [16.6-20.0] \\
 & Icelandic & 92.0\% [85.0-95.9] & 89.5\% [88.1-90.8] &  & Icelandic & 28.0\% [20.1-37.5] & 26.2\% [24.4-28.2] \\
\addlinespace[2pt]
Qwen3.5-0.8B (reasoning off) & Chinese & 33.0\% [24.6-42.7] & 25.9\% [24.0-27.8] & EuroLLM-1.7B-Instruct & Chinese & 0.0\% [0.0-3.7] & 0.2\% [0.1-0.6] \\
 & Hindi & 0.0\% [0.0-3.7] & 0.3\% [0.1-0.7] &  & Hindi & 0.0\% [0.0-3.7] & 0.4\% [0.2-0.7] \\
 & English & 2.0\% [0.6-7.0] & 44.8\% [42.6-46.9] &  & English & 1.0\% [0.2-5.4] & 0.6\% [0.3-1.0] \\
 & English metric & 54.0\% [44.3-63.4] & 45.1\% [43.0-47.3] &  & English metric & 1.0\% [0.2-5.4] & 0.5\% [0.3-0.9] \\
 & Arabic & 8.0\% [4.1-15.0] & 7.4\% [6.3-8.6] &  & Arabic & 0.0\% [0.0-3.7] & 0.1\% [0.1-0.4] \\
 & Japanese & 12.5\% [8.6-17.8] & 10.1\% [9.2-11.0] &  & Japanese & 0.5\% [0.1-2.8] & 0.5\% [0.3-0.7] \\
 & Russian & 15.0\% [9.3-23.3] & 11.2\% [9.9-12.7] &  & Russian & 0.0\% [0.0-3.7] & 0.2\% [0.1-0.5] \\
 & German & 11.0\% [7.4-16.1] & 10.3\% [9.4-11.3] &  & German & 0.5\% [0.1-2.8] & 0.2\% [0.1-0.4] \\
 & Marathi & 1.0\% [0.2-5.4] & 0.2\% [0.1-0.6] &  & Marathi & 1.0\% [0.2-5.4] & 0.5\% [0.3-1.0] \\
 & French & 19.0\% [12.5-27.8] & 13.9\% [12.5-15.5] &  & French & 0.0\% [0.0-3.7] & 0.4\% [0.2-0.7] \\
 & Italian & 14.0\% [8.5-22.1] & 13.1\% [11.7-14.6] &  & Italian & 0.0\% [0.0-3.7] & 0.3\% [0.1-0.7] \\
 & Ukrainian & 3.0\% [1.0-8.5] & 5.1\% [4.3-6.2] &  & Ukrainian & 0.0\% [0.0-3.7] & 0.4\% [0.2-0.8] \\
 & Dutch & 15.0\% [9.3-23.3] & 10.1\% [8.9-11.5] &  & Dutch & 0.0\% [0.0-3.7] & 0.4\% [0.2-0.9] \\
 & Danish & 5.0\% [2.7-9.0] & 4.9\% [4.3-5.6] &  & Danish & 0.5\% [0.1-2.8] & 0.4\% [0.2-0.6] \\
 & Estonian & 5.0\% [2.2-11.2] & 2.1\% [1.6-2.8] &  & Estonian & 1.0\% [0.2-5.4] & 0.5\% [0.3-0.9] \\
 & Icelandic & 6.0\% [2.8-12.5] & 1.7\% [1.2-2.4] &  & Icelandic & 1.0\% [0.2-5.4] & 0.6\% [0.3-1.0] \\
\addlinespace[2pt]
Qwen3.5-4B (reasoning off) & Chinese & 95.0\% [88.8-97.8] & 90.0\% [88.6-91.2] & EuroLLM-9B-Instruct-2512 & Chinese & 50.0\% [40.4-59.6] & 50.3\% [48.1-52.5] \\
 & Hindi & 52.0\% [42.3-61.5] & 47.9\% [45.8-50.1] &  & Hindi & 50.0\% [40.4-59.6] & 41.7\% [39.6-43.9] \\
 & English & 98.0\% [93.0-99.4] & 93.7\% [92.5-94.6] &  & English & 54.0\% [44.3-63.4] & 51.2\% [49.1-53.4] \\
 & English metric & 94.0\% [87.5-97.2] & 93.0\% [91.8-94.0] &  & English metric & 58.0\% [48.2-67.2] & 51.2\% [49.0-53.4] \\
 & Arabic & 84.0\% [75.6-89.9] & 82.0\% [80.3-83.6] &  & Arabic & 57.0\% [47.2-66.3] & 48.5\% [46.4-50.7] \\
 & Japanese & 87.5\% [82.2-91.4] & 82.9\% [81.7-84.0] &  & Japanese & 39.0\% [32.5-45.9] & 35.1\% [33.6-36.6] \\
 & Russian & 88.0\% [80.2-93.0] & 84.2\% [82.5-85.7] &  & Russian & 56.0\% [46.2-65.3] & 50.6\% [48.4-52.8] \\
 & German & 90.0\% [85.1-93.4] & 87.5\% [86.5-88.5] &  & German & 62.5\% [55.6-68.9] & 55.5\% [53.9-57.0] \\
 & Marathi & 47.0\% [37.5-56.7] & 35.7\% [33.6-37.8] &  & Marathi & 1.0\% [0.2-5.4] & 1.2\% [0.8-1.8] \\
 & French & 93.0\% [86.3-96.6] & 87.0\% [85.5-88.4] &  & French & 63.0\% [53.2-71.8] & 56.1\% [54.0-58.3] \\
 & Italian & 93.0\% [86.3-96.6] & 91.2\% [89.9-92.4] &  & Italian & 66.0\% [56.3-74.5] & 59.2\% [57.0-61.3] \\
 & Ukrainian & 87.0\% [79.0-92.2] & 82.3\% [80.6-84.0] &  & Ukrainian & 62.0\% [52.2-70.9] & 53.4\% [51.3-55.6] \\
 & Dutch & 82.0\% [73.3-88.3] & 84.0\% [82.4-85.6] &  & Dutch & 61.0\% [51.2-70.0] & 53.5\% [51.3-55.7] \\
 & Danish & 88.0\% [82.8-91.8] & 84.9\% [83.7-85.9] &  & Danish & 60.5\% [53.6-67.0] & 53.7\% [52.1-55.2] \\
 & Estonian & 81.0\% [72.2-87.5] & 74.9\% [73.0-76.8] &  & Estonian & 53.0\% [43.3-62.5] & 46.0\% [43.8-48.1] \\
 & Icelandic & 69.0\% [59.4-77.2] & 63.7\% [61.6-65.8] &  & Icelandic & 10.0\% [5.5-17.4] & 3.7\% [3.0-4.6] \\
\addlinespace[2pt]
Qwen3.5-9B (reasoning off) & Chinese & 93.0\% [86.3-96.6] & 92.0\% [90.7-93.1] & EuroLLM-22B-Instruct-2512 & Chinese & 72.0\% [62.5-79.9] & 58.8\% [56.6-60.9] \\
 & Hindi & 73.0\% [63.6-80.7] & 73.0\% [71.1-74.9] &  & Hindi & 36.0\% [27.3-45.8] & 31.9\% [29.9-34.0] \\
 & English & 98.0\% [93.0-99.4] & 95.2\% [94.2-96.1] &  & English & 61.0\% [51.2-70.0] & 63.0\% [60.9-65.1] \\
 & English metric & 95.0\% [88.8-97.8] & 94.9\% [93.8-95.8] &  & English metric & 63.0\% [53.2-71.8] & 62.9\% [60.8-65.0] \\
 & Arabic & 89.0\% [81.4-93.7] & 90.0\% [88.6-91.2] &  & Arabic & 50.0\% [40.4-59.6] & 45.2\% [43.0-47.4] \\
 & Japanese & 90.5\% [85.6-93.8] & 88.6\% [87.6-89.5] &  & Japanese & 46.5\% [39.7-53.4] & 42.4\% [40.9-44.0] \\
 & Russian & 94.0\% [87.5-97.2] & 88.6\% [87.1-89.9] &  & Russian & 71.0\% [61.5-79.0] & 60.0\% [57.8-62.1] \\
 & German & 93.5\% [89.2-96.2] & 90.4\% [89.4-91.3] &  & German & 75.0\% [68.6-80.5] & 68.4\% [67.0-69.8] \\
 & Marathi & 65.0\% [55.3-73.6] & 50.6\% [48.5-52.8] &  & Marathi & 3.0\% [1.0-8.5] & 2.1\% [1.5-2.8] \\
 & French & 90.0\% [82.6-94.5] & 89.5\% [88.0-90.7] &  & French & 61.0\% [51.2-70.0] & 62.5\% [60.4-64.6] \\
 & Italian & 94.0\% [87.5-97.2] & 93.2\% [92.0-94.2] &  & Italian & 75.0\% [65.7-82.5] & 71.2\% [69.2-73.2] \\
 & Ukrainian & 92.0\% [85.0-95.9] & 89.1\% [87.7-90.4] &  & Ukrainian & 73.0\% [63.6-80.7] & 68.9\% [66.8-70.9] \\
 & Dutch & 90.0\% [82.6-94.5] & 88.1\% [86.6-89.4] &  & Dutch & 81.0\% [72.2-87.5] & 69.2\% [67.1-71.1] \\
 & Danish & 95.0\% [91.0-97.3] & 90.8\% [89.9-91.7] &  & Danish & 69.5\% [62.8-75.5] & 63.3\% [61.8-64.8] \\
 & Estonian & 91.0\% [83.8-95.2] & 89.7\% [88.3-91.0] &  & Estonian & 62.0\% [52.2-70.9] & 52.4\% [50.2-54.6] \\
 & Icelandic & 89.0\% [81.4-93.7] & 83.2\% [81.5-84.8] &  & Icelandic & 7.0\% [3.4-13.7] & 7.4\% [6.3-8.6] \\
\addlinespace[2pt]
Qwen3.5-27B (reasoning off) & Chinese & 97.0\% [91.5-99.0] & 97.2\% [96.3-97.8] & Apertus-8B-Instruct-2509 & Chinese & 29.0\% [21.0-38.5] & 23.5\% [21.7-25.4] \\
 & Hindi & 93.0\% [86.3-96.6] & 91.2\% [89.9-92.4] &  & Hindi & 30.0\% [21.9-39.6] & 23.2\% [21.4-25.1] \\
 & English & 98.0\% [93.0-99.4] & 97.9\% [97.1-98.4] &  & English & 32.0\% [23.7-41.7] & 25.2\% [23.3-27.1] \\
 & English metric & 98.0\% [93.0-99.4] & 97.5\% [96.7-98.1] &  & English metric & 31.0\% [22.8-40.6] & 23.8\% [22.0-25.8] \\
 & Arabic & 93.0\% [86.3-96.6] & 95.2\% [94.1-96.0] &  & Arabic & 34.0\% [25.5-43.7] & 28.4\% [26.5-30.5] \\
 & Japanese & 97.7\% [95.3-98.9] & 94.3\% [93.5-95.0] &  & Japanese & 21.5\% [16.4-27.7] & 16.0\% [14.8-17.1] \\
 & Russian & 95.0\% [88.8-97.8] & 91.2\% [89.9-92.4] &  & Russian & 56.0\% [46.2-65.3] & 41.8\% [39.6-43.9] \\
 & German & 97.0\% [94.4-98.4] & 93.5\% [92.6-94.2] &  & German & 45.0\% [38.3-51.9] & 44.0\% [42.5-45.6] \\
 & Marathi & 89.0\% [81.4-93.7] & 86.9\% [85.4-88.3] &  & Marathi & 23.0\% [15.8-32.2] & 17.9\% [16.3-19.6] \\
 & French & 97.0\% [91.5-99.0] & 93.5\% [92.3-94.5] &  & French & 34.0\% [25.5-43.7] & 34.2\% [32.1-36.3] \\
 & Italian & 96.0\% [90.2-98.4] & 95.5\% [94.4-96.3] &  & Italian & 46.0\% [36.6-55.7] & 45.2\% [43.0-47.4] \\
 & Ukrainian & 96.0\% [90.2-98.4] & 95.2\% [94.2-96.1] &  & Ukrainian & 39.0\% [30.0-48.8] & 36.4\% [34.3-38.5] \\
 & Dutch & 96.0\% [90.2-98.4] & 91.3\% [90.0-92.5] &  & Dutch & 40.0\% [30.9-49.8] & 31.5\% [29.5-33.6] \\
 & Danish & 98.0\% [95.7-99.1] & 96.2\% [95.6-96.8] &  & Danish & 34.0\% [27.8-40.8] & 30.6\% [29.1-32.0] \\
 & Estonian & 97.0\% [91.5-99.0] & 96.4\% [95.5-97.1] &  & Estonian & 33.0\% [24.6-42.7] & 32.6\% [30.5-34.6] \\
 & Icelandic & 95.0\% [88.8-97.8] & 92.8\% [91.6-93.9] &  & Icelandic & 30.0\% [21.9-39.6] & 28.1\% [26.1-30.1] \\
\addlinespace[2pt]
Qwen3.5-0.8B & Chinese & 1.0\% [0.2-5.4] & 1.2\% [0.8-1.8] & Apertus-70B-Instruct-2509 & Chinese & 70.0\% [60.4-78.1] & 63.1\% [61.0-65.2] \\
 & Hindi & 0.0\% [0.0-3.7] & 0.0\% [0.0-0.2] &  & Hindi & 55.0\% [45.2-64.4] & 54.1\% [51.9-56.3] \\
 & English & 2.0\% [0.6-7.0] & 1.4\% [1.0-2.0] &  & English & 23.0\% [15.8-32.2] & 22.7\% [20.9-24.6] \\
 & English metric & 3.0\% [1.0-8.5] & 1.3\% [0.9-1.9] &  & English metric & 28.0\% [20.1-37.5] & 23.3\% [21.5-25.2] \\
 & Arabic & 0.0\% [0.0-3.7] & 0.1\% [0.0-0.3] &  & Arabic & 64.0\% [54.2-72.7] & 60.2\% [58.0-62.3] \\
 & Japanese & 0.5\% [0.1-2.8] & 0.4\% [0.2-0.6] &  & Japanese & 53.8\% [48.9-58.6] & 49.6\% [48.3-50.8] \\
 & Russian & 0.0\% [0.0-3.7] & 0.1\% [0.0-0.3] &  & Russian & 72.0\% [62.5-79.9] & 64.3\% [62.2-66.4] \\
 & German & 0.0\% [0.0-1.9] & 0.1\% [0.0-0.2] &  & German & 74.5\% [70.0-78.5] & 65.6\% [64.1-67.1] \\
 & Marathi & 0.0\% [0.0-3.7] & 0.0\% [0.0-0.2] &  & Marathi & 47.0\% [37.5-56.7] & 37.8\% [35.7-39.9] \\
 & French & 0.0\% [0.0-3.7] & 0.4\% [0.2-0.7] &  & French & 76.0\% [66.8-83.3] & 66.1\% [64.0-68.1] \\
 & Italian & 0.0\% [0.0-3.7] & 0.1\% [0.0-0.3] &  & Italian & 72.0\% [62.5-79.9] & 66.4\% [64.3-68.4] \\
 & Ukrainian & 0.0\% [0.0-3.7] & 0.0\% [0.0-0.2] &  & Ukrainian & 74.0\% [64.6-81.6] & 61.2\% [59.0-63.3] \\
 & Dutch & 0.0\% [0.0-3.7] & 0.4\% [0.2-0.7] &  & Dutch & 59.0\% [49.2-68.1] & 59.7\% [57.5-61.8] \\
 & Danish & 2.0\% [0.8-5.0] & 0.1\% [0.1-0.3] &  & Danish & 55.2\% [50.4-60.0] & 59.7\% [58.2-61.2] \\
 & Estonian & 0.0\% [0.0-3.7] & 0.0\% [0.0-0.2] &  & Estonian & 63.0\% [53.2-71.8] & 56.1\% [53.9-58.3] \\
 & Icelandic & 0.0\% [0.0-3.7] & 0.0\% [0.0-0.2] &  & Icelandic & 57.0\% [47.2-66.3] & 55.3\% [53.1-57.5] \\
\end{longtable}
\endgroup

\end{document}